\documentclass[runningheads]{llncs}
\usepackage[T1]{fontenc}

\usepackage{graphicx}
\usepackage{url}
\usepackage{cite}
\usepackage{amsmath}
\usepackage{subfig}
\usepackage{xspace}

\input{preamble}

\begin{document}
\title{Road Grip Uncertainty Estimation Through Surface State Segmentation}
%
\author{Jyri Maanp\"a\"a\inst{1,2}\orcidID{0000-0001-6772-9611} \and
Julius Pesonen\inst{1,2}\orcidID{0009-0000-9175-7129} \and
Iaroslav Melekhov\thanks{The work was done prior to joining Amazon}$^,$\inst{2}\orcidID{0000-0003-3819-5280} \and
Heikki Hyyti\inst{1}\orcidID{0000-0003-4664-6221} \and
Juha Hyypp\"a\inst{1}\orcidID{0000-0001-5360-4017}}
\authorrunning{J. Maanp\"a\"a et al.}
%
\institute{Finnish Geospatial Research Institute FGI, National Land Survey of Finland, 02150 Espoo, Finland  \and
Department of Computer Science, Aalto University, 02150 Espoo, Finland \\ \email{jyri.maanpaa@nls.fi}}
\maketitle              
\begin{abstract}
Slippery road conditions pose significant challenges for autonomous driving. Beyond predicting road grip, it is crucial to estimate its uncertainty reliably to ensure safe vehicle control. In this work, we benchmark several uncertainty prediction methods to assess their effectiveness for grip uncertainty estimation. Additionally, we propose a novel approach that leverages road surface state segmentation to predict grip uncertainty. Our method estimates a pixel-wise grip probability distribution based on inferred road surface conditions. Experimental results indicate that the proposed approach enhances the robustness of grip uncertainty prediction.

\end{abstract}
\section{Introduction}

Adverse weather conditions present several challenges for autonomous driving. 
According to the Road Weather Management Program by the U.S. Department of Transportation, snowy and icy roads substantially increase crash risks: 24\% of weather-related vehicle crashes in the U.S. occur on snowy, slushy, or icy pavement, and 15\% happen during snowfall or sleet each year~\cite{rwmbyus}. To achieve fully autonomous driving, reliable road grip prediction methods are essential for safe operation in slippery conditions.

A key factor in enhancing the reliability of perception methods for autonomous vehicles is incorporating uncertainty estimates. By quantifying uncertainty, vehicles can proactively adjust their control strategies to prepare for the worst-case scenarios. 

In the context of road area grip prediction, a practical approach to uncertainty estimation is to compute the lower bound of a confidence interval for the predicted grip value probability distribution. This lower confidence limit is critical, as it represents the minimum expected grip with a given probability, directly influencing vehicle safety. An overestimated grip could lead to slipping, while an underestimated grip may result in overly cautious driving. The upper bound of the confidence interval is less critical, as autonomous vehicles should prioritize worst-case scenarios based on the lower limit. 

\begin{figure}[t]
    \centering
    \includegraphics[width=0.9\textwidth]{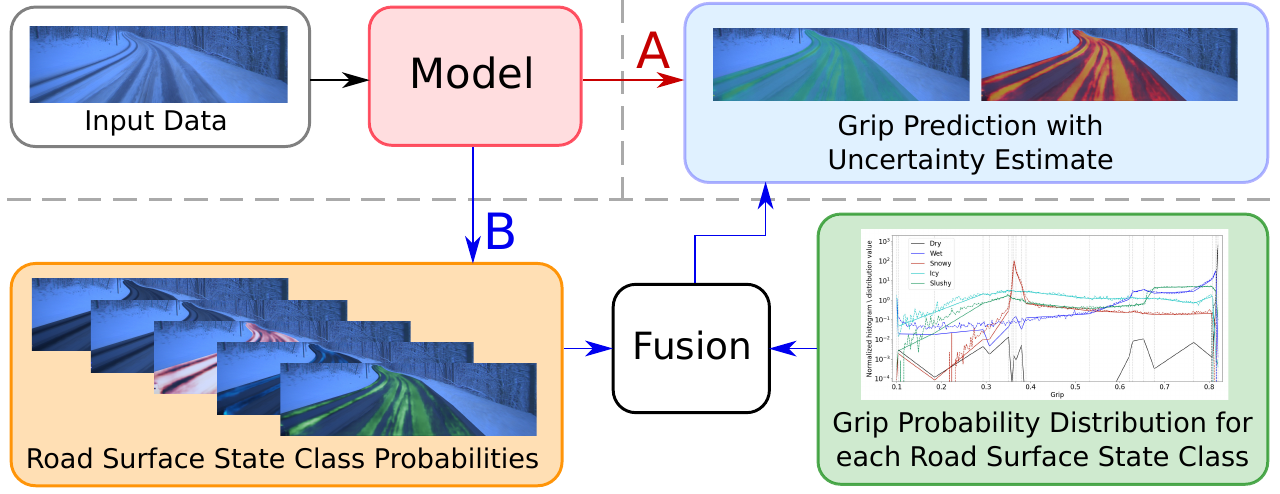}
    \caption{Two ways of approaching the grip uncertainty estimation problem: (A) using a regression model with predictive uncertainty or (B) leveraging the proposed method, which combines road surface state segmentation with grip probability distributions for each surface type to estimate grip and its uncertainty.} \label{fig:main_idea}
\end{figure}

Grip prediction can be approached in two fundamentally different ways: regression-based prediction, where grip values are directly estimated for each road region, and surface state segmentation, where grip is inferred based on the detected road surface layer type (\eg, dry asphalt, ice, snow). In slippery conditions, road surfaces are often covered with varying layers of water, ice, or snow, which strongly influence grip values. For instance, dry asphalt typically has a grip value of 0.82, while snowy conditions generally range between 0.3 and 0.4~\cite{maanpaa2024}. The effectiveness of these approaches also depends on sensor capabilities,~\ie, some sensors can detect surface types but may lack the precision to measure layer thickness or other properties affecting grip. Furthermore, road surface segmentation can be leveraged to estimate grip uncertainty more effectively. Since each surface type has a characteristic grip distribution, the uncertainty in grip prediction can be derived by combining class probabilities with their respective grip distributions which could lead to a more robust uncertainty estimate compared to direct regression methods. The distinction between these two methods is illustrated in~\figref{fig:main_idea}.

In this work, we propose a novel grip uncertainty estimation approach based on road surface state segmentation. We benchmark various uncertainty estimation methods for the task and explore whether grip prediction should be framed as a segmentation task rather than a direct regression problem. All experiments are conducted using data from our previous work in Maanp\"a\"a~\etal~\cite{maanpaa2024}.

\section{Related Work}

Predicting road grip ahead of a vehicle is challenging due to the difficulty of collecting ground truth data. Measuring road grip is complex, and associating these measurements with long-range sensor data across numerous samples adds further complications. Several non-contact methods exist for grip estimation, including computer vision, infrared spectroscopy, optical polarization, and radar detection~\cite{ma2022current}.

Maanp\"a\"a~\etal in~\cite{maanpaa2024} proposed a direct grip regression approach, where grip measurements from an optical road weather sensor are matched with data from an RGB camera, thermal cameras, and a LiDAR reflectance channel. This dataset was used to train a convolutional neural network to predict pixel-wise road grip values. Similarly, Ojala~\etal~\cite{ojala2024} utilized grip measurements from an optical road weather sensor and RGB images to predict a single grip estimate per frame, along with an uncertainty interval. In contrast, several works treat grip prediction as a road surface state classification problem. Roychowdhury~\etal~\cite{roychowdhury2018} classified road conditions from RGB images to estimate grip on a $3\times5$ grid over the road. Other studies adopt a two-stage approach, first classifying the road surface state and then performing grip regression on the classified samples~\cite{langstrand2023, sabanovic2020}.

Various methods exist for estimating uncertainty in neural network regression tasks. Common approaches include ensembling~\cite{ensembling} and Monte Carlo dropout sampling~\cite{mcdropout}, which generate multiple predictions for the same sample, allowing a Gaussian distribution to be fitted based on the mean and standard deviation. Gaussian regression~\cite{gaussian} directly models a Gaussian distribution, while quantile regression~\cite{quantile} predicts confidence intervals without explicitly modeling the output distribution. However, many of these methods struggle with overconfidence under domain shifts~\cite{gustafsson2023how, ovadia2019}, highlighting the need for more robust uncertainty estimation techniques.

\section{Methods}

In this section, we present the uncertainty estimation methods used in our benchmark and our method to predict road grip via road surface state segmentation. 

\subsection{Baseline Uncertainty Regression Methods}

We chose the following four standard uncertainty regression methods for our benchmark:

\boldparagraph{Ensembling.} We train an ensemble $\{f_{\theta_1},...,f_{\theta_M}\}$ of $M=5$ models on the regular grip prediction task. During inference, one can estimate the predicted uncertainty via a Gaussian distribution defined by the mean~$\mu(x)=\frac{1}{M}\sum_{i=1}^Mf_{\theta_i}(x)$ and variance~$\sigma^2(x)=\frac{1}{M}\sum_{i=1}^M\left(\mu(x) - f_{\theta_i}(x)\right)^2$ of the predictions from the~$M$ trained models. The models are trained with different random generator initializations and slight changes in training parameters.

The loss used in the regular grip prediction task is the weighted mean square error:
\begin{align}\label{eq:grip_loss}
    \mathcal{L}(x,y,w|\theta) = \frac{1}{N}\sum_{i=1}^N w_i (y_i - f_i(x|\theta))^2
\end{align}
Here $x$ is the input image tensor, $N$ is the number of pixels containing ground truth grip values in the sample, $w_i$ is the weight for pixel~$i$, $y_i$, and $f_i(x|\theta)$ are the ground truth and model output for grip value at pixel~$i$ and~$\theta$ corresponds to the model parameters.

\boldparagraph{Monte Carlo Dropout Sampling.} We train a model with a dropout layer with dropout parameter~$p=0.5$ as the second last layer of the model on the regular grip prediction task. During inference, the dropout layer is used in training mode and sampled $M=10$ times for each input sample, resulting in $M$ different model outputs. These are similarly used to estimate the mean~$\mu(x)$ and variance~$\sigma^2(x)$ for a Gaussian distribution as in the ensembling approach.

\boldparagraph{Gaussian Probabilistic Regression.} We train the model to predict two outputs which directly correspond to the mean~$\mu(x)$ and variance~$\sigma^2(x)$ of a Gaussian distribution. The model has two output channels from the last layer, which correspond to these mean and variance estimates. The model is trained with the following loss~\cite{gaussian}:
\begin{align}\label{eq:gaussian_loss}
    \mathcal{L}(x,y,\mu,\sigma,w|\theta) =  \frac{1}{N}\sum_{i=1}^N w_i \left(\frac{1}{2\sigma^2_i(x|\theta)}(y_i - \mu_i(x|\theta))^2 + \frac{1}{2}\log\sigma^2_i(x|\theta)\right)~.
\end{align}
As the practical implementation requires, we applied a change of variables and predicted $s(x|\theta):=\log\sigma^2(x|\theta)$, which stabilizes the training process by avoiding the division by zero in the loss function.

\boldparagraph{Quantile Regression.} We train a model to predict two outputs which correspond to the lower quantile~$q_{\text{low}}(x)$ and upper quantile $q_{\text{high}}(x)$ of the grip distribution, forming a confidence interval~$\left[\alpha_{\text{low}}, \alpha_{\text{high}}\right]$. These quantile predictions are obtained from the two output channels of the network, and they are trained with the following loss:
\begin{align}\label{eq:loss_quantile}
    \mathcal{L}(x,y,w|\theta) := \frac{1}{N}\sum_{i=1}^N w_i\left(\rho\left( y_i, q_{\text{low},i}(x|\theta), \alpha_{\text{low}} \right) + \rho\left(y_i, q_{\text{high},i}(x|\theta), \alpha_{\text{high}} \right) \right)
\end{align}
where $q_{\text{low},i}(x_i|\theta)$ and $q_{\text{high},i}(x_i|\theta)$ correspond the lower and upper quantile predictions for pixel~$i$ and $\rho$ corresponds to the pinball loss:
\begin{align}
    \rho(y,\hat{y}, \alpha) := \begin{cases}
    \alpha(y - \hat{y}) & \text{if}~ y - \hat{y} > 0 \\
    (1-\alpha)(\hat{y} - y) & \text{otherwise}.
    \end{cases}
\end{align}
In the experiments, we choose $\alpha_{\text{low}}=0.05$ and $\alpha_{\text{high}}=0.95$ to predict a 90\% confidence interval. As $\alpha_{\text{low}} + \alpha_{\text{high}} = 1$, one can use the mean of predictions $q_{\alpha_{\text{low}},i}(x_i|\theta)$ and $q_{\alpha_{\text{high}}}(x)$ as grip distribution mean. 

\subsection{Grip Uncertainty Prediction via Road Surface State Segmentation (GvRS)}

In the dataset from~\cite{maanpaa2024} each grip measurement also contains a road surface state class estimated by the road weather sensor. These classes are dry, moist, wet, snowy, icy, and slushy. Our initial experiments showed that the dry and moist classes have mostly identical grip value distribution and therefore we merged these classes into a single class which we refer to as the dry class in this work. 

In our approach Grip via Road State (GvRS), we train a model to segment input image pixels to these five classes based on the pixels that contain ground truth for the road surface state class. 
We used the focal loss~\cite{Lin_2017_ICCV} in training:
\begin{align}\label{eq:focal_loss}
    \mathcal{L}(x,y,w|\theta) = \frac{1}{N}\sum_{i=1}^N w_i \left( -(1-p(x|\theta)_{y_i})^\gamma\log(p(x|\theta)_{y_i})\right)~.
\end{align}
Here $p(x|\theta)_{y_i}$ is the softmax output from the model for the ground truth road state class~$y_i$ in the pixel~$i$.

In addition, each class has a grip probability distribution interpolated from the histogram of the corresponding class grip distribution in the training dataset. As the classes are independent, we can fuse the predicted road surface state class probabilities with the grip distributions for each class as follows:
\begin{align}
    p_\text{grip}(g,x) = \sum_{c=1}^K p_c(x|\theta)q_c(g)~,
\end{align}
where $p_\text{grip}(g,x)$ is the grip probability distribution for the input sample~$x$, $p_c(x|\theta)$ is the predicted probability for the class $c=1,..,K$ by the model, and $q_c(g)$ is the grip probability distribution for the class~$c$ evaluated from the training dataset grip values for that class.

In this method, we obtain a continuous grip probability distribution for each input image pixel. As the grip distribution would need to be integrated for each input pixel to obtain the grip confidence limits in online processing, the grip distribution processing needs an efficient implementation presented in Section~\ref{sec:grip_rw_approximation}.
In addition, one could use the ground truth road state classes as road state class predictions to simulate the grip uncertainty output of a perfectly accurate road state prediction model (Ideal GvRS in results).

\section{Experiments}

In this section, we first define our dataset and describe our approach for efficient grip probability distribution evaluation using road surface state predictions. We then present our training setup, evaluation metrics, and both quantitative and qualitative results on the test datasets.

\subsection{Dataset}

We used the dataset from Maanp\"a\"a~\etal~\cite{maanpaa2024}, 
which contains 37~hours (1538~kilometers) of car sensor data collected under diverse driving conditions in Finland.
The dataset includes data from an RGB camera, a LiDAR sensor, and three thermal cameras, all matched pixel-wise with measurements from a Vaisala Mobile Detector MD30 road weather sensor. This matching was performed using 3D transformations, sensor calibrations, and an accurately post-processed GNSS trajectory. In total, the dataset comprises 237\ 067 samples recorded at 2FPS. Further details on the dataset and preprocessing can be found in~\cite{maanpaa2024, pesonen2023pixelwise}.

\begin{figure}[t]
    \centering
    \subfloat{\includegraphics[width=0.95\textwidth]{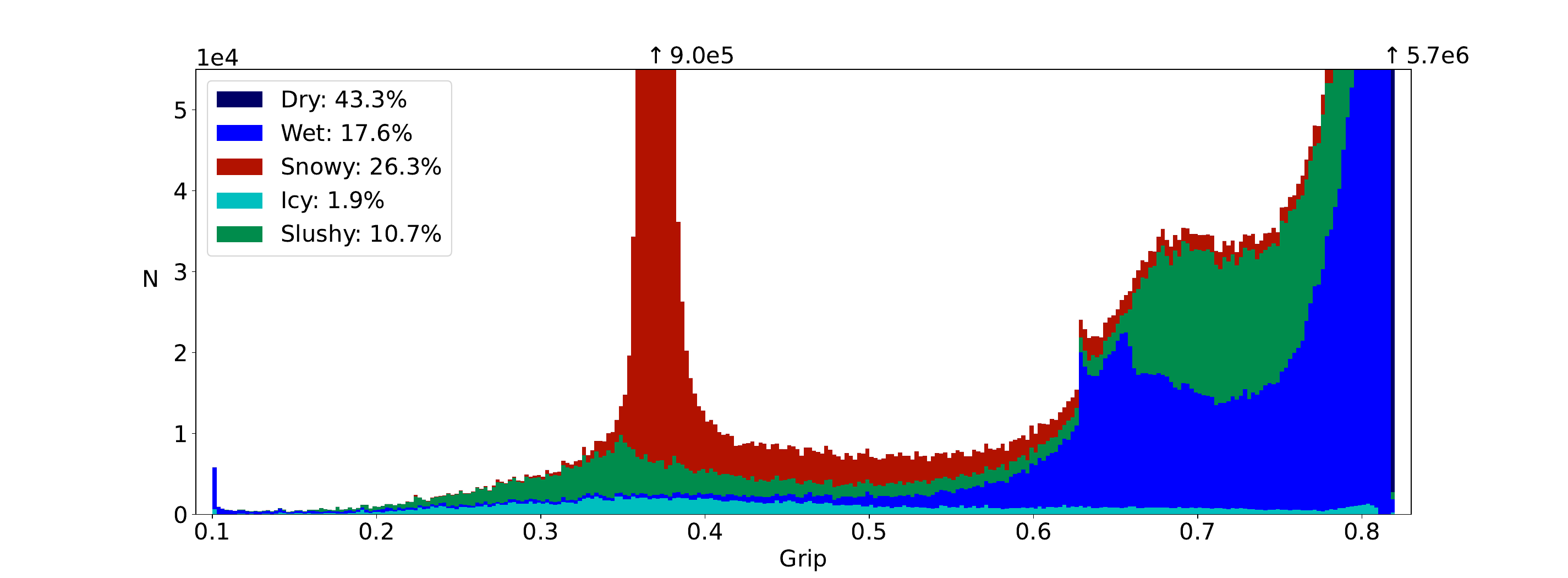}}
    \caption{Stacked histogram of grip values for different road surface state classes in the training set.} \label{fig:dataset_histograms}
\end{figure}

In this work, we used only the RGB camera samples along with corresponding road weather sensor measurements, as previous work~\cite{maanpaa2024} demonstrated that RGB data alone provides sufficient accuracy for grip prediction.  
The road weather sensor records multiple surface-related measurements, from which we utilized grip values and road surface state classifications. The histogram of road surface state measurements with respect to the corresponding grip values is in~\figref{fig:dataset_histograms}.

The dataset was split following the geofencing approach in~\cite{maanpaa2024}, ensuring a minimum 55-meter separation between samples across different data splits. The validation and test sets have grip distributions similar to the training set. After filtering, the dataset consists of 159\ 801 training samples (79.1\%), 15\ 343 validation samples (7.6\%), and 26\ 783 test samples (13.3\%).

In addition, we use three independent test drives (Test Drive 1, 2, and 3 from~\cite{maanpaa2024}) to evaluate out-of-distribution performance. These test drive datasets, collected on different days and roads, contain 5746, 2042, and 8351 samples, respectively. Test Drive 1 was recorded during snowfall in dark, snowy conditions, Test Drive 2 on a snowy road in daylight, and Test Drive 3 in wet and slushy conditions.

\subsection{Grip Distribution Models for Road Surface State Classes}\label{sec:grip_rw_approximation}

\begin{figure}[t]
    \centering
    \includegraphics[width=0.9\textwidth]{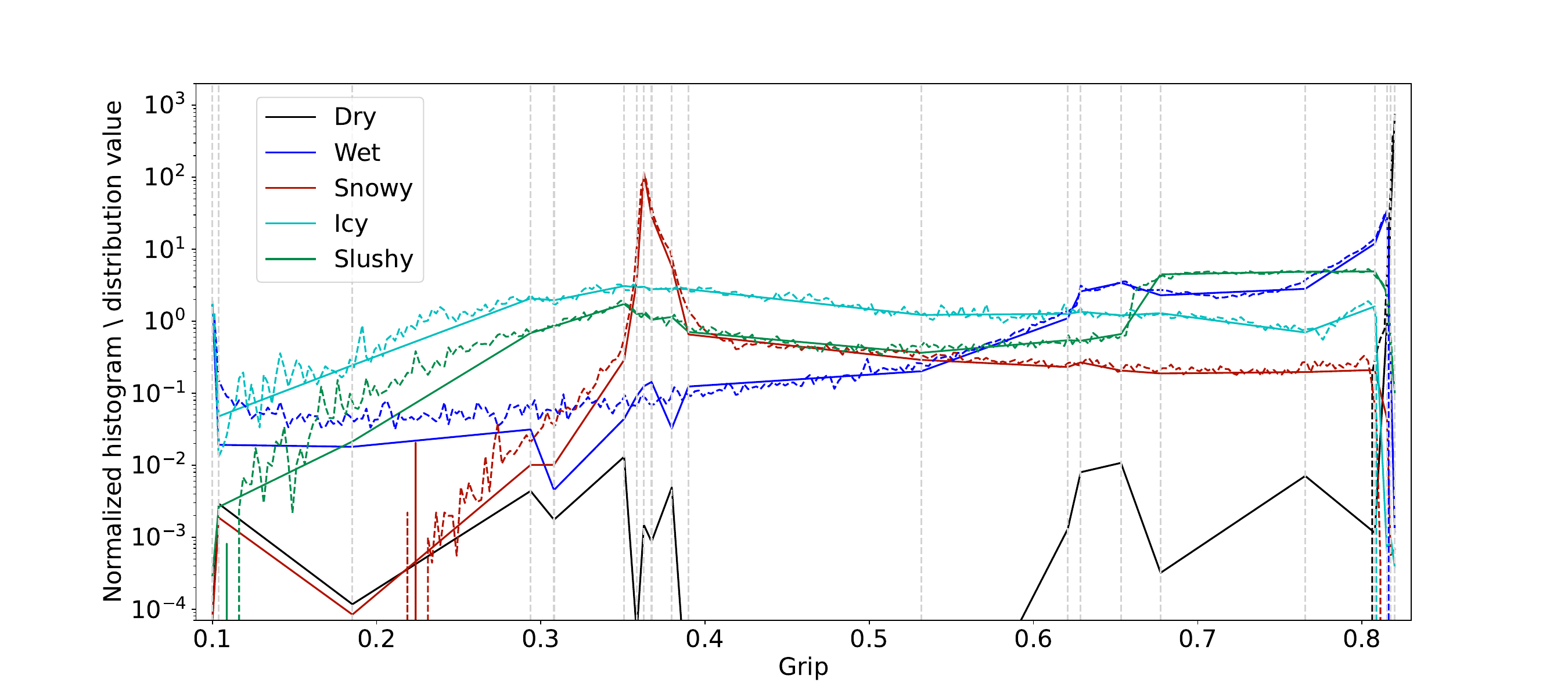}
    \caption{Normalized grip histograms (dashed line) and corresponding fitted grip distributions (solid line) for each road surface state class. Vertical lines show the knots of the piecewise linear functions approximating the grip distributions.} \label{fig:roadstate_grip_distributions}
\end{figure}

The proposed method for predicting grip uncertainty via road surface state segmentation requires an efficient approach to estimate the grip probability density distribution for each road surface state class. 
To achieve this, we approximate these distributions using piecewise linear functions with 20 intervals. The segment boundaries are optimized to minimize the sum of mean squared errors between the normalized grip histograms from the training set and the corresponding piecewise linear functions for each class. 
We initialized the segment boundaries with a manual guess and refined them with MATLAB's \texttt{fminsearch} function, which utilizes the Nedler-Mead simplex method~\cite{simplex} for optimization. The resulting probability density function approximations are visualized in~\figref{fig:roadstate_grip_distributions}. These piecewise linear functions can also be integrated to obtain percentile-based grip confidence limits.

\subsection{Training parameters}

We used a Feature Pyramid Network (FPN)~\cite{lin2017feature} with a ResNet18~\cite{he2016deep} encoder for all experiments in this benchmark. This model was selected based on its highest validation accuracy in previous work~\cite{maanpaa2024} and its sufficient complexity for our study. All models were trained using the AdamW optimizer~\cite{loshchilov2019decoupled} with a learning rate of ~$1\mathrm{e}-3$ and a weight decay of ~$1\mathrm{e}-3$. We used $\gamma=0.5$ in the focal loss~\eqref{eq:focal_loss} when training our Grip via Road State -method. Training was conducted for 40 epochs with a batch size of 32, and model validation was performed using the same loss function applied during training. 

Following the approach in~\cite{maanpaa2024}, we incorporated pixel-wise weighting in all loss functions to balance prediction accuracy across the entire road area, as most samples contain a higher density of pixels farther from the vehicle. The weight assigned to each pixel is determined by its $y$-coordinate in the image, decreasing linearly from the bottom of the estimated horizon level, with normalization ensuring a mean weight of one.

To augment the training data, we applied the following transformations to each sample: random color jitter (30\% probability), slight random blur (30\% probability), small random scaling and rotation (30\% probability), and horizontal flipping (50\% probability).

\subsection{Metrics}

We present several metrics to evaluate the reliability of the uncertainty estimates produced by the tested models. All metrics are first computed per sample and then averaged over the entire test set. To assess grip distribution accuracy, we report the root mean squared error (RMSE) for the mean and median grip predictions ($\text{RMSE}(\mu_\text{mean})$ and $\text{RMSE}(\mu_\text{median})$), calculated as the weighted mean squared error per sample, averaged across the test set, followed by a square root operation.  
For uncertainty evaluation, We report the fraction of ground truth grip values that fall within the predicted 68.3\% confidence interval $F(g\in L_\sigma)$, the 90\% confidence interval $F(g\in L_\text{90\%})$, and the fraction exceeding the 5th percentile bound $F(g >P_\text{5\%})$, which corresponds to the lower limit of the 90\% confidence interval. Additionally, we provide the average confidence interval length~$\mu(P_\text{95\%}-P_\text{5\%})$ and the mean 5th percentile value~$\mu(P_\text{5\%})$ to offer insight into model behavior. 

To ensure reasonable classification of dry road samples, confidence intervals are constrained within $[0.1,0.82]$. However, for evaluation, the lower bound is restricted to $[0.1,0.81]$, and upper bound values above $0.81$ are rounded up to $0.82$. This adjustment prevents overly narrow confidence intervals from misclassifying dry road conditions. 

Finally, we analyze cases where the ground truth grip falls below the predicted 5th percentile ($g<P_\text{5\%}$), reporting the median~$p_\text{50\%}$, 70th percentile~$p_\text{70\%}$, and 90th percentile~$p_\text{90\%}$ of the deviation between the ground truth and the lower confidence limit $(P_\text{5\%} - g)$. These metrics characterize the accuracy of the 5th percentile predictions.

\begin{figure}[t!]
    \centering
    \subfloat{\includegraphics[width=0.34\textwidth]{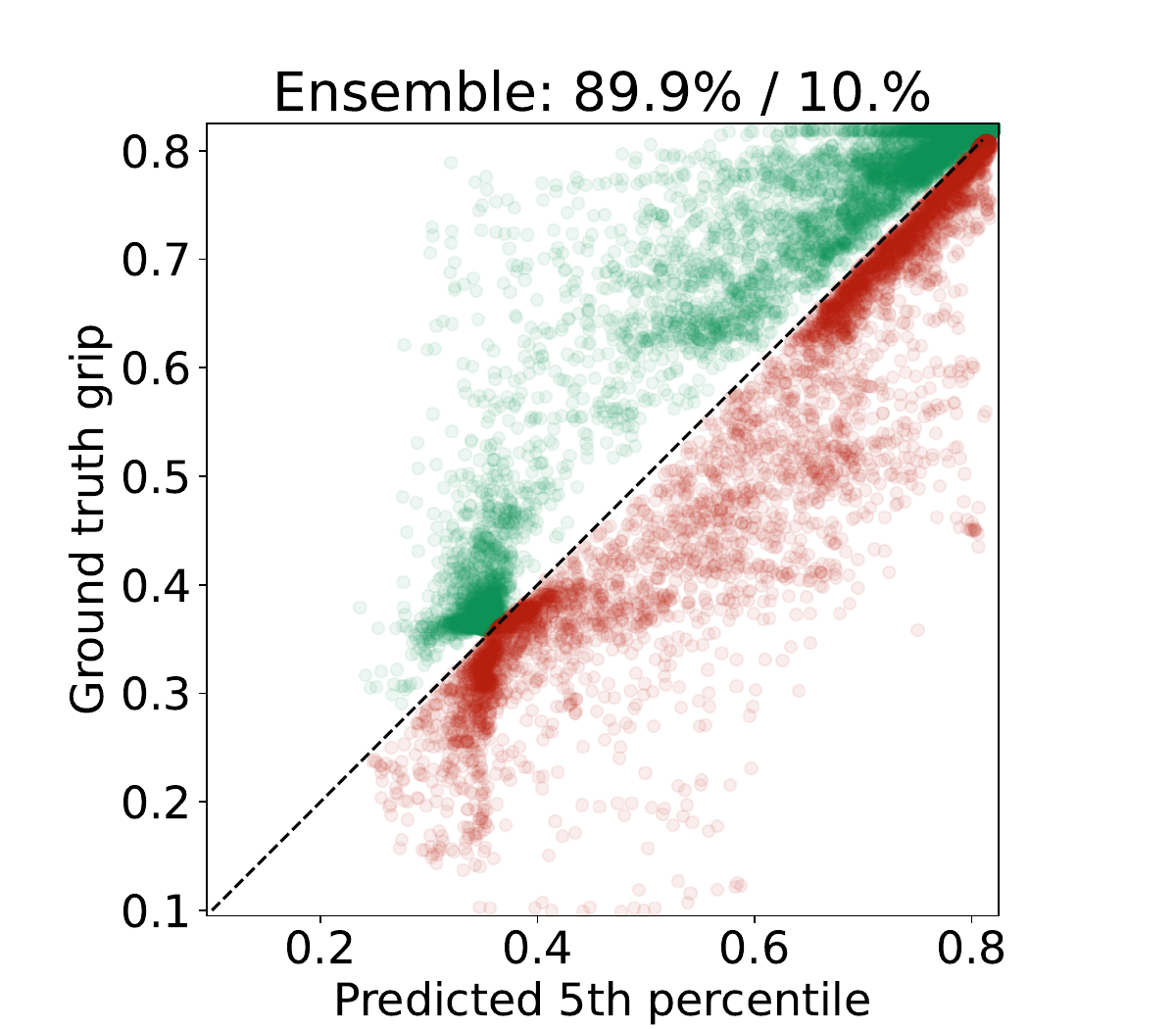}}
    \hspace{-0.4cm}
    \subfloat{\includegraphics[width=0.34\textwidth]{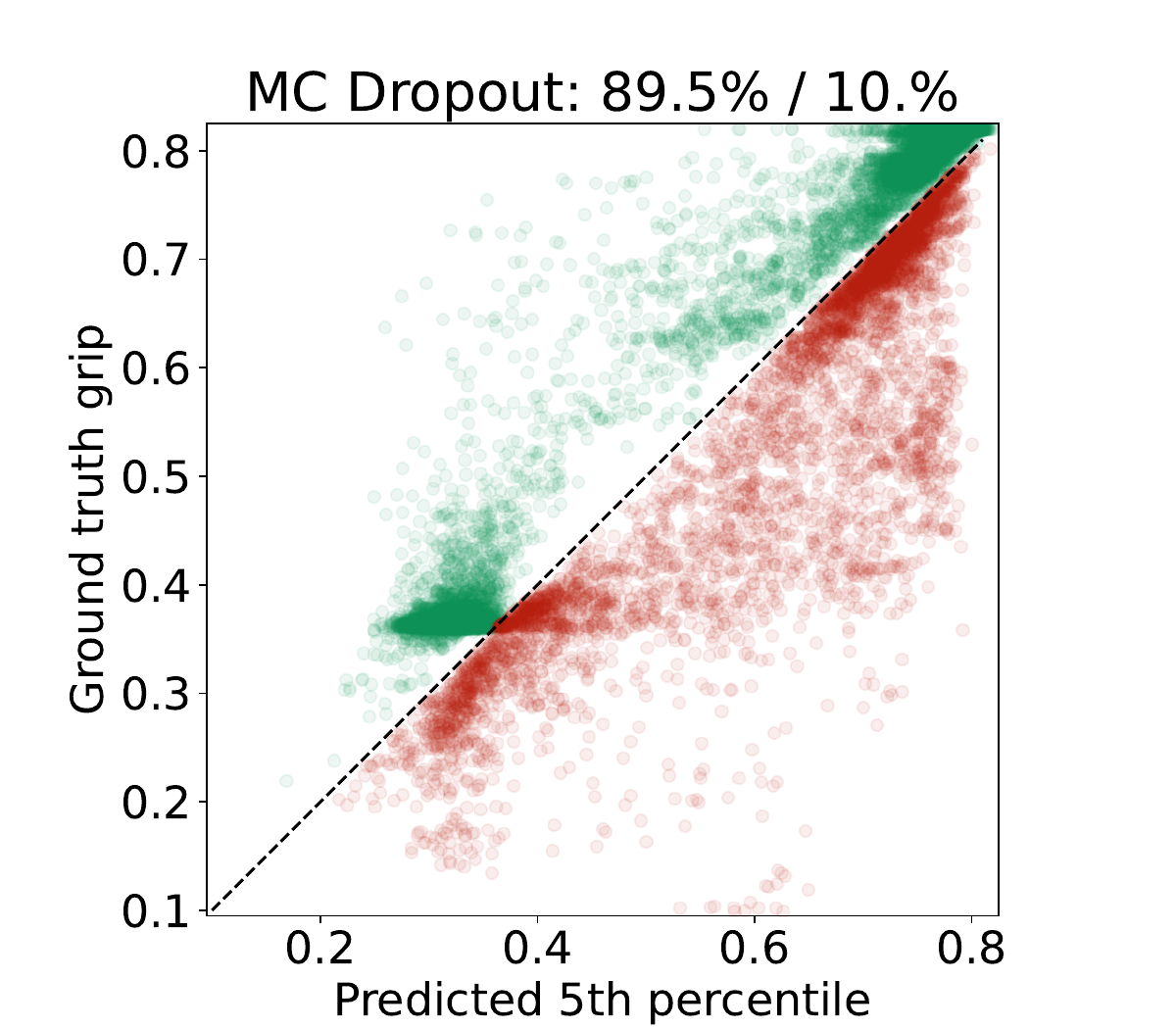}}
    \hspace{-0.4cm}
    \subfloat{\includegraphics[width=0.34\textwidth]{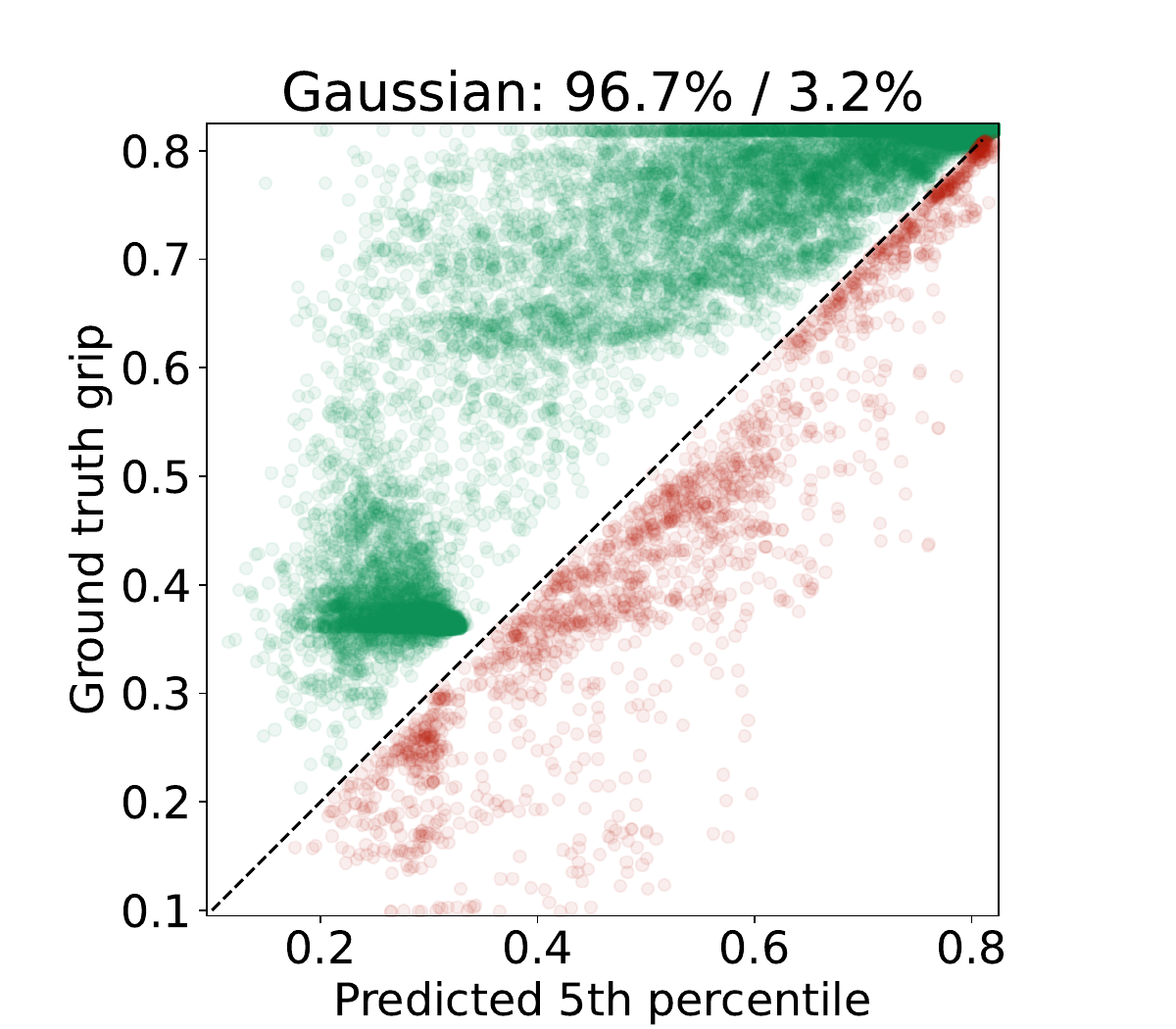}} \\
    \vspace{-0.2cm}
    \subfloat{\includegraphics[width=0.34\textwidth]{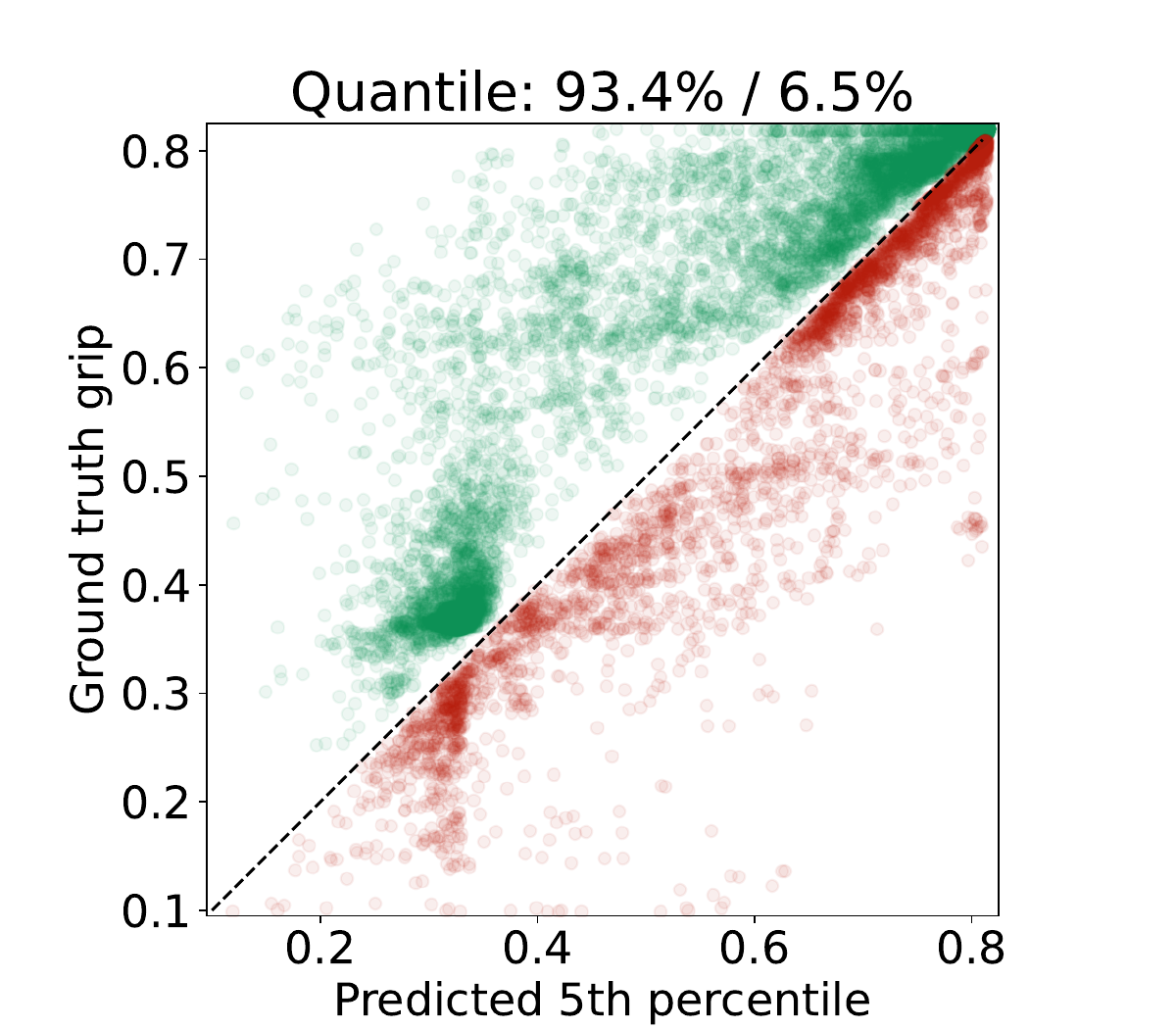}}
    \hspace{-0.4cm}
    \subfloat{\includegraphics[width=0.34\textwidth]{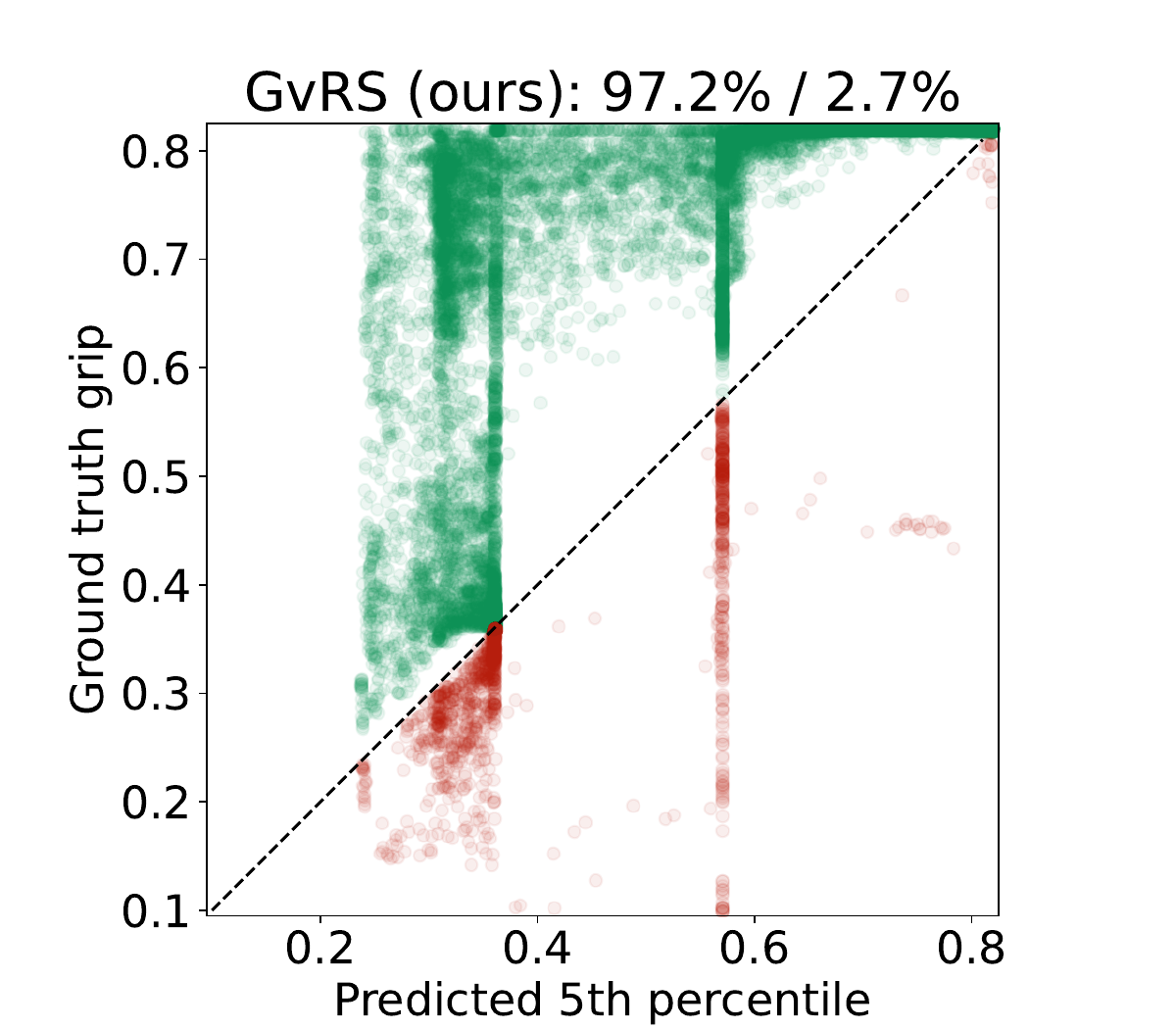}}
    \hspace{-0.4cm}
    \subfloat{\includegraphics[width=0.34\textwidth]{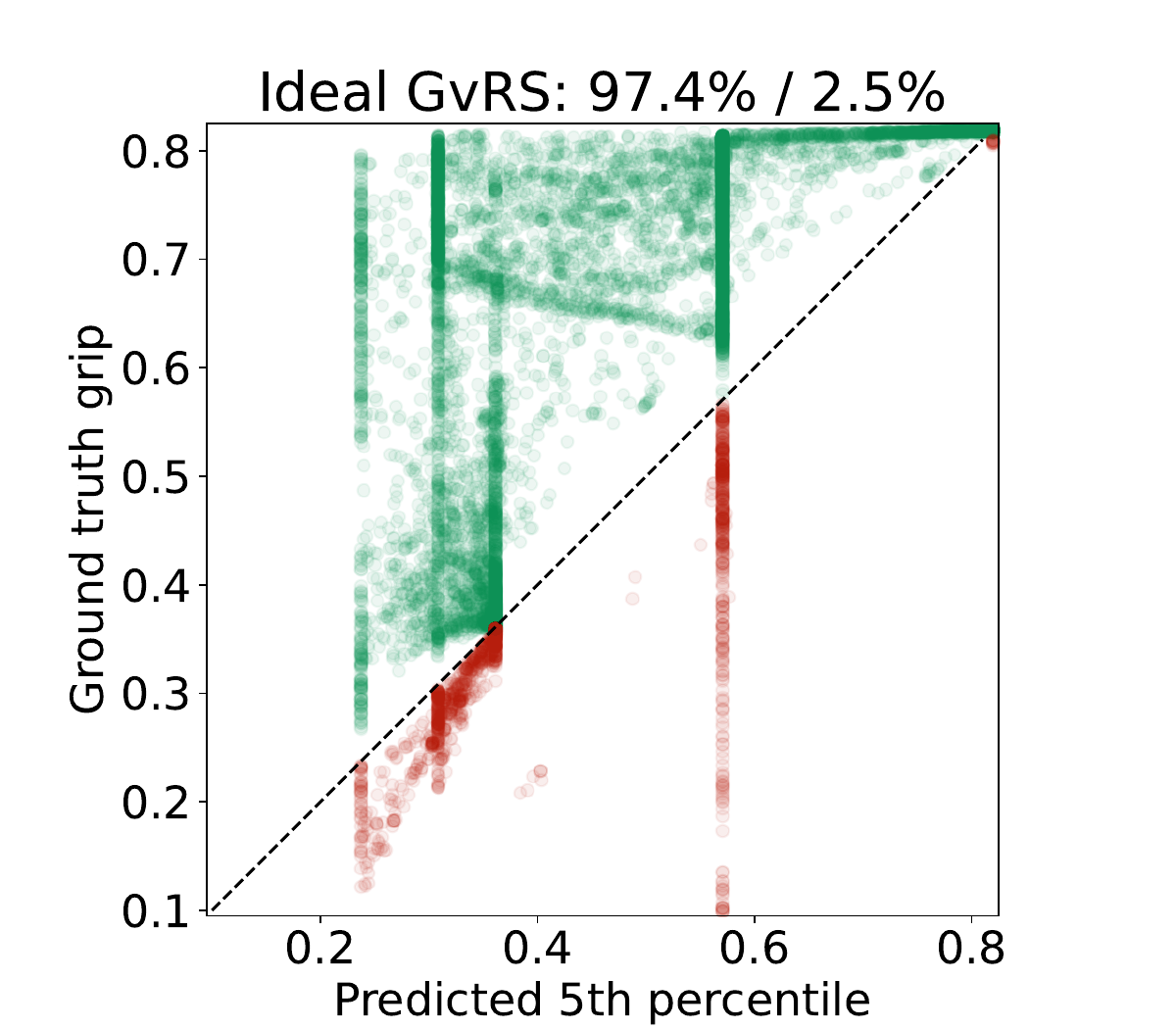}}
    \caption{Scatter plots of the test set samples showing the relation between ground truth grip and the corresponding predicted 5th percentile for each model. The average percentage of points over / under the 5th percentile limit are shown in each title. Each point represents the averaged metrics in one test set sample, shown in green class if 90\% of the ground truth grip measurements in the sample are over the 5th percentile limit, and shown in red class otherwise.} \label{fig:outlier_scatters} 
\end{figure}

\subsection{Quantitative Results}

\boldparagraph{Test Set.} We present the benchmark results of model performance on the test set in ~\tabref{tab:test_set}. All models achieve sufficient accuracy in predicting the mean and median grip values, with RMSE values below $0.1$. 
The ensemble and Monte Carlo dropout models demonstrate the highest accuracy for the 68.3\% confidence interval; however, they struggle with the 90\% confidence interval, indicating difficulty in capturing the tails of the grip probability distribution. 
The quantile regression model performs best in predicting the 90\% confidence interval, as it is explicitly trained for confidence interval estimation. In contrast, the Gaussian probabilistic regression and the proposed GvRS models tend to predict wider confidence intervals. Despite this, the GvRS model exhibits robust predictions for the lower 5th percentile limit, as indicated by the low deviation $P_\text{5\%} - g$ for cases where $g<P_\text{5\%}$. 
The ideal GvRS reference further suggests that improving road surface state prediction would enhance this robustness. The GvRS model achieves 94.9\% accuracy in road surface state classification. Additional validation results, including model runtime analysis, are provided in the supplementary material. 

\begin{table}[t]
\centering
\caption{Benchmark on different metrics for models on the test set. The best result among models is bolded for each metric if there is an interpretation for optimal metric value (shown in brackets). This excludes 'Ideal GvRS' as it is only an ideal reference for GvRS model accuracy and not a real model.}
\label{tab:test_set}
\begin{tabular}{|lrrrrrr|}
\hline
\multicolumn{1}{|l|}{\textbf{Metric}}                                              & \multicolumn{1}{l|}{\textbf{Ensemble}} & \multicolumn{1}{l|}{\textbf{\begin{tabular}[c]{@{}l@{}}MC\\ Dropout\end{tabular}}} & \multicolumn{1}{l|}{\textbf{Gaussian}} & \multicolumn{1}{l|}{\textbf{Quantile}} & \multicolumn{1}{l|}{\textbf{\begin{tabular}[c]{@{}l@{}}GvRS\\ (ours)\end{tabular}}} & \multicolumn{1}{l|}{\begin{tabular}[c]{@{}l@{}}Ideal\\ GvRS\end{tabular}} \\ \hline
\multicolumn{1}{|l|}{$\text{RMSE}(\mu_\text{mean})$ ($\downarrow$)}                          & \multicolumn{1}{r|}{\textbf{0.0580}}   & \multicolumn{1}{r|}{0.0626}                                                        & \multicolumn{1}{r|}{0.0700}            & \multicolumn{1}{r|}{0.0602}            & \multicolumn{1}{r|}{0.0882}                                                         & 0.0866                                                                    \\ \cline{1-1}
\multicolumn{1}{|l|}{$\text{RMSE}(\mu_\text{median})$ ($\downarrow$)}                        & \multicolumn{1}{r|}{\textbf{0.0580}}   & \multicolumn{1}{r|}{0.0626}                                                        & \multicolumn{1}{r|}{0.0700}            & \multicolumn{1}{r|}{-}                 & \multicolumn{1}{r|}{0.0958}                                                         & 0.0932                                                                    \\ \cline{1-1}
\multicolumn{1}{|l|}{$F(g\in L_\sigma)$ ($\rightarrow$ 68.3)[$\%$]}    & \multicolumn{1}{r|}{\textbf{69.4}}     & \multicolumn{1}{r|}{73.3}                                                          & \multicolumn{1}{r|}{91.1}              & \multicolumn{1}{r|}{-}                 & \multicolumn{1}{r|}{87.9}                                                           & 84.5                                                                      \\ \cline{1-1}
\multicolumn{1}{|l|}{$F(g\in L_\text{90\%})$ ($\rightarrow$ 90)[$\%$]} & \multicolumn{1}{r|}{79.6}              & \multicolumn{1}{r|}{79.6}                                                          & \multicolumn{1}{r|}{96.2}              & \multicolumn{1}{r|}{\textbf{90.2}}     & \multicolumn{1}{r|}{96.5}                                                           & 95.8                                                                      \\ \cline{1-1}
\multicolumn{1}{|l|}{$F(g >P_\text{5\%})$ ($\rightarrow$ 95)[$\%$]}    & \multicolumn{1}{r|}{89.9}              & \multicolumn{1}{r|}{89.5}                                                          & \multicolumn{1}{r|}{96.7}              & \multicolumn{1}{r|}{\textbf{93.4}}     & \multicolumn{1}{r|}{97.3}                                                           & 97.4                                                                      \\ \cline{1-1}
\multicolumn{1}{|l|}{$\mu(P_\text{95\%}-P_\text{5\%})$}                            & \multicolumn{1}{r|}{0.0375}            & \multicolumn{1}{r|}{0.0586}                                                        & \multicolumn{1}{r|}{0.1294}            & \multicolumn{1}{r|}{0.0679}            & \multicolumn{1}{r|}{0.1995}                                                         & 0.1573                                                                    \\ \cline{1-1}
\multicolumn{1}{|l|}{$\mu(P_\text{5\%})$}                                          & \multicolumn{1}{r|}{0.656}             & \multicolumn{1}{r|}{0.643}                                                         & \multicolumn{1}{r|}{0.605}             & \multicolumn{1}{r|}{0.639}             & \multicolumn{1}{r|}{0.576}                                                          & 0.601                                                                     \\ \hline
\multicolumn{7}{|l|}{\textbf{For} $g<P_\text{5\%}$:}                                                                                                                                                                                                                                                                                                                                                                                                                 \\ \hline
\multicolumn{1}{|l|}{$p_\text{50\%}(P_\text{5\%} - g)$ ($\downarrow$)}             & \multicolumn{1}{r|}{0.0213}            & \multicolumn{1}{r|}{0.0316}                                                        & \multicolumn{1}{r|}{0.0360}            & \multicolumn{1}{r|}{0.0212}            & \multicolumn{1}{r|}{\textbf{0.0197}}                                                & 0.0125                                                                    \\ \cline{1-1}
\multicolumn{1}{|l|}{$p_\text{70\%}(P_\text{5\%} - g)$ ($\downarrow$)}             & \multicolumn{1}{r|}{0.0452}            & \multicolumn{1}{r|}{0.0642}                                                        & \multicolumn{1}{r|}{0.0676}            & \multicolumn{1}{r|}{\textbf{0.0438}}   & \multicolumn{1}{r|}{0.0467}                                                         & 0.0306                                                                    \\ \cline{1-1}
\multicolumn{1}{|l|}{$p_\text{90\%}(P_\text{5\%} - g)$ ($\downarrow$)}             & \multicolumn{1}{r|}{0.1317}            & \multicolumn{1}{r|}{0.1844}                                                        & \multicolumn{1}{r|}{0.1444}            & \multicolumn{1}{r|}{\textbf{0.1186}}   & \multicolumn{1}{r|}{0.1224}                                                         & 0.0899                                                                    \\ \hline
\end{tabular}
\end{table}

\figref{fig:outlier_scatters} illustrates the behavior of the predicted 5th percentile across models in relation to the ground truth grip values. The ensemble and Monte Carlo dropout models produce 5th percentile estimates close to the ground truth grip value, but this percentile is overestimated more frequently. 
Meanwhile, the Gaussian and quantile models show greater deviation, often placing the 5th percentile below the actual grip value. The GvRS model, in contrast, tends to predict specific grip values for the 5th percentile, likely due to the higher probability densities associated with certain grip distributions. This results in instances where the predicted 5th percentile is excessively low, even when the uncertainty in grip estimation is minimal. 

\boldparagraph{Outside of Distribution Examples.} \tabref{tab:extra_test_set} presents the chosen error metrics ($\text{RMSE}(\mu_\text{mean})$, $F(g >P_\text{5\%})$ and $p_\text{90\%}(P_\text{5\%} - g)$ for $g<P_\text{5\%}$) for each model on the out-of-distribution test drives. Most models exhibit challenges with overconfident predictions, as indicated by $F(g >P_\text{5\%})$ values falling below 95\%. However, the Gaussian and GvRS models provide more reliable uncertainty estimates in these scenarios. Additionally, the 90\% percentile of the deviation $P_\text{5\%} - g$ for $g<P_\text{5\%}$ remains around $0.1$ in most cases, suggesting that the predicted 5th percentile rarely deviates significantly from the actual grip value in these out-of-distribution tests.

\begin{table}[t]
\caption{Chosen error metrics evaluated on the three test drives for each model.}
\label{tab:extra_test_set}
\begin{tabular}{|l|r|r|r|r|r|r|}
\hline
\textbf{Test Drive / Metric}                                       & \multicolumn{1}{l|}{\textbf{{Ensemble}}} & \multicolumn{1}{l|}{\textbf{\begin{tabular}[c]{@{}l@{}}{MC}\\ {Dropout}\end{tabular}}} & \multicolumn{1}{l|}{\textbf{Gaussian}} & \multicolumn{1}{l|}{\textbf{Quantile}} & \multicolumn{1}{l|}{\textbf{\begin{tabular}[c]{@{}l@{}}GvRS\\ (ours)\end{tabular}}} & \multicolumn{1}{l|}{\begin{tabular}[c]{@{}l@{}}Ideal\\ GvRS\end{tabular}} \\ \hline
\textbf{TD 1}: $\text{RMSE}(\mu_\text{mean})$ $\downarrow$                  & \textbf{0.102}                         & 0.109                                                                              & 0.106                                  & \textbf{0.102}                         & 0.103                                                                               & 0.133                                                                     \\
$F(g >P_\text{5\%})$ ($\rightarrow$ 95){[}$\%${]}                  & 82.2                                   & 80.1                                                                               & \textbf{97.0}                          & 84.8                                   & 88.8                                                                                & 93.3                                                                      \\
$g<P_\text{5\%}$: $p_\text{90\%}(P_\text{5\%} - g)$ $\downarrow$ & 0.0705                                 & 0.1169                                                                             & 0.0780                                 & 0.0875                                 & \textbf{0.0518}                                                                     & 0.0388                                                                    \\ \hline
\textbf{TD 2}: $\text{RMSE}(\mu_\text{mean})$ $\downarrow$                  & 0.124                                  & 0.137                                                                              & 0.126                                  & 0.138                                  & \textbf{0.123}                                                                      & 0.179                                                                     \\
$F(g >P_\text{5\%})$ ($\rightarrow$ 95){[}$\%${]}                  & 91.6                                   & 86.6                                                                               & 92.8                                   & \textbf{93.6}                          & 99.6                                                                                & 99.7                                                                      \\
$g<P_\text{5\%}$: $p_\text{90\%}(P_\text{5\%} - g)$ $\downarrow$ & 0.0849                                 & 0.1227                                                                             & 0.1171                                 & 0.1141                                 & \textbf{0.0442}                                                                     & 0.0120                                                                    \\ \hline
\textbf{TD 3}: $\text{RMSE}(\mu_\text{mean})$ $\downarrow$                  & 0.0902                                 & \textbf{0.0892}                                                                    & 0.0968                                 & 0.0988                                 & 0.0949                                                                              & 0.0920                                                                    \\
$F(g >P_\text{5\%})$ ($\rightarrow$ 95){[}$\%${]}                  & 75.4                                   & 67.5                                                                               & 82.7                                   & 72.4                                   & \textbf{97.7}                                                                       & 95.4                                                                      \\
$g<P_\text{5\%}$: $p_\text{90\%}(P_\text{5\%} - g)$ $\downarrow$ & 0.1024                                 & 0.0981                                                                             & \textbf{0.0882}                        & 0.0898                                 & 0.1637                                                                              & 0.1387                                                                    \\ \hline
\end{tabular}
\end{table}

\subsection{Qualitative Results}

\begin{figure}[t]
\centering
\begin{tabular}{ccc}
\includegraphics[width=0.3\textwidth]{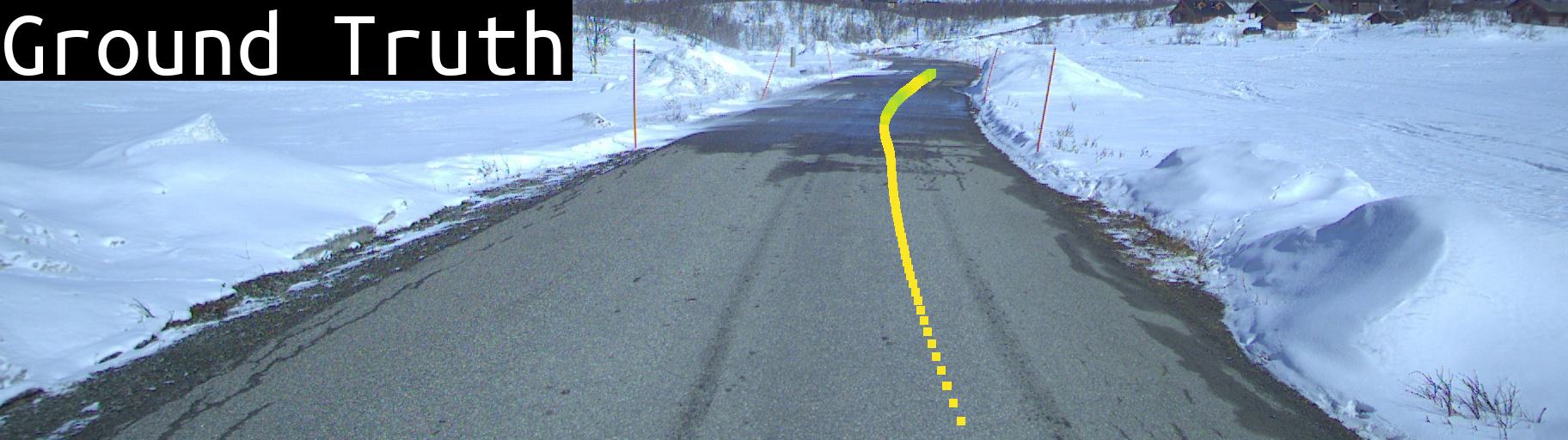} & \includegraphics[width=0.3\textwidth]{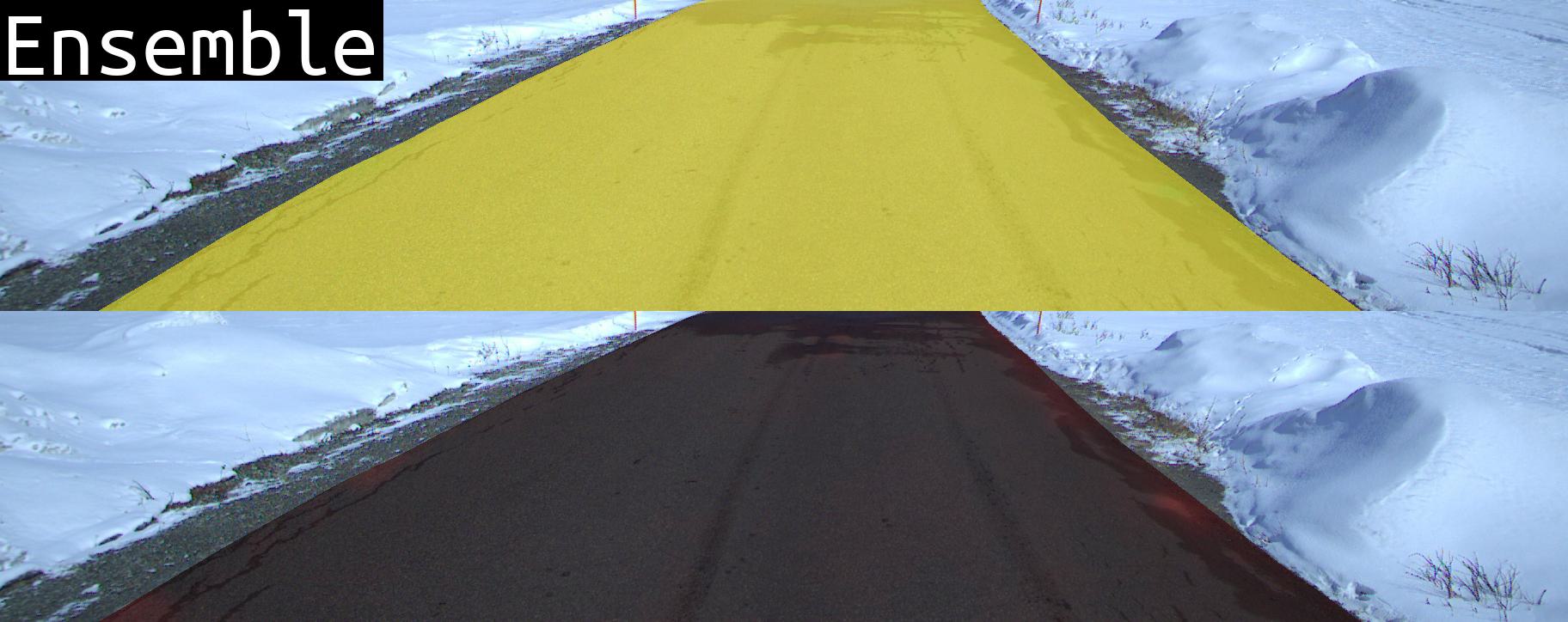} & \includegraphics[width=0.3\textwidth]{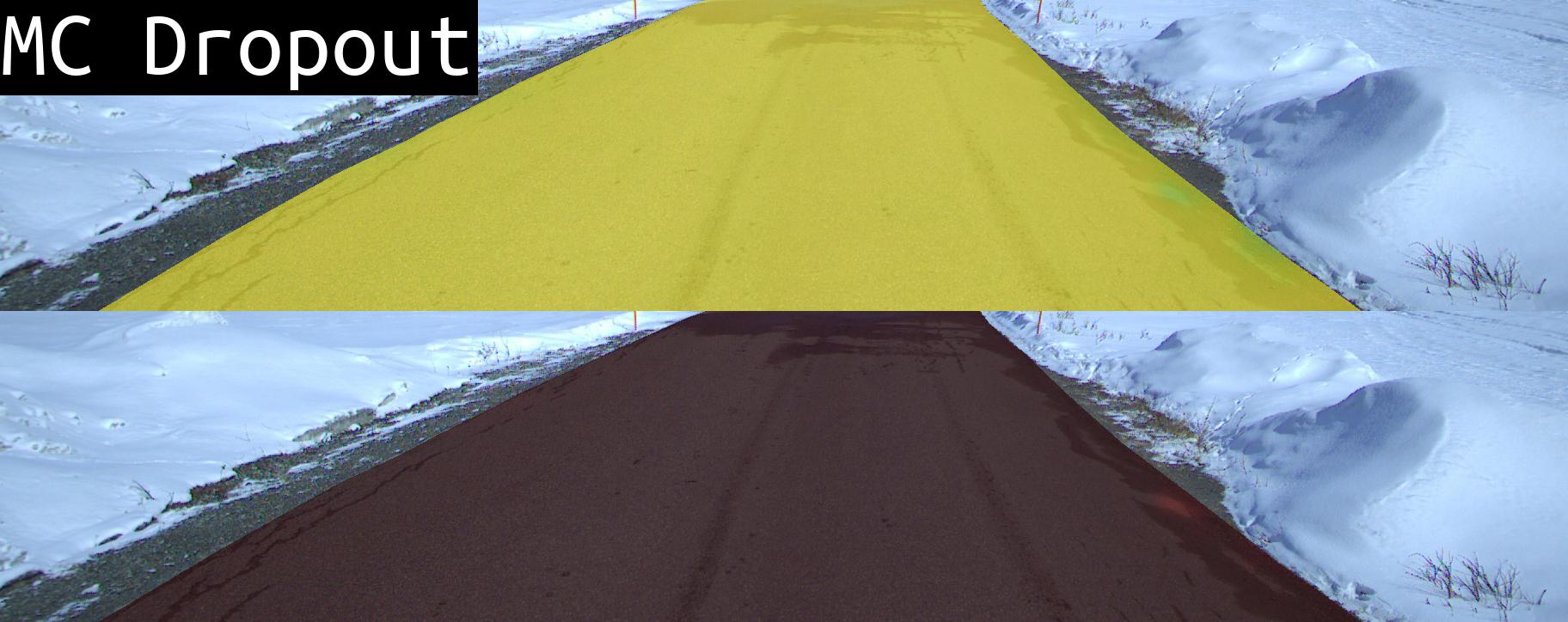} \\
\includegraphics[width=0.3\textwidth]{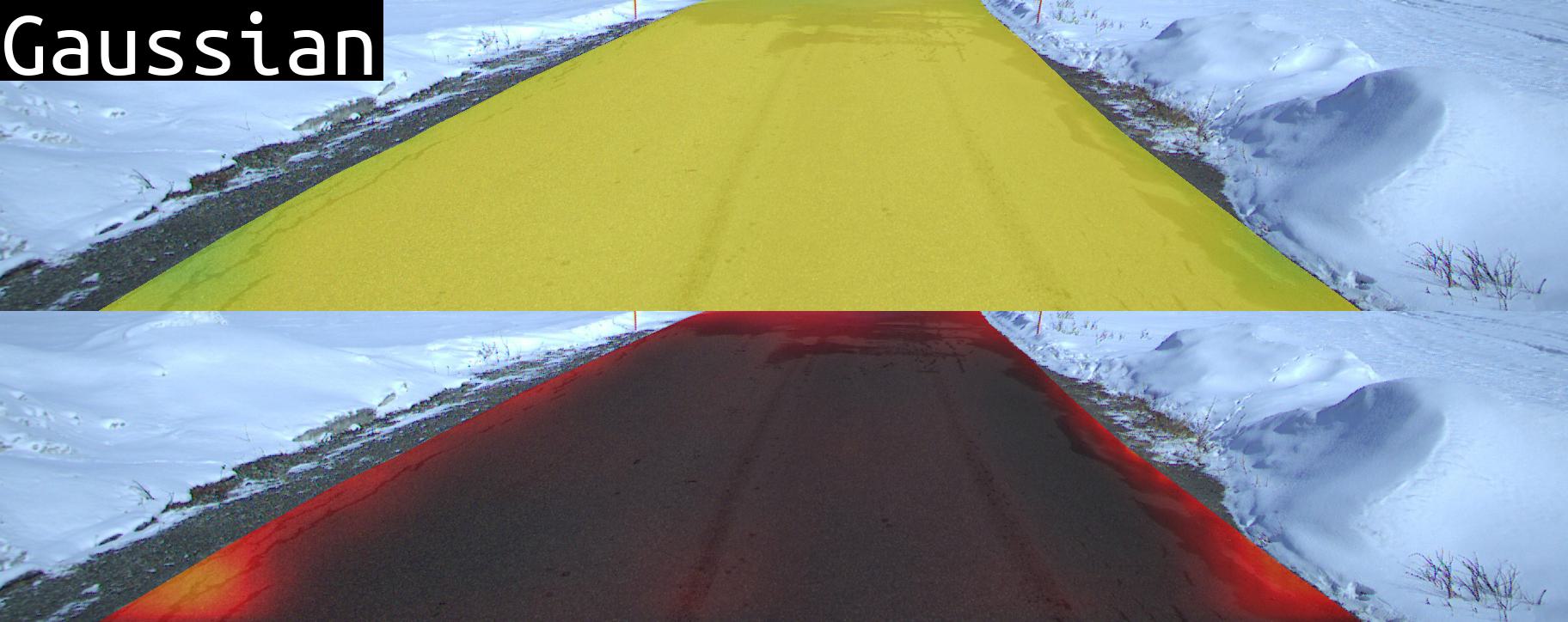} & \includegraphics[width=0.3\textwidth]{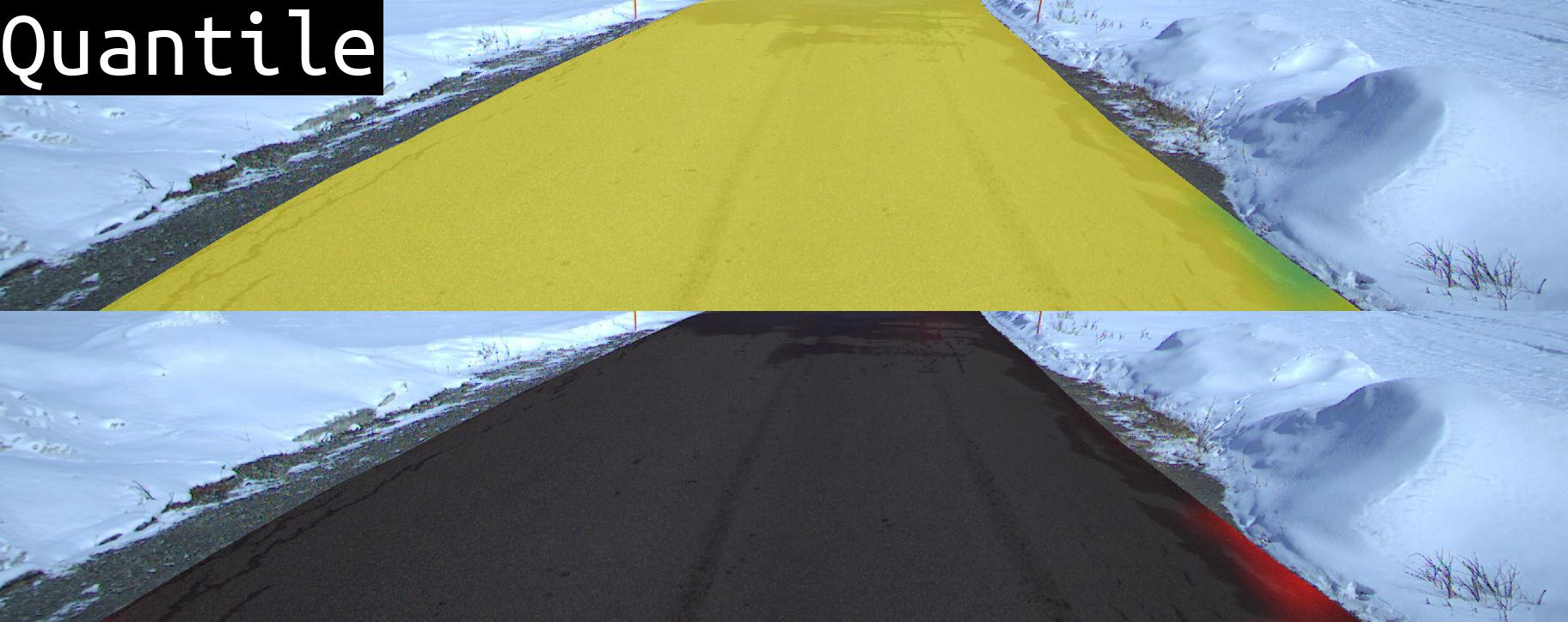} & \includegraphics[width=0.3\textwidth]{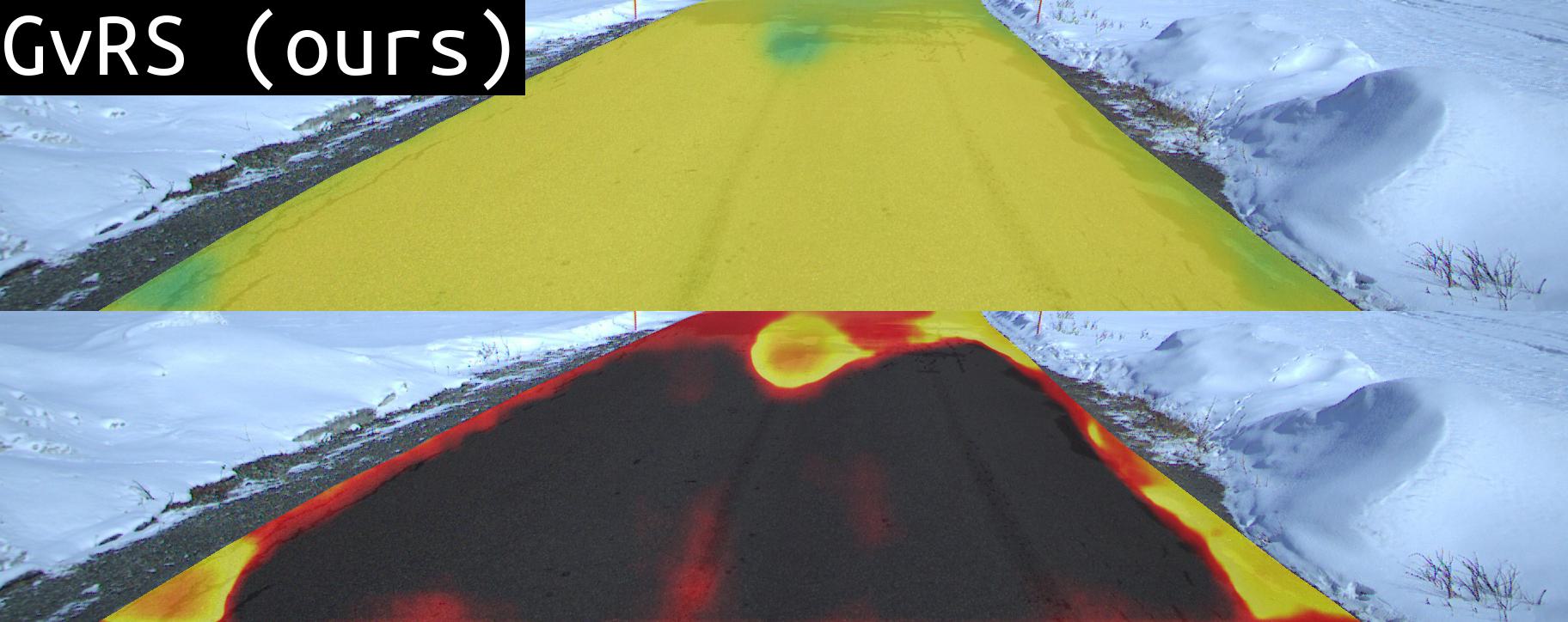} \\ \hline
\includegraphics[width=0.3\textwidth]{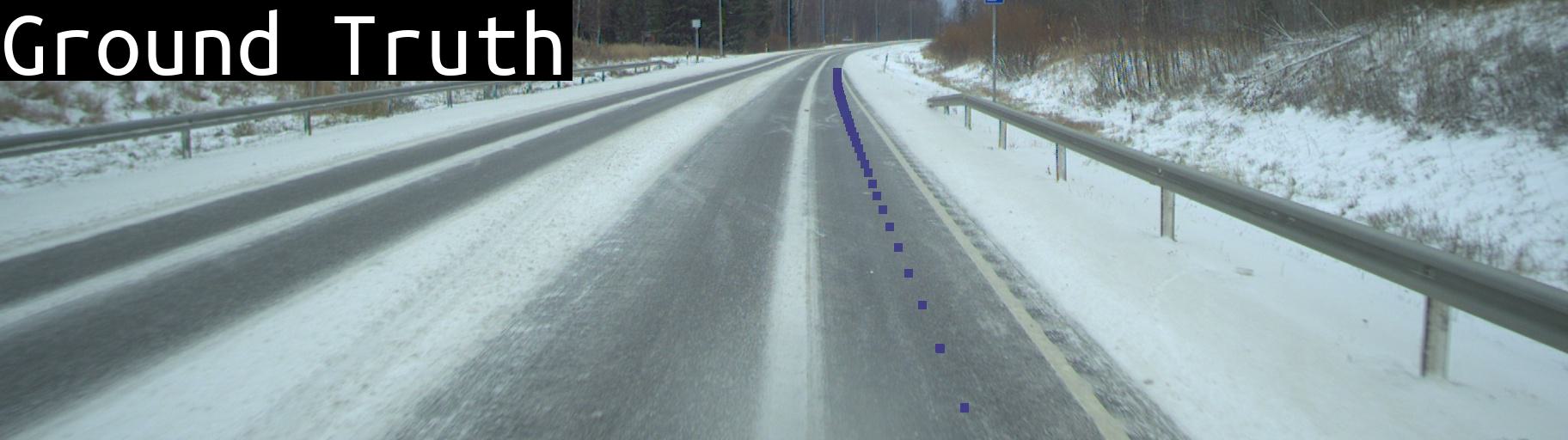} & \includegraphics[width=0.3\textwidth]{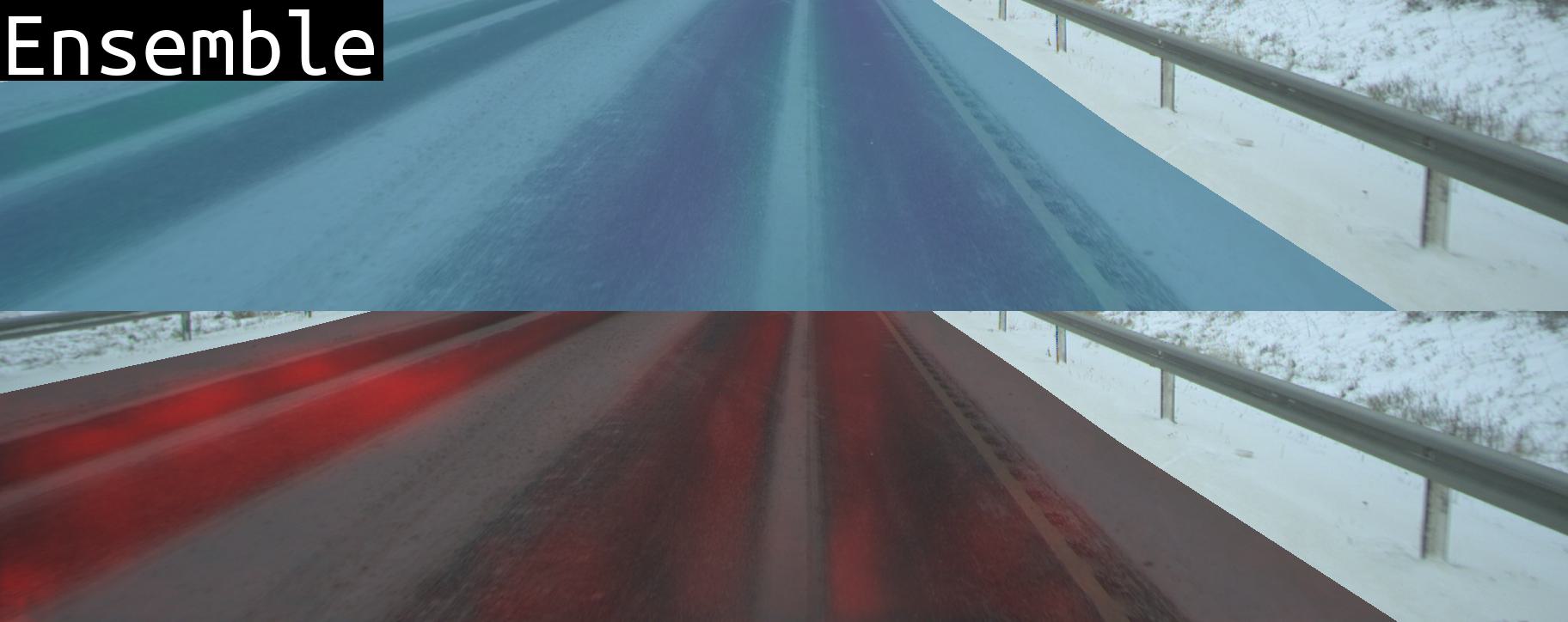} & \includegraphics[width=0.3\textwidth]{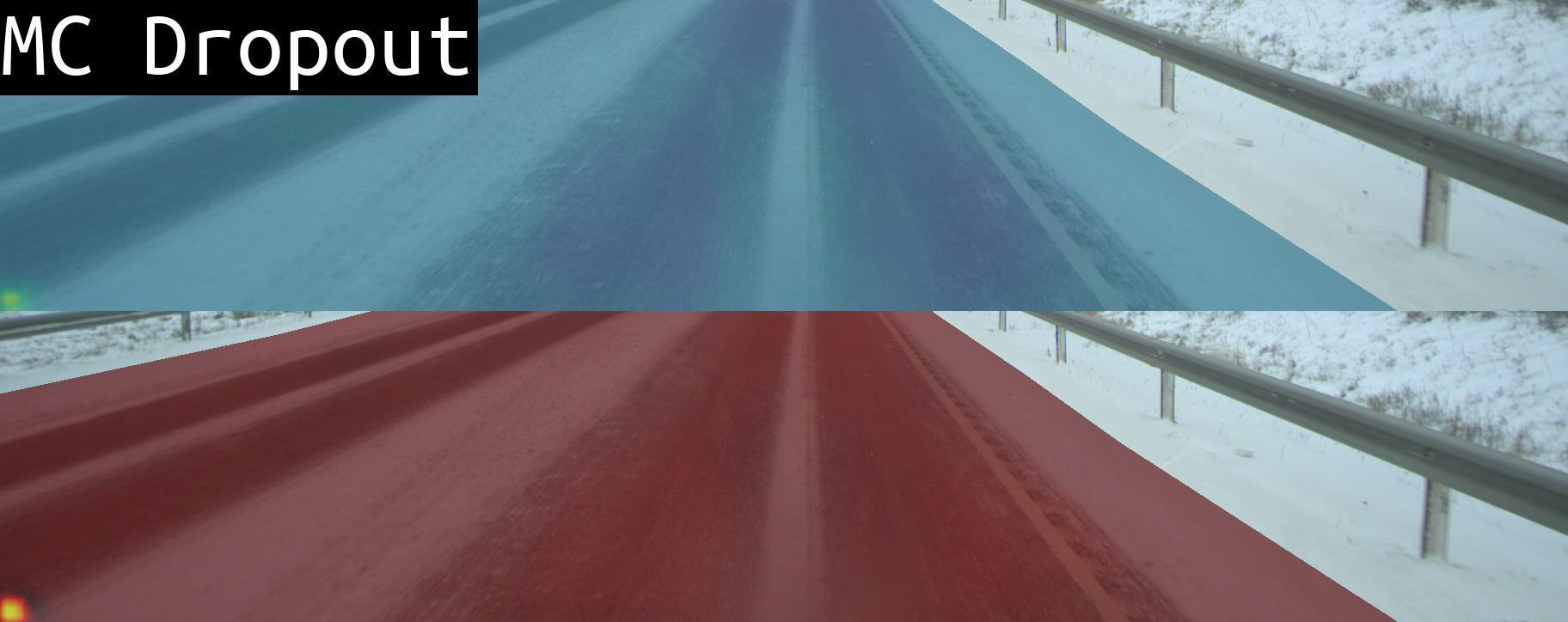} \\
\includegraphics[width=0.3\textwidth]{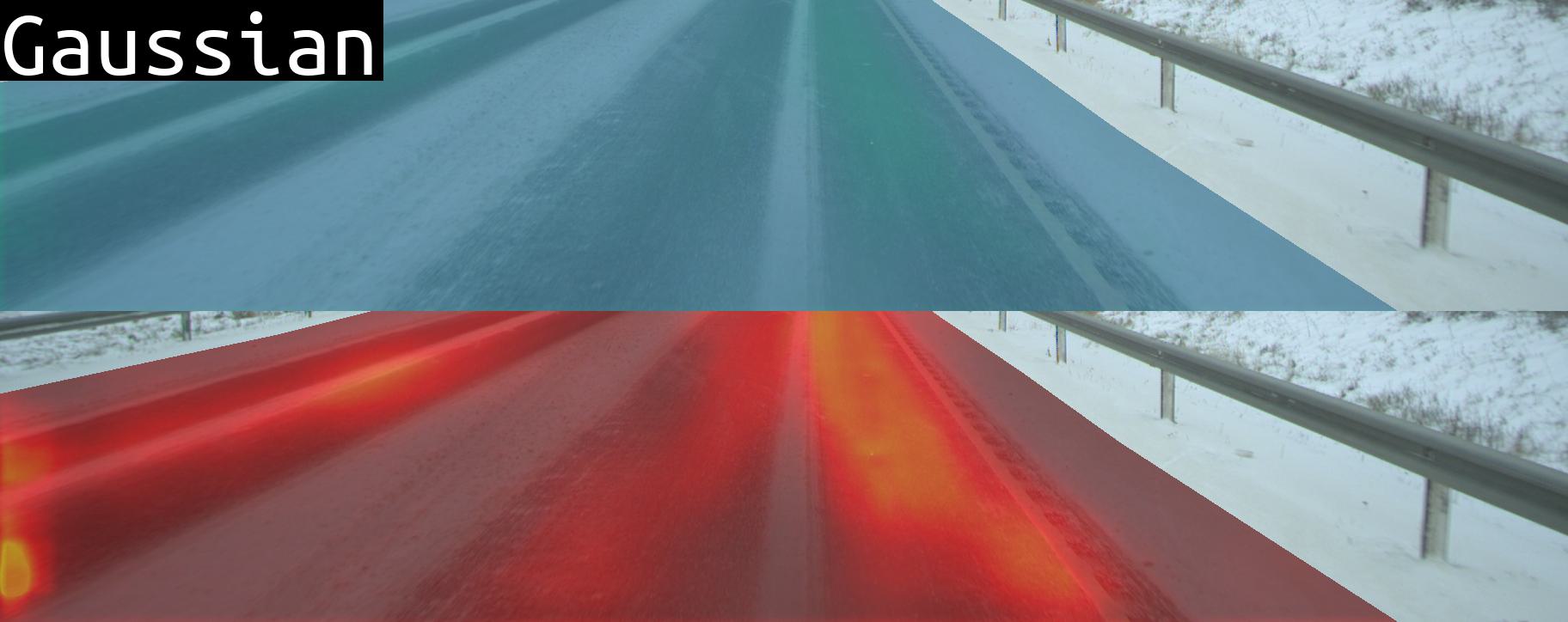} & \includegraphics[width=0.3\textwidth]{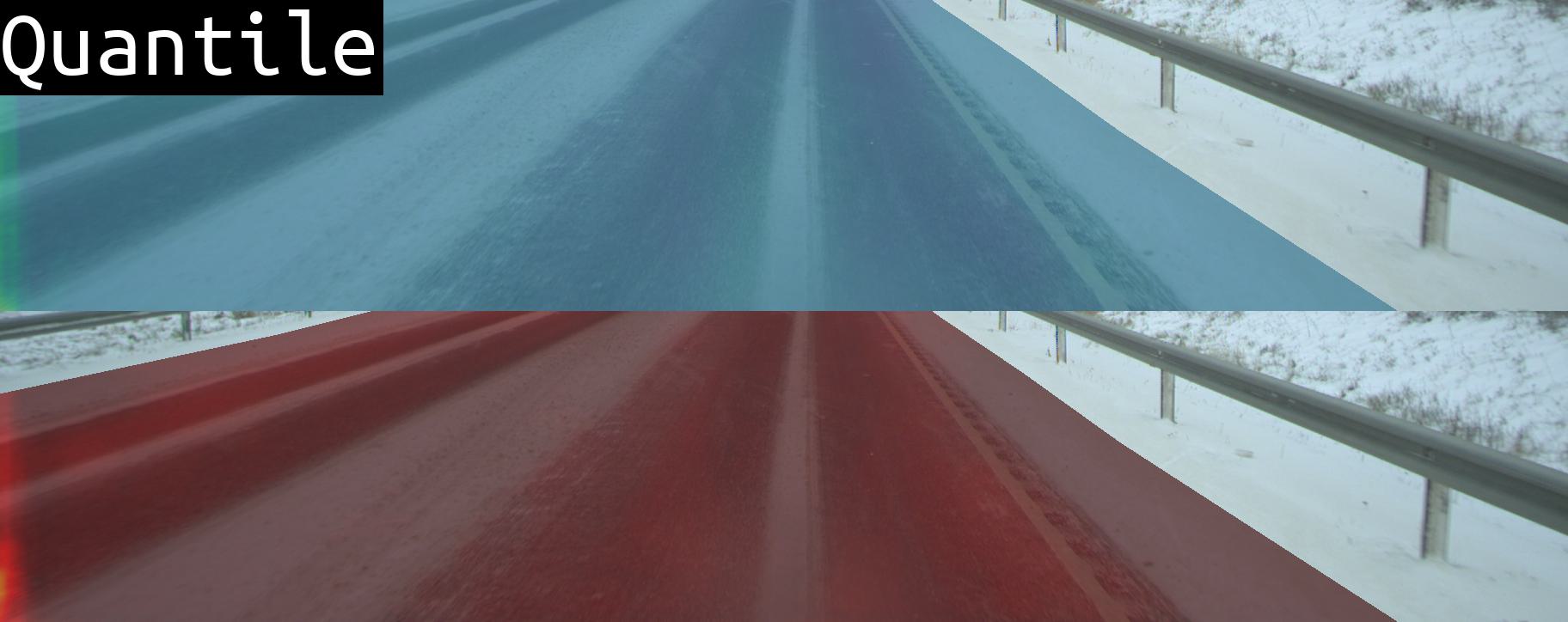} & \includegraphics[width=0.3\textwidth]{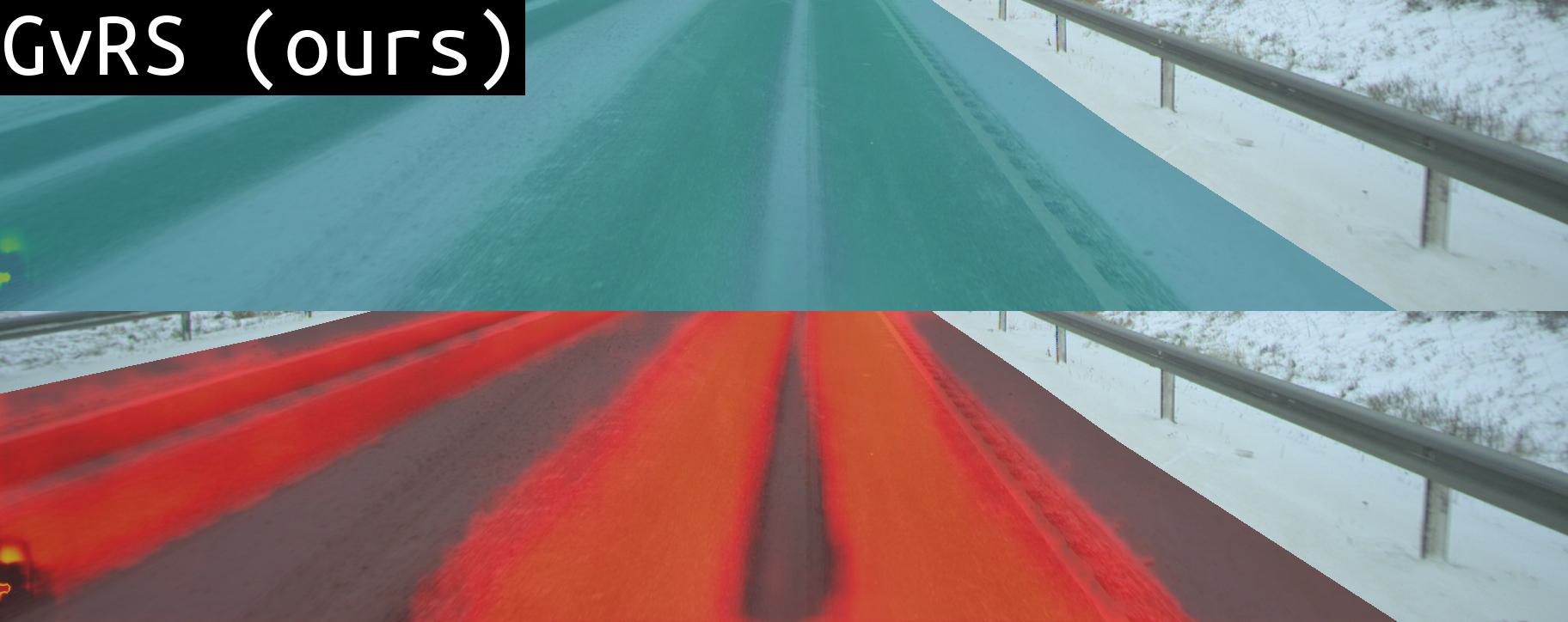} \\ \hline
\includegraphics[width=0.3\textwidth]{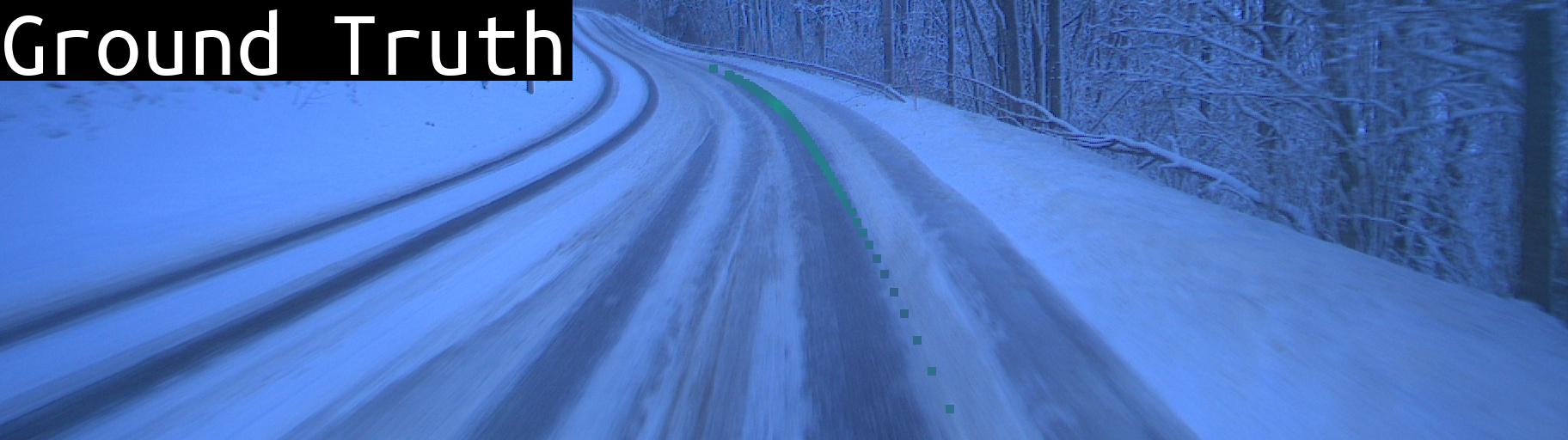} & \includegraphics[width=0.3\textwidth]{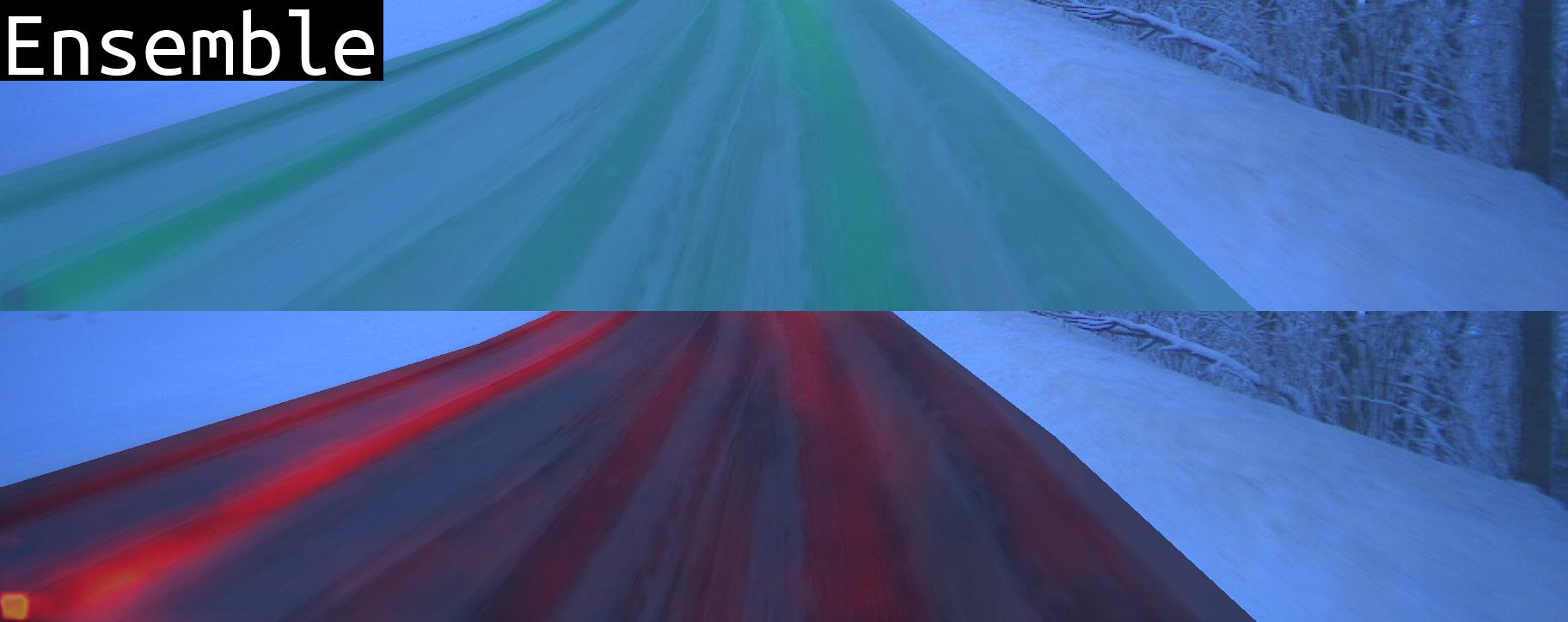} & \includegraphics[width=0.3\textwidth]{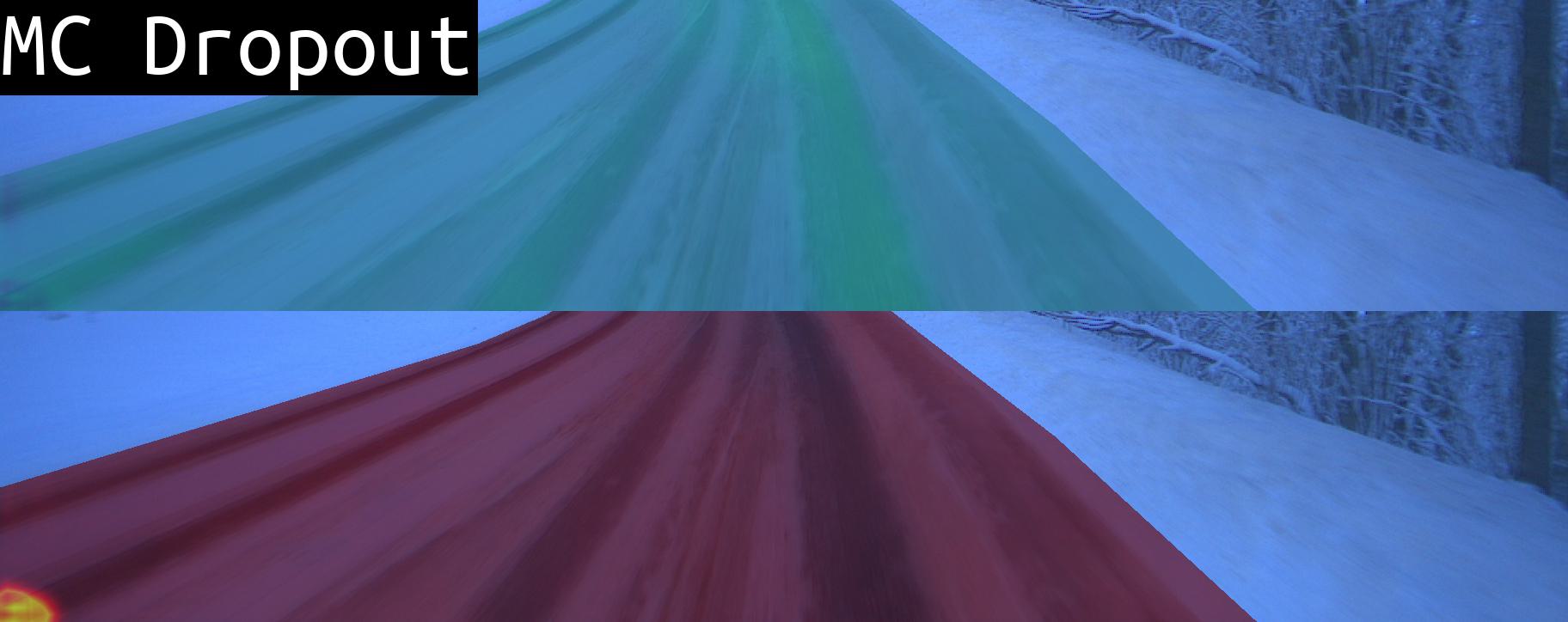} \\
\includegraphics[width=0.3\textwidth]{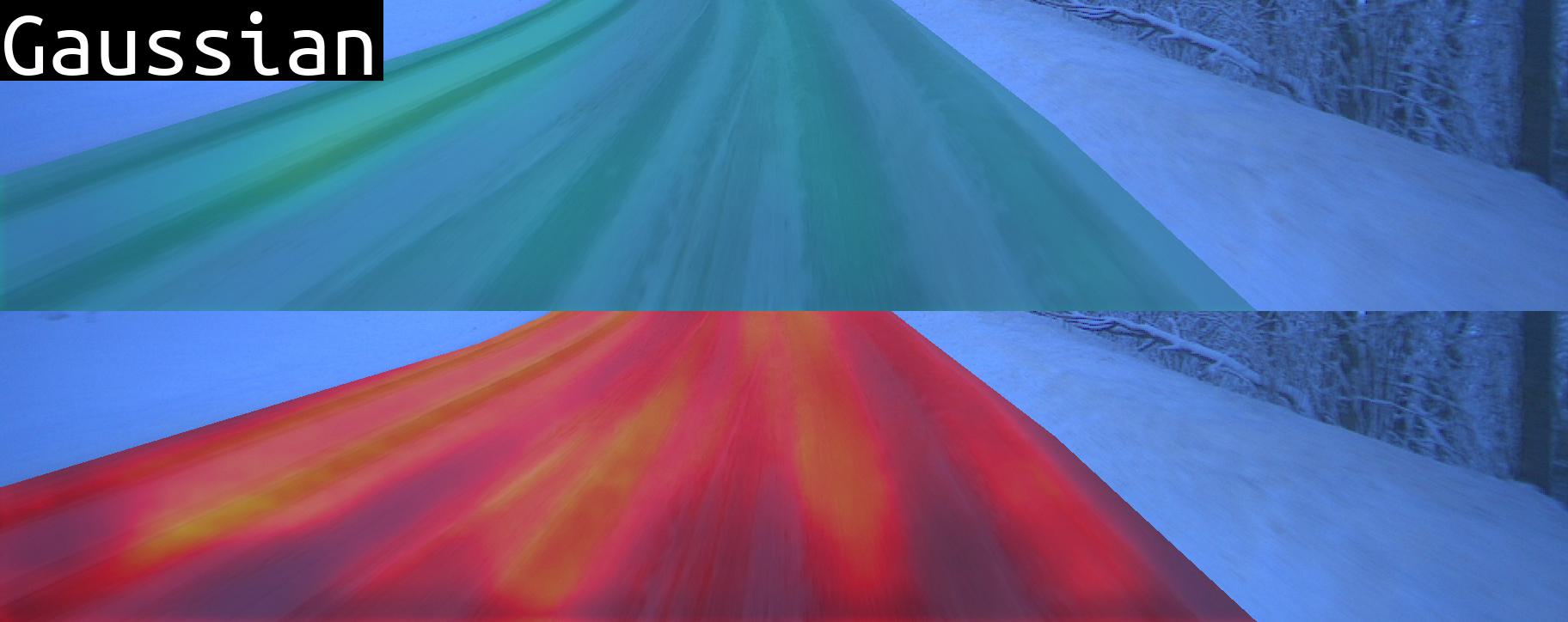} & \includegraphics[width=0.3\textwidth]{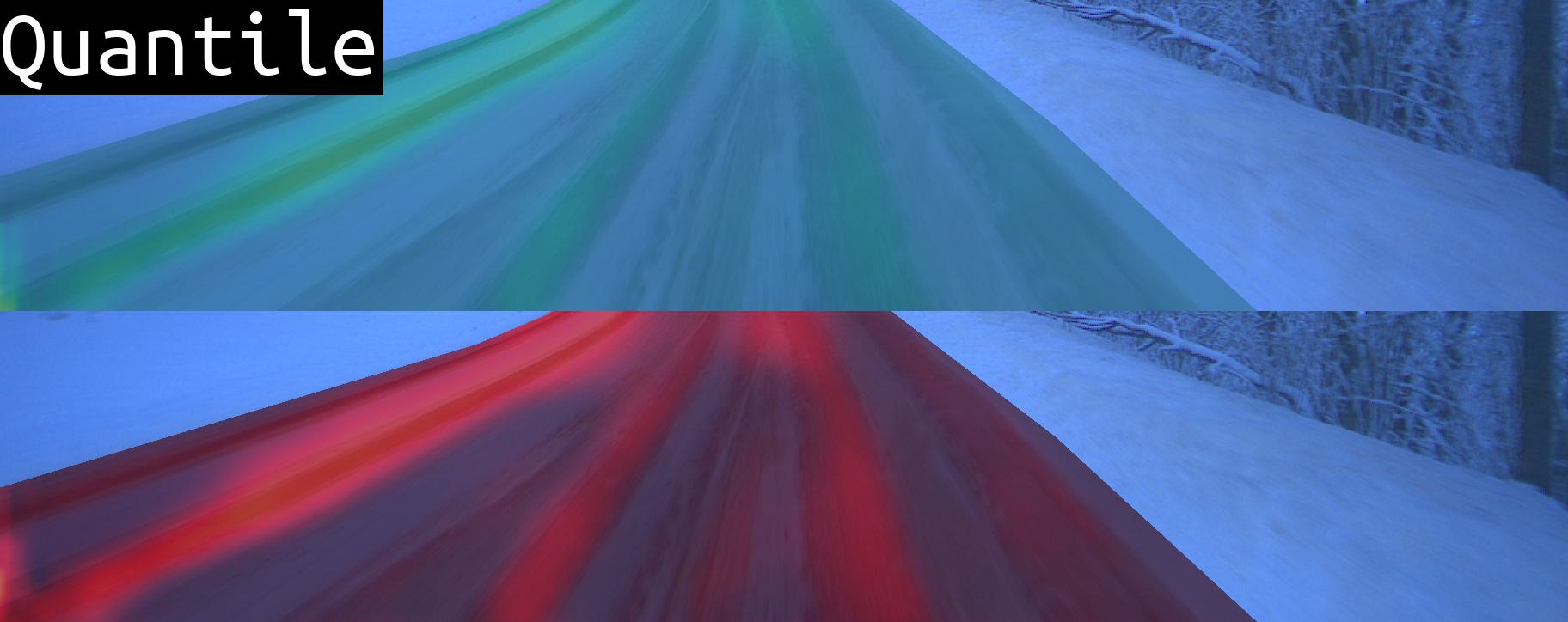} & \includegraphics[width=0.3\textwidth]{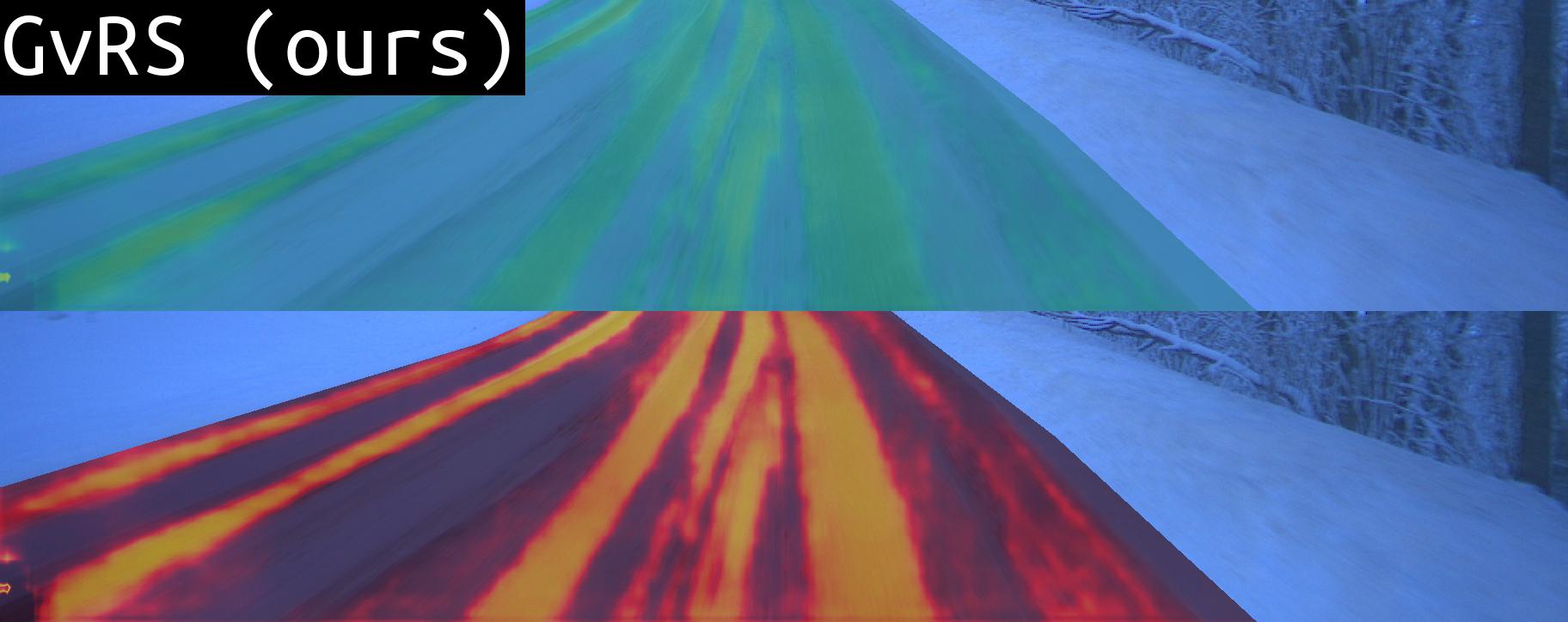} \\ \hline
\includegraphics[width=0.3\textwidth]{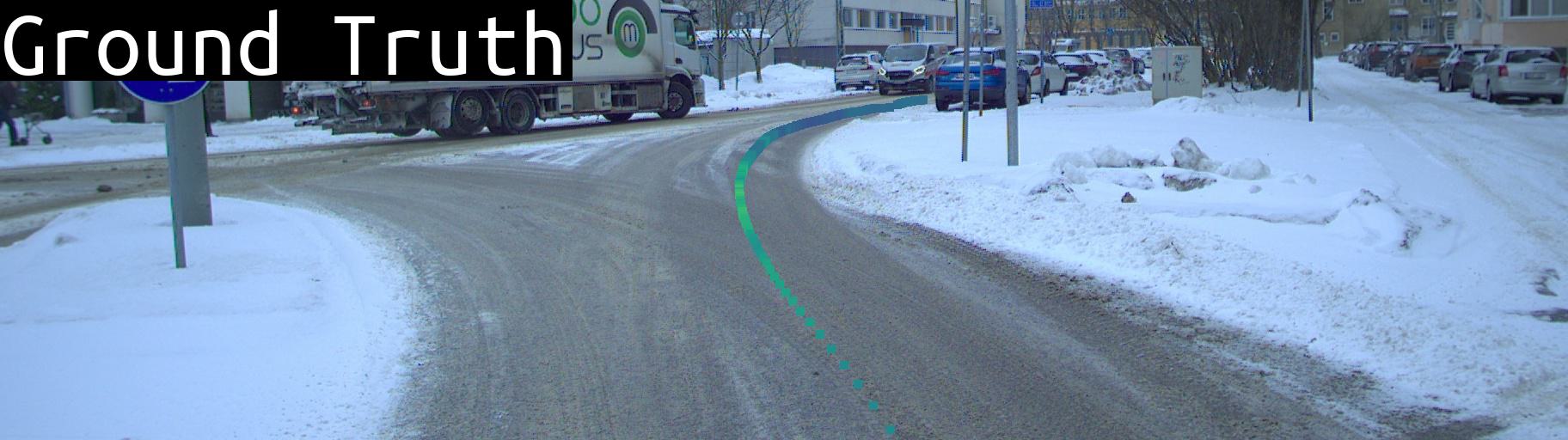} & \includegraphics[width=0.3\textwidth]{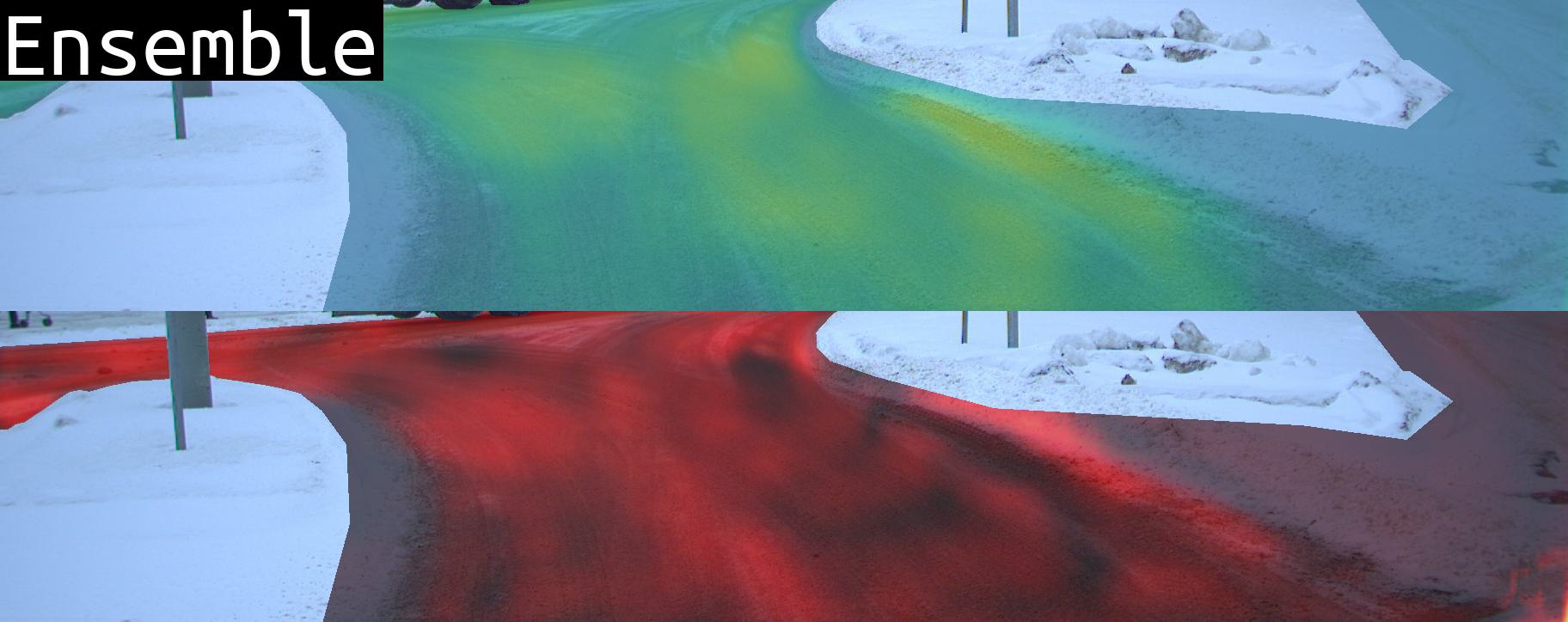} & \includegraphics[width=0.3\textwidth]{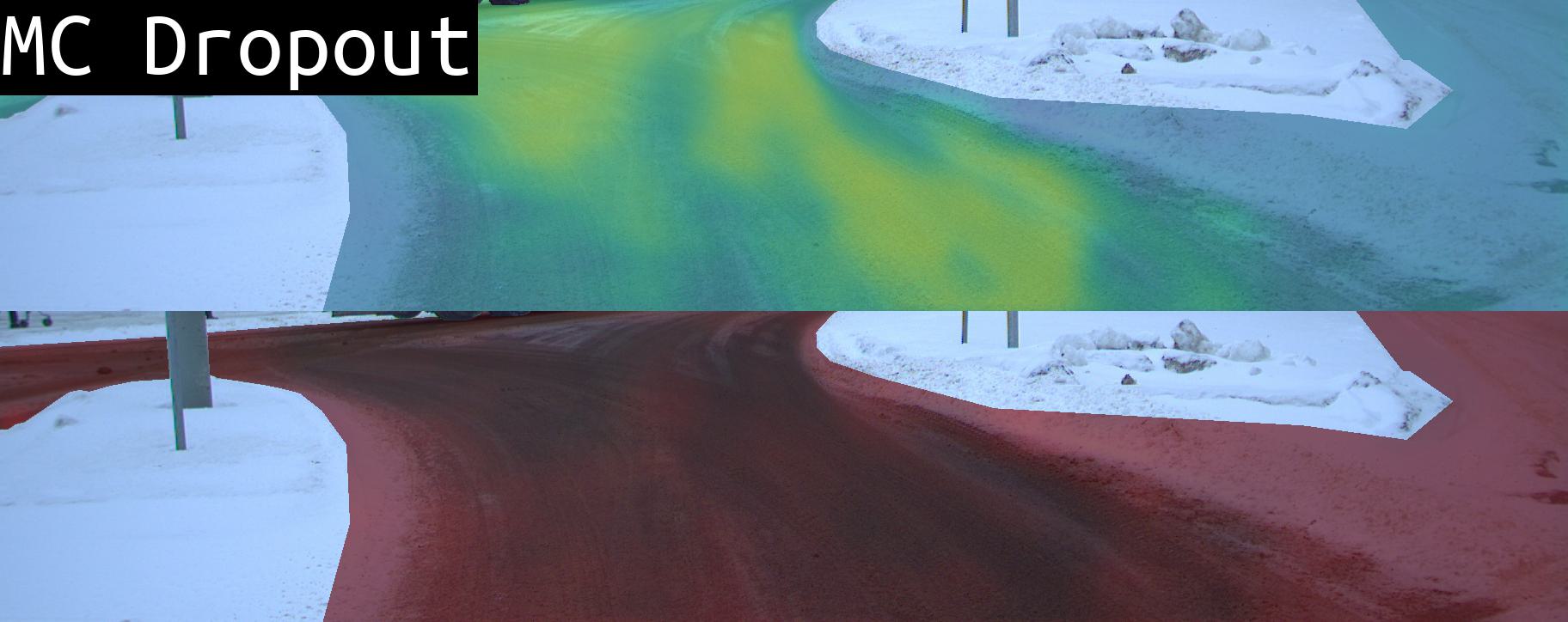} \\
\includegraphics[width=0.3\textwidth]{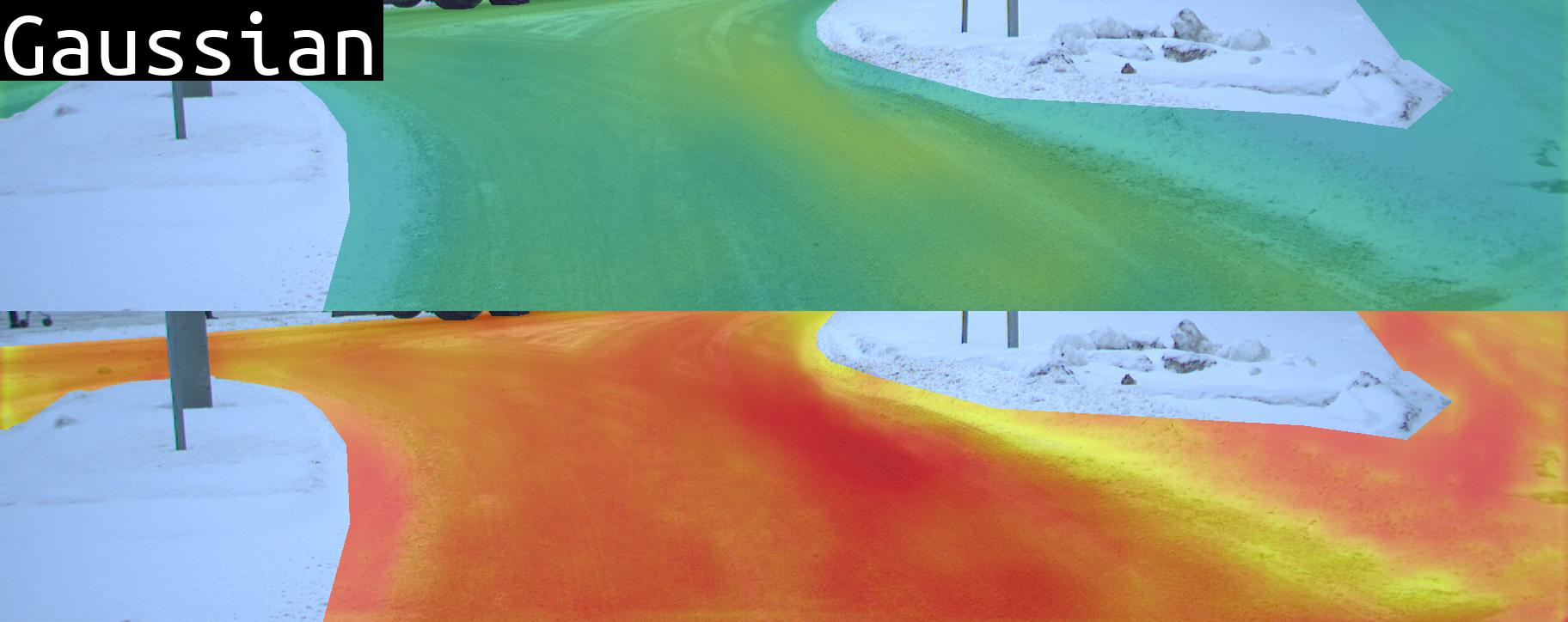} & \includegraphics[width=0.3\textwidth]{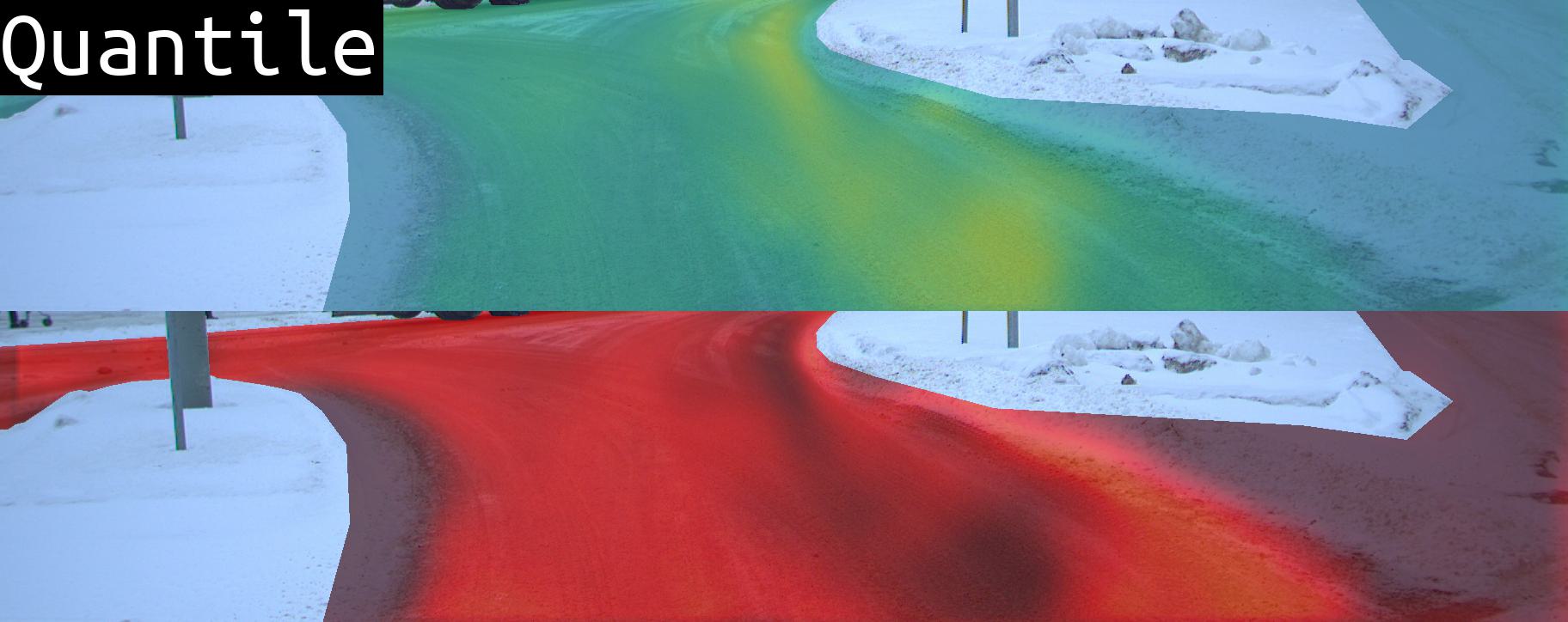} & \includegraphics[width=0.3\textwidth]{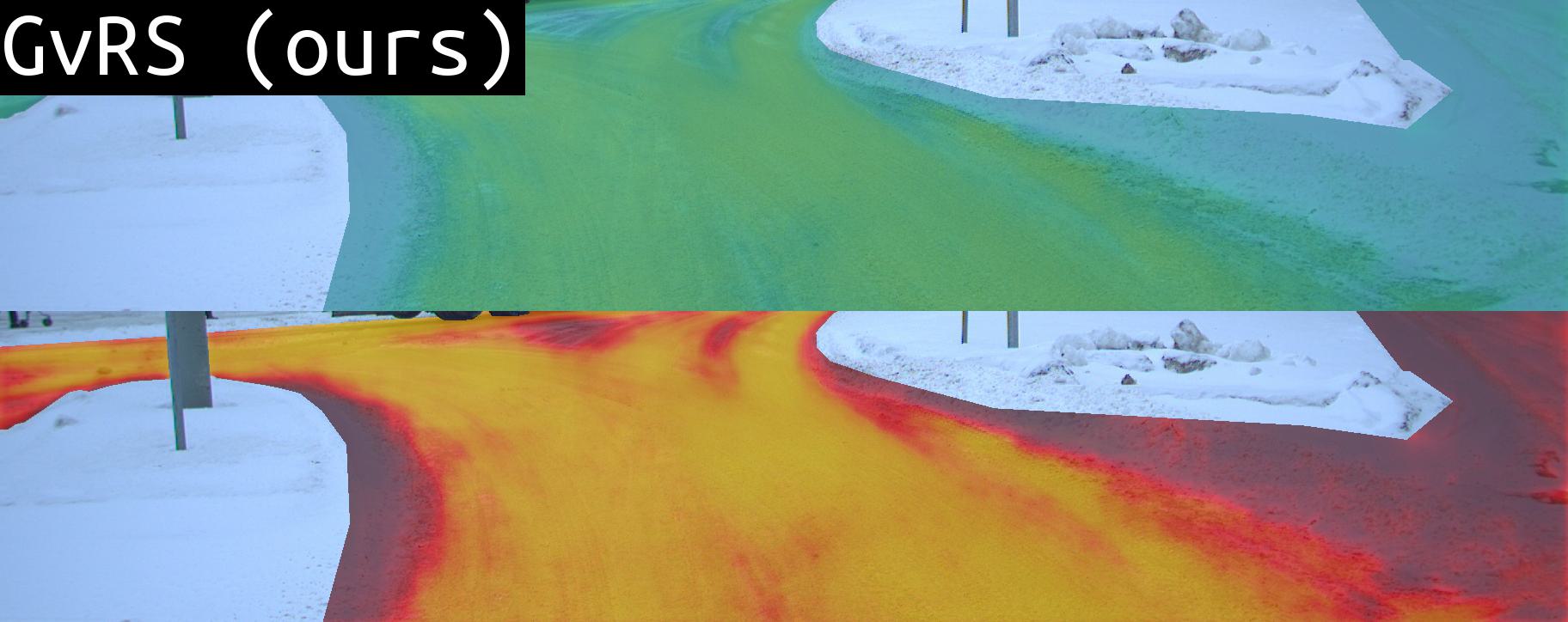} \\ \hline
\includegraphics[width=0.3\textwidth]{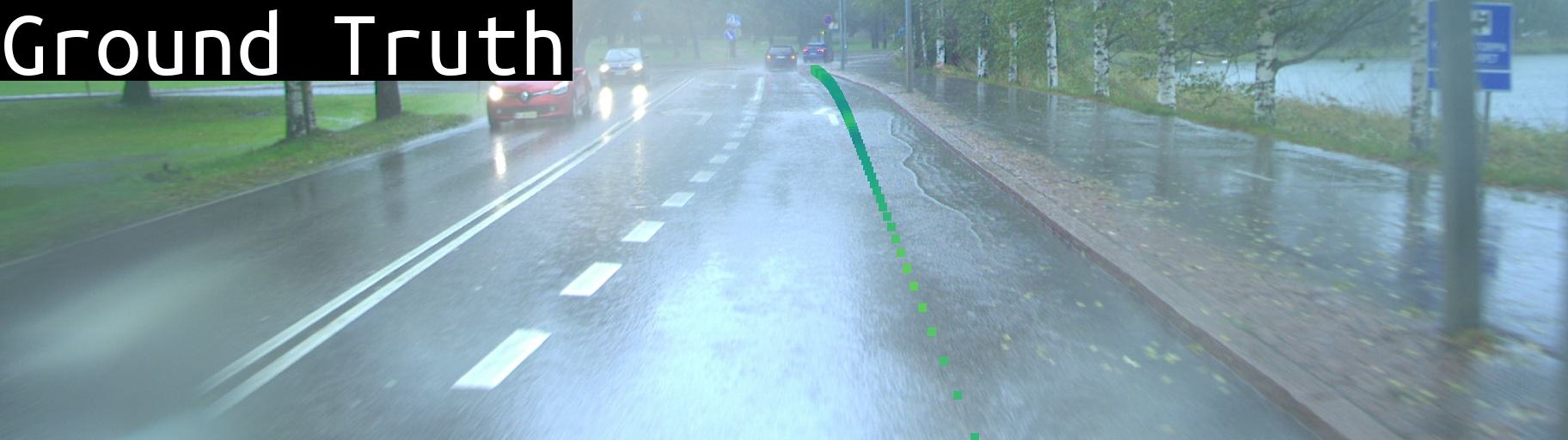} & \includegraphics[width=0.3\textwidth]{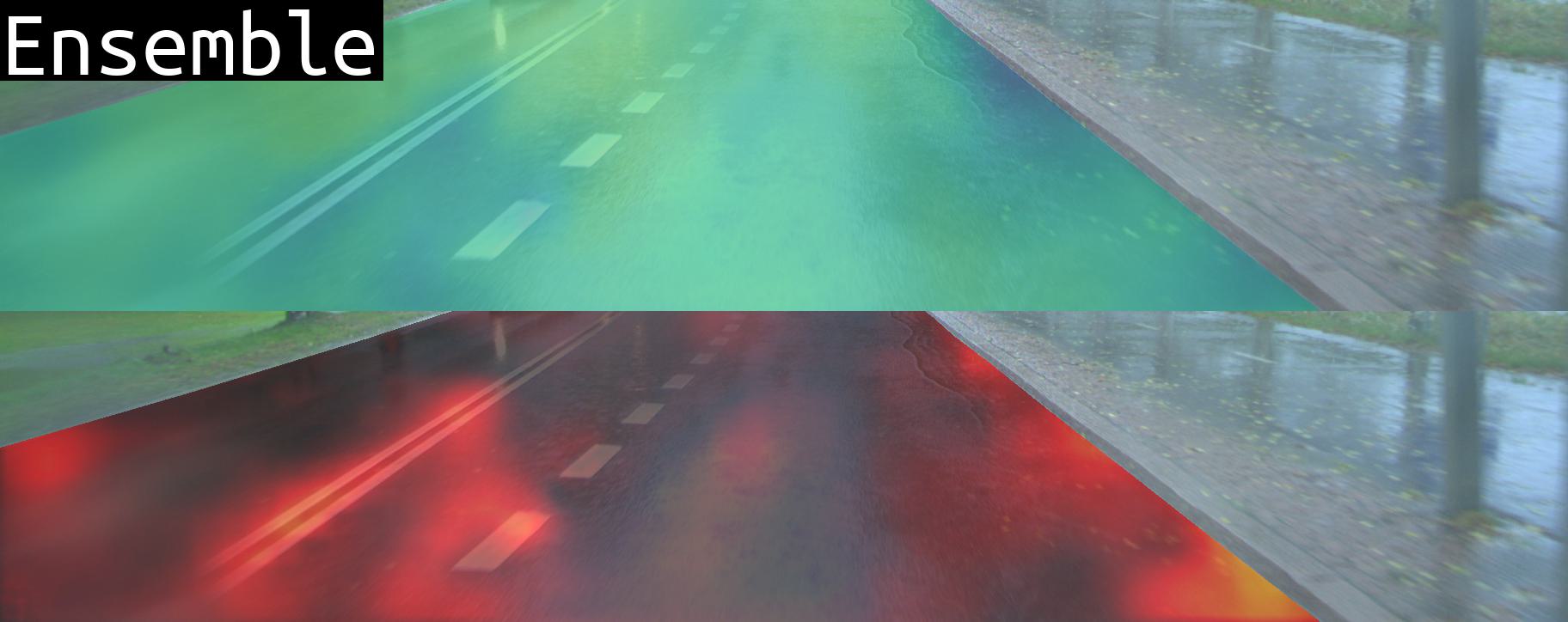} & \includegraphics[width=0.3\textwidth]{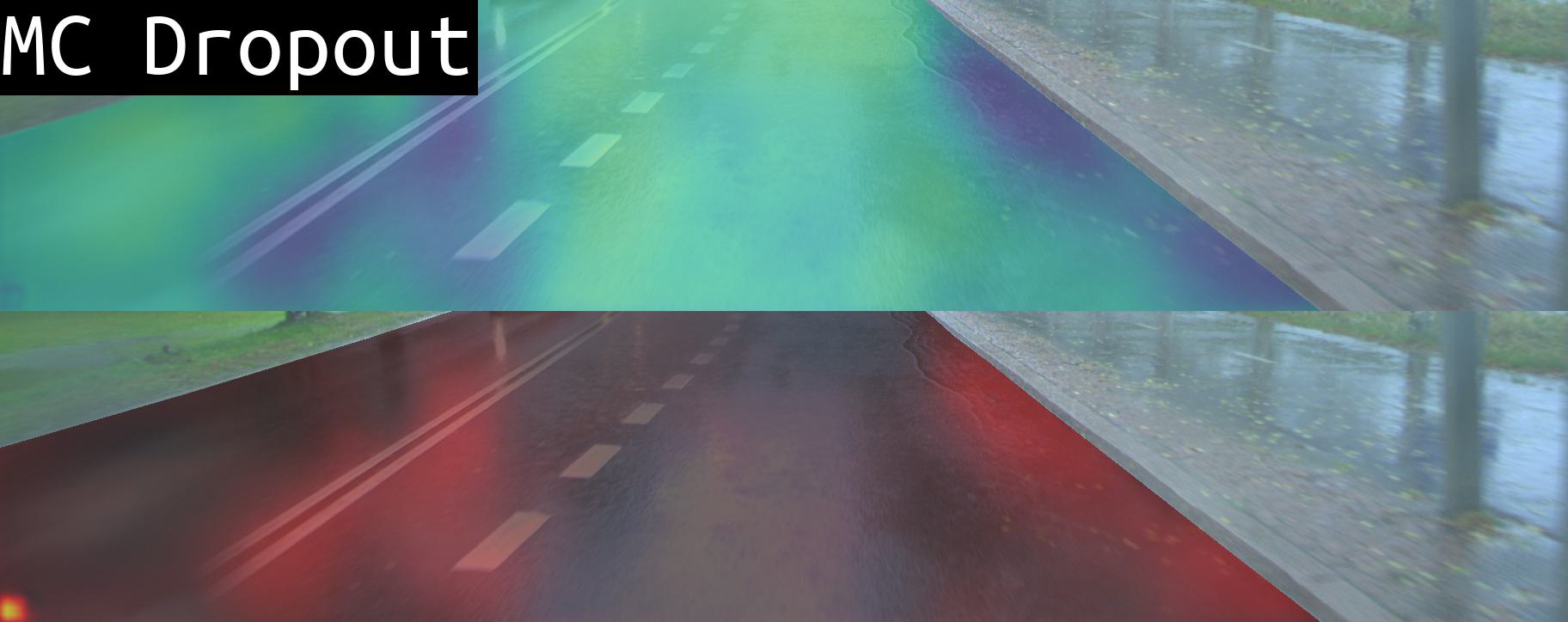} \\
\includegraphics[width=0.3\textwidth]{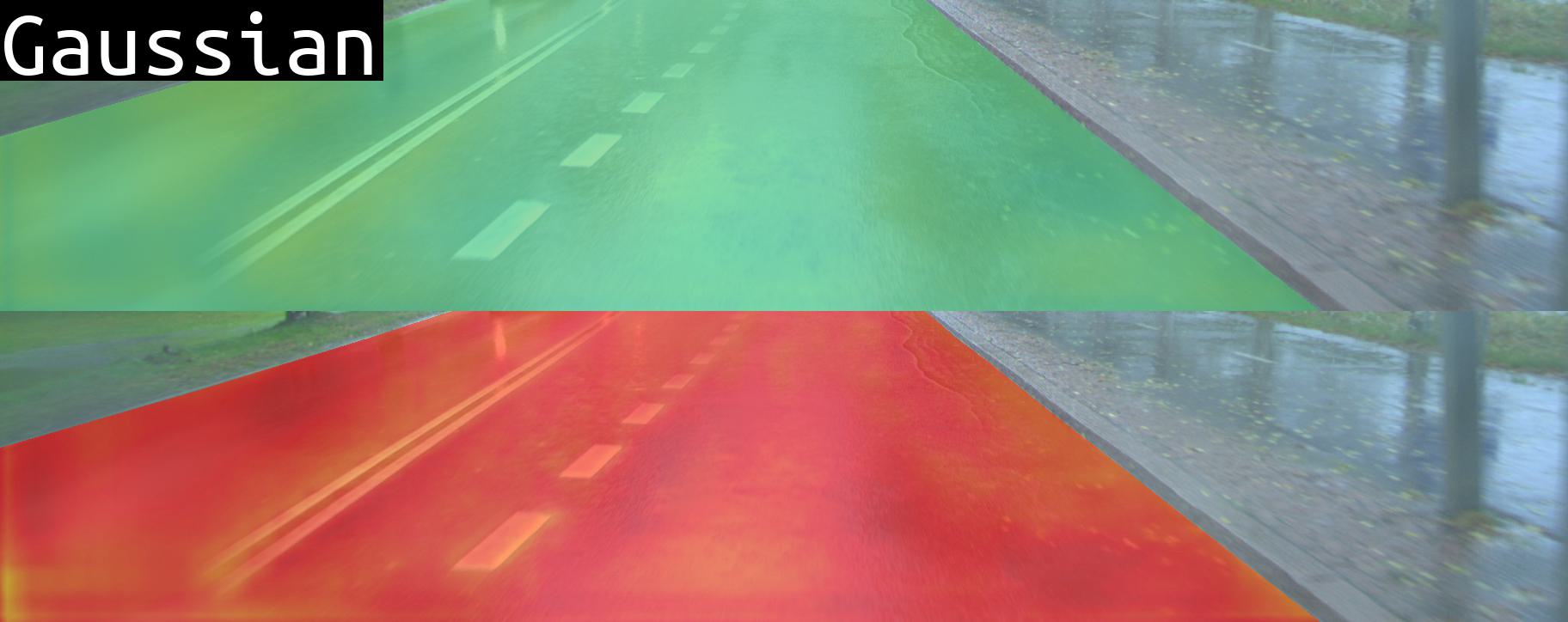} & \includegraphics[width=0.3\textwidth]{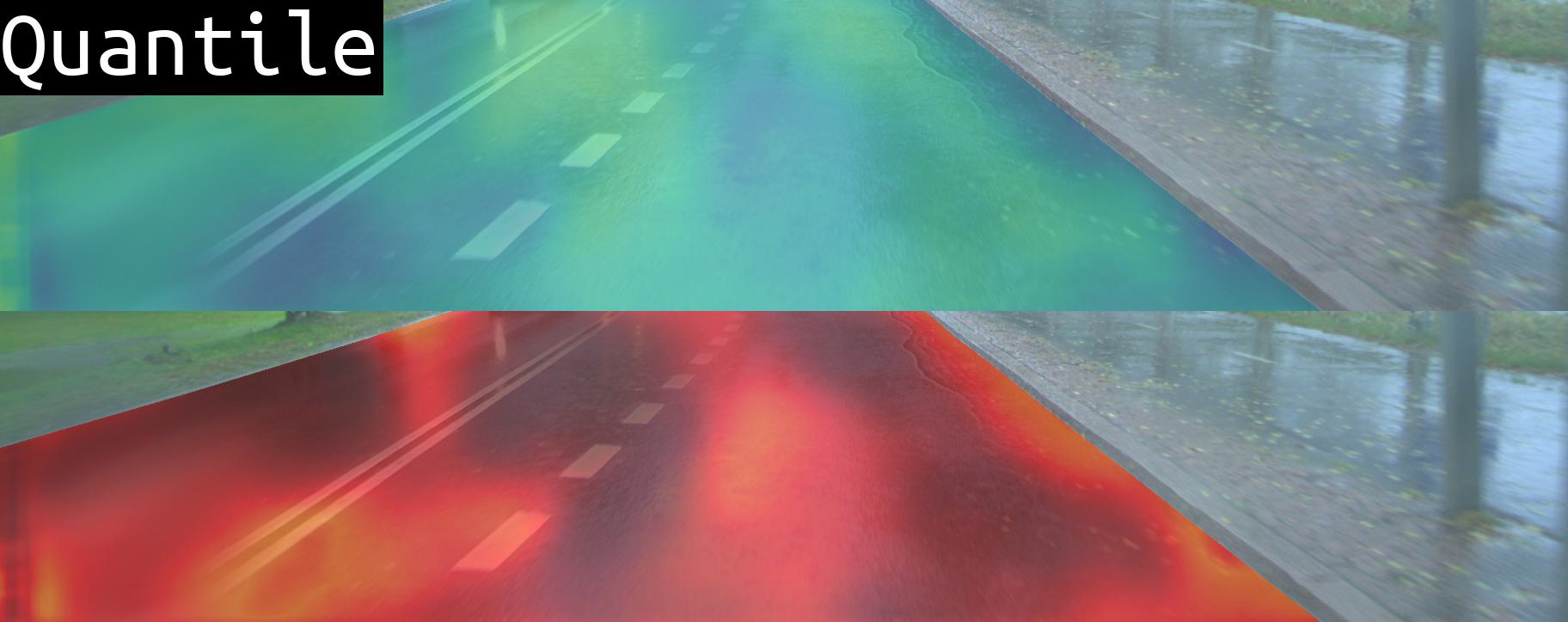} & \includegraphics[width=0.3\textwidth]{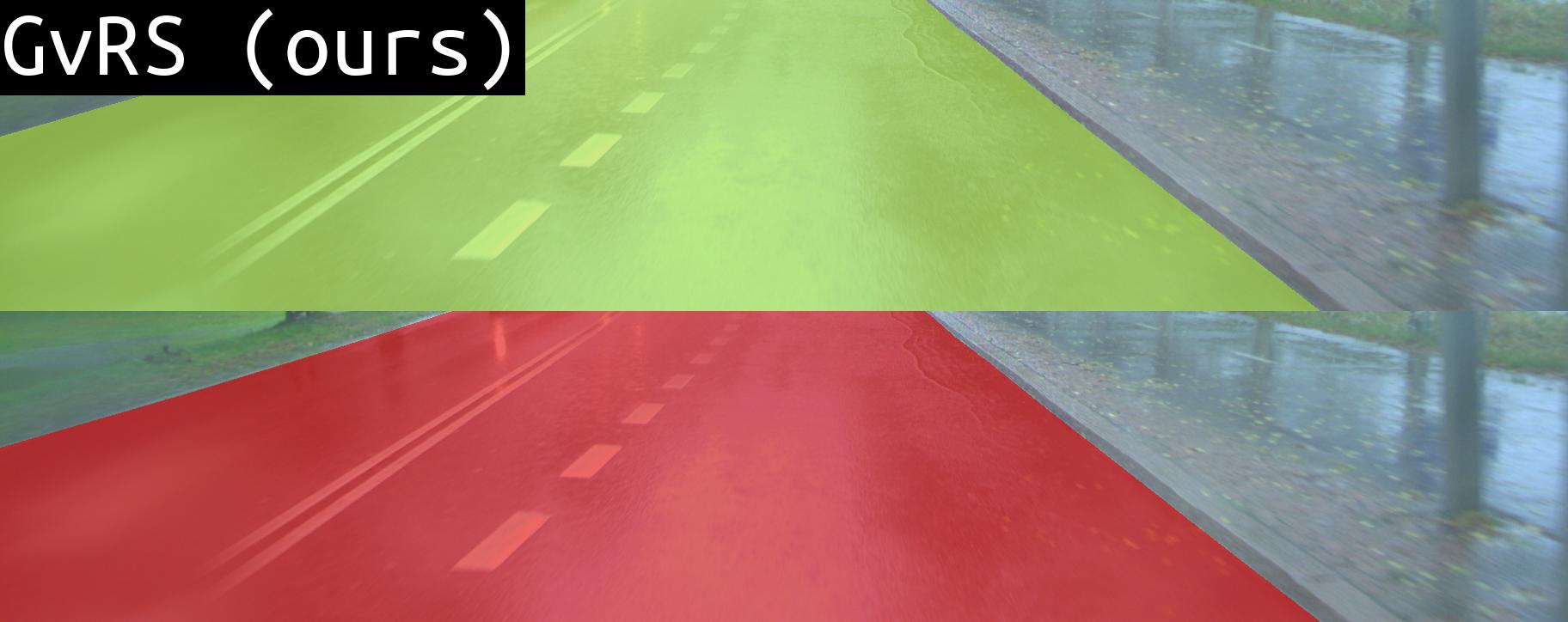} 
\end{tabular}
\includegraphics[width=0.8\textwidth]{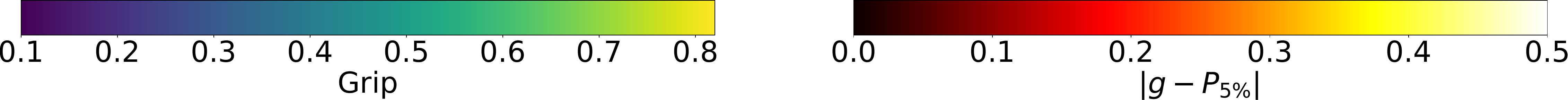}
\caption{Visualizations of grip and grip uncertainty output on the test set images. The first image for each example shows the ground truth data from the road weather sensor. For each model output, the upper image shows the predicted grip distribution mean and the lower image shows the distance between the predicted 5th percentile limit and the predicted grip distribution mean. The road area is manually segmented in the images for clarity.}\label{fig:output_images}
\end{figure}

We visualize the grip predictions along with uncertainty estimates for different models in~\figref{fig:output_images}, focusing on various road surface conditions. In general, the Monte Carlo dropout model often produces highly confident predictions, whereas the Gaussian and GvRS models lead to wider confidence intervals. The GvRS model typically provides sharper edges in its predictions but tends to exhibit excessive uncertainty under certain conditions. Additional examples can be found in the supplementary material. 

\section{Discussion}

Based on our experiments, all tested methods demonstrate relatively strong performance in the grip uncertainty prediction task. However, ensembling and Monte Carlo dropout sampling exhibited lower accuracy in estimating the lower confidence limit of the 90\% confidence interval. Gaussian regression and the proposed GvRS method tended to predict wider confidence intervals, which, in many cases, appeared excessively broad. The quantile regression method achieved the highest accuracy on the test set but performed less reliably in out-of-distribution tests. 

Our GvRS approach demonstrates the most robust performance overall, as it mitigates the uncertainty estimation challenge by leveraging fixed grip probability distributions for each road surface state class. However, it still lacked accuracy in several scenarios, leading to predictions with excessive uncertainty. As observed in~\figref{fig:outlier_scatters}, the lower confidence limits predicted by the GvRS model tend to cluster around specific values due to the fixed grip distributions associated with each road state class. While using models that predict road surface state instead of directly regressing grip with uncertainty estimates can enhance robustness, achieving more accurate predictive uncertainty still requires direct regression on grip values.

Interpreting these results requires considering potential error sources. The reference road weather sensor’s classification of road surface states may contain inaccuracies, and its grip estimates may also introduce errors. However, most key error metrics were close to $0.1$ within the grip scale, which aligns with the expected accuracy of the road weather sensor. This level of precision is sufficient for autonomous driving applications, where grip dynamics between the tires and the road can be affected by multiple other factors.

\section{Conclusion}
We presented a benchmark of several standard regression methods with predictive uncertainty in the novel task of road-area grip uncertainty prediction. Additionally, we introduced a new approach that estimates grip uncertainty by leveraging predicted road surface state probabilities and precomputed grip probability distributions for different surface conditions. Our findings suggest that incorporating road surface state segmentation can enhance the robustness of grip uncertainty prediction: in our out-of-distribution tests, the fraction of ground truth grip values over the 5th percentile predicted by our GvRS model was never below 88\% as with the tested regression models. 
However, achieving higher accuracy still requires direct regression with predictive uncertainty. Future work should explore hybrid approaches that combine these techniques to improve both robustness and accuracy in grip uncertainty estimation.

\section*{Acknowledgements}

Co-funded by the European Union. Views and opinions expressed are however those of the authors only and do not necessarily reflect those of the European Union or European Climate, Infrastructure and Environment Executive Agency (CINEA). Neither the European Union nor the granting authority can be held responsible for them. Project grant no. 101069576.

The contribution of the authors is as follows: Maanp{\"a}{\"a} planned the study, performed experiments, and wrote the manuscript. Hyyti developed the optimization of grip distributions in subsection~\ref{sec:grip_rw_approximation}. Pesonen, Iaroslav and Hyyti advised in the planning of the study and revised the manuscript. Hyypp{\"a} supervised the project.

\clearpage
\newpage


%
%
%
\bibliographystyle{splncs04}
\bibliography{bibliography}

\clearpage
\newpage

\section*{Supplementary material}
\thispagestyle{empty}
\setcounter{equation}{0}
\setcounter{figure}{0}
\setcounter{table}{0}
\setcounter{page}{1}
\makeatletter
\renewcommand{\theequation}{S\arabic{equation}}
\renewcommand{\thefigure}{S\arabic{figure}}

\begin{table}[h]
\caption{Benchmark on different metrics for models on the validation set. The best result among models is bolded for each metric if there is an interpretation for optimal metric value (shown in brackets). This excludes 'Ideal GvRS' as it is only an ideal reference for GvRS model accuracy and not a real model.}
\begin{tabular}{|lrrrrrr|}
\hline
\multicolumn{1}{|l|}{\textbf{Metric}}                                              & \multicolumn{1}{l|}{\textbf{Ensemble}} & \multicolumn{1}{l|}{\textbf{\begin{tabular}[c]{@{}l@{}}MC\\ Dropout\end{tabular}}} & \multicolumn{1}{l|}{\textbf{Gaussian}} & \multicolumn{1}{l|}{\textbf{Quantile}} & \multicolumn{1}{l|}{\textbf{\begin{tabular}[c]{@{}l@{}}GvRS\\ (ours)\end{tabular}}} & \multicolumn{1}{l|}{\begin{tabular}[c]{@{}l@{}}Ideal\\ GvRS\end{tabular}} \\ \hline
\multicolumn{1}{|l|}{$\text{RMSE}(\mu_\text{mean})$ ($\downarrow$)}                          & \multicolumn{1}{r|}{\textbf{0.0631}}   & \multicolumn{1}{r|}{0.0673}                                                        & \multicolumn{1}{r|}{0.0754}            & \multicolumn{1}{r|}{0.0659}            & \multicolumn{1}{r|}{0.0940}                                                         & 0.0874                                                                    \\ \cline{1-1}
\multicolumn{1}{|l|}{$\text{RMSE}(\mu_\text{median})$ ($\downarrow$)}                        & \multicolumn{1}{r|}{\textbf{0.0631}}   & \multicolumn{1}{r|}{0.0673}                                                        & \multicolumn{1}{r|}{0.0754}            & \multicolumn{1}{r|}{-}                 & \multicolumn{1}{r|}{0.1024}                                                         & 0.0933                                                                    \\ \cline{1-1}
\multicolumn{1}{|l|}{$F(g\in L_\sigma)$ ($\rightarrow$ 68.3)[$\%$]}    & \multicolumn{1}{r|}{\textbf{68.0}}     & \multicolumn{1}{r|}{71.8}                                                          & \multicolumn{1}{r|}{90.1}              & \multicolumn{1}{r|}{-}                 & \multicolumn{1}{r|}{86.6}                                                           & 84.5                                                                      \\ \cline{1-1}
\multicolumn{1}{|l|}{$F(g\in L_\text{90\%})$ ($\rightarrow$ 90)[$\%$]} & \multicolumn{1}{r|}{78.8}              & \multicolumn{1}{r|}{78.2}                                                          & \multicolumn{1}{r|}{95.8}              & \multicolumn{1}{r|}{\textbf{89.5}}     & \multicolumn{1}{r|}{95.3}                                                           & 94.4                                                                      \\ \cline{1-1}
\multicolumn{1}{|l|}{$F(g >P_\text{5\%})$ ($\rightarrow$ 95)[$\%$]}    & \multicolumn{1}{r|}{89.6}              & \multicolumn{1}{r|}{89.0}                                                          & \multicolumn{1}{r|}{96.8}              & \multicolumn{1}{r|}{93.3}              & \multicolumn{1}{r|}{\textbf{96.2}}                                                  & 96.4                                                                      \\ \cline{1-1}
\multicolumn{1}{|l|}{$\mu(P_\text{95\%}-P_\text{5\%})$}                            & \multicolumn{1}{r|}{0.0412}            & \multicolumn{1}{r|}{0.0603}                                                        & \multicolumn{1}{r|}{0.1388}            & \multicolumn{1}{r|}{0.0773}            & \multicolumn{1}{r|}{0.2080}                                                         & 0.1640                                                                    \\ \cline{1-1}
\multicolumn{1}{|l|}{$\mu(P_\text{5\%})$}                                          & \multicolumn{1}{r|}{0.643}             & \multicolumn{1}{r|}{0.630}                                                         & \multicolumn{1}{r|}{0.588}             & \multicolumn{1}{r|}{0.622}             & \multicolumn{1}{r|}{0.563}                                                          & 0.589                                                                     \\ \hline
\multicolumn{7}{|l|}{\textbf{For $g<P_\text{5\%}$:}}                                                                                                                                                                                                                                                                                                                                                                                                                 \\ \hline
\multicolumn{1}{|l|}{$p_\text{50\%}(P_\text{5\%} - g)$ ($\downarrow$)}             & \multicolumn{1}{r|}{0.0252}            & \multicolumn{1}{r|}{0.0408}                                                        & \multicolumn{1}{r|}{0.0406}            & \multicolumn{1}{r|}{0.0266}            & \multicolumn{1}{r|}{\textbf{0.0219}}                                                & 0.0141                                                                    \\ \cline{1-1}
\multicolumn{1}{|l|}{$p_\text{70\%}(P_\text{5\%} - g)$ ($\downarrow$)}             & \multicolumn{1}{r|}{0.0607}            & \multicolumn{1}{r|}{0.0889}                                                        & \multicolumn{1}{r|}{0.0763}            & \multicolumn{1}{r|}{\textbf{0.0545}}   & \multicolumn{1}{r|}{0.0575}                                                         & 0.0306                                                                    \\ \cline{1-1}
\multicolumn{1}{|l|}{$p_\text{90\%}(P_\text{5\%} - g)$ ($\downarrow$)}             & \multicolumn{1}{r|}{0.1528}            & \multicolumn{1}{r|}{0.1905}                                                        & \multicolumn{1}{r|}{0.1790}            & \multicolumn{1}{r|}{0.1501}            & \multicolumn{1}{r|}{\textbf{0.1459}}                                                & 0.1041                                                                    \\ \hline
\end{tabular}
\end{table}

\begin{table}[]
\caption{Benchmark on model runtimes for a single image. The runtime was tested for 1000 loops on a single Nvidia RTX A6000 GPU.}
\begin{tabular}{|l|l|l|l|l|l|}
\hline
\textbf{}                 & \textbf{Ensemble}         & \textbf{\begin{tabular}[c]{@{}l@{}}MC\\ Dropout\end{tabular}} & \textbf{Gaussian}         & \textbf{Quantile}         & \textbf{\begin{tabular}[c]{@{}l@{}}GvRS\\ (ours)\end{tabular}} \\ \hline
\textbf{Runtime} {[}ms{]} & \multicolumn{1}{r|}{74.1} & \multicolumn{1}{r|}{17.9}                                     & \multicolumn{1}{r|}{14.7} & \multicolumn{1}{r|}{14.7} & \multicolumn{1}{r|}{32.3}                                      \\ \hline
\end{tabular}
\end{table}

\begin{figure}[h]
\centering
\begin{tabular}{ccc}
\includegraphics[width=0.3\textwidth]{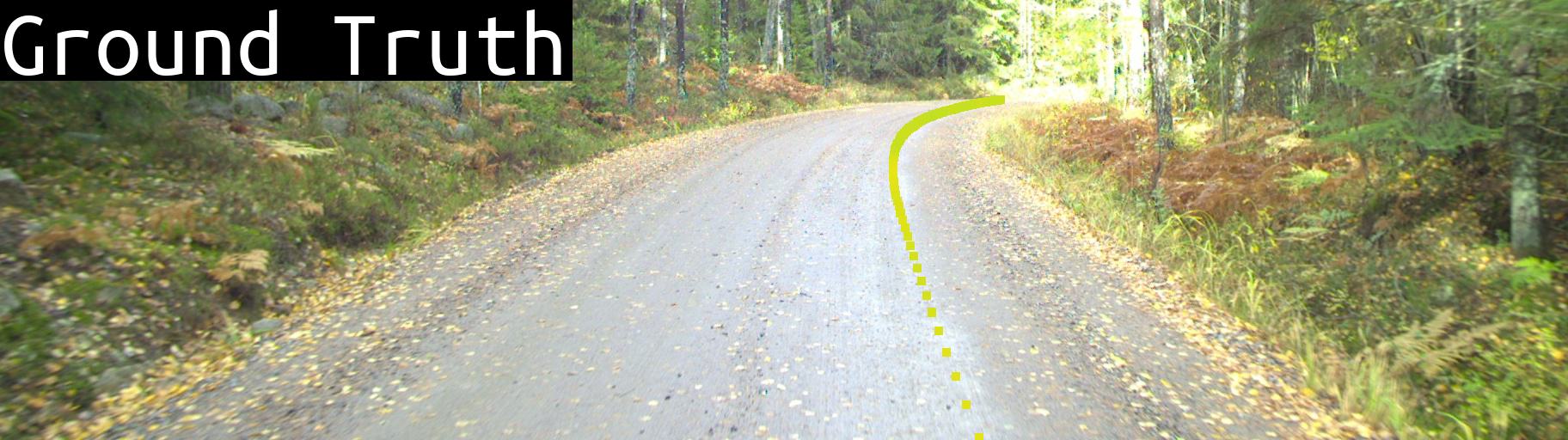} & \includegraphics[width=0.3\textwidth]{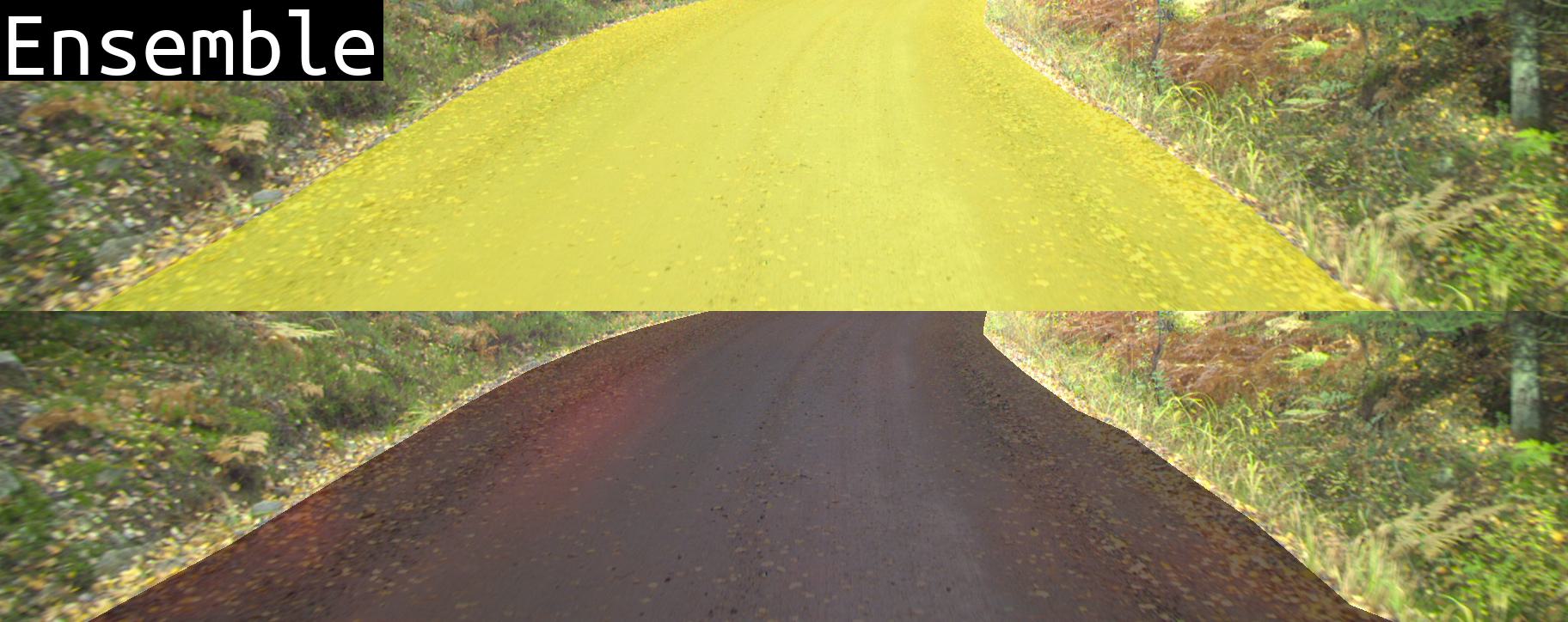} & \includegraphics[width=0.3\textwidth]{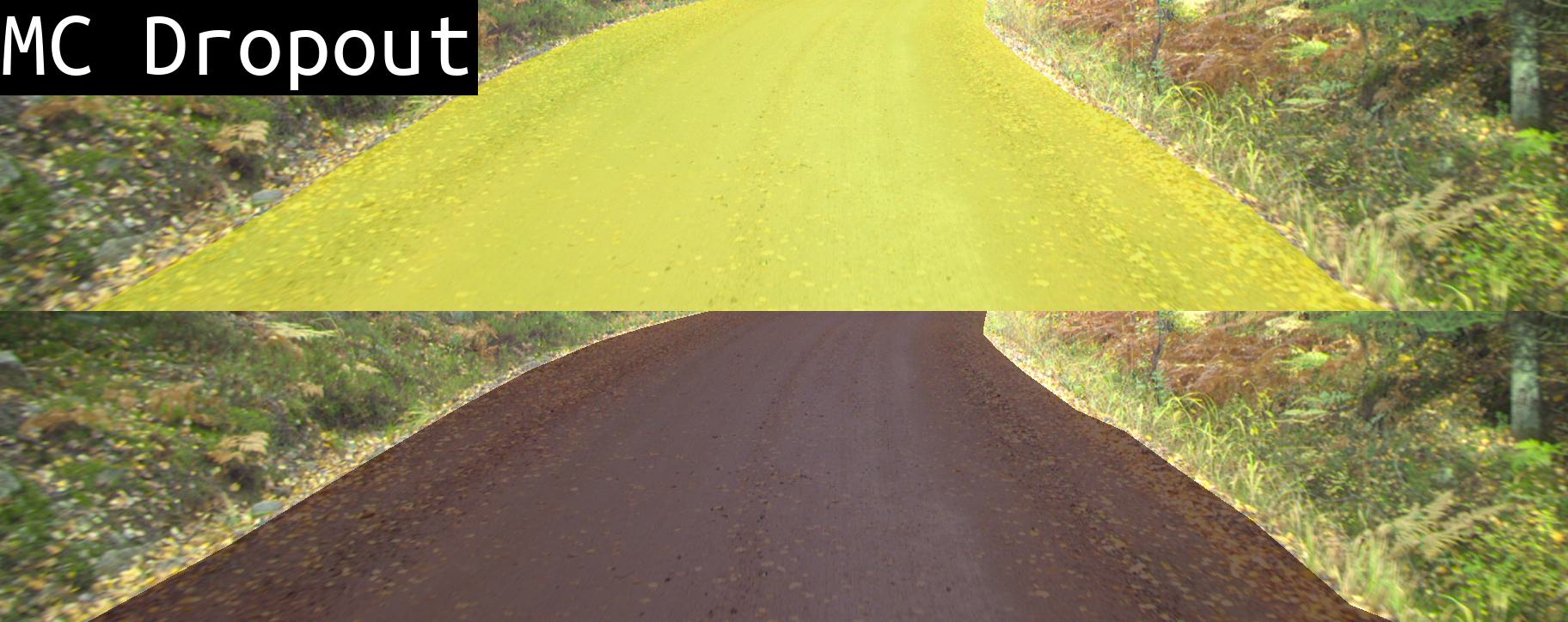} \\
\includegraphics[width=0.3\textwidth]{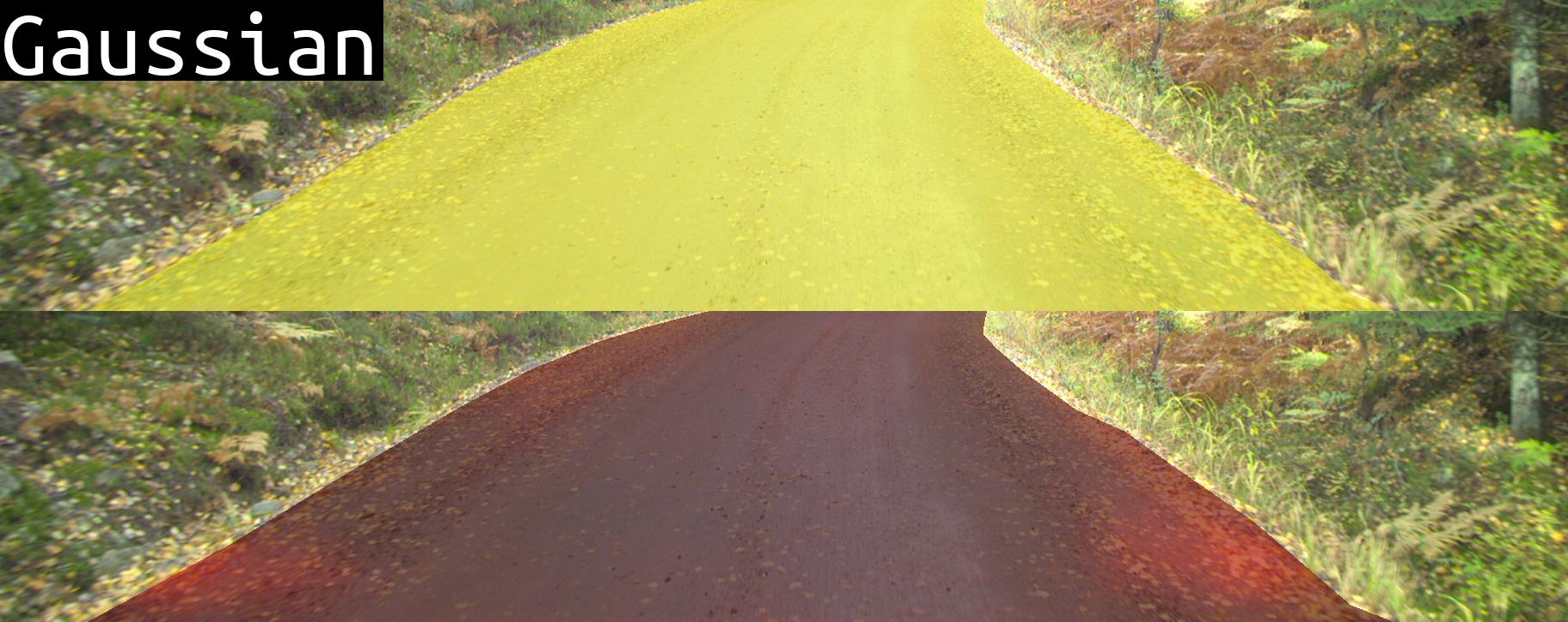} & \includegraphics[width=0.3\textwidth]{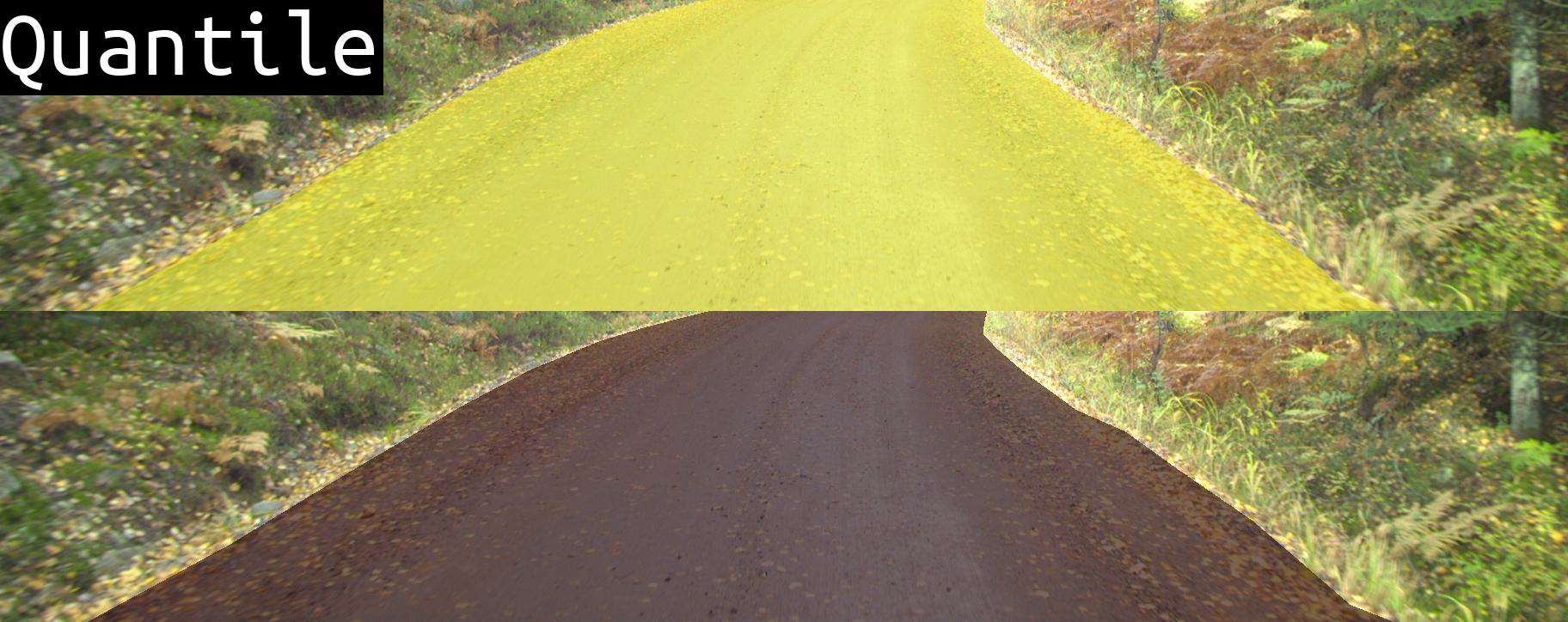} & \includegraphics[width=0.3\textwidth]{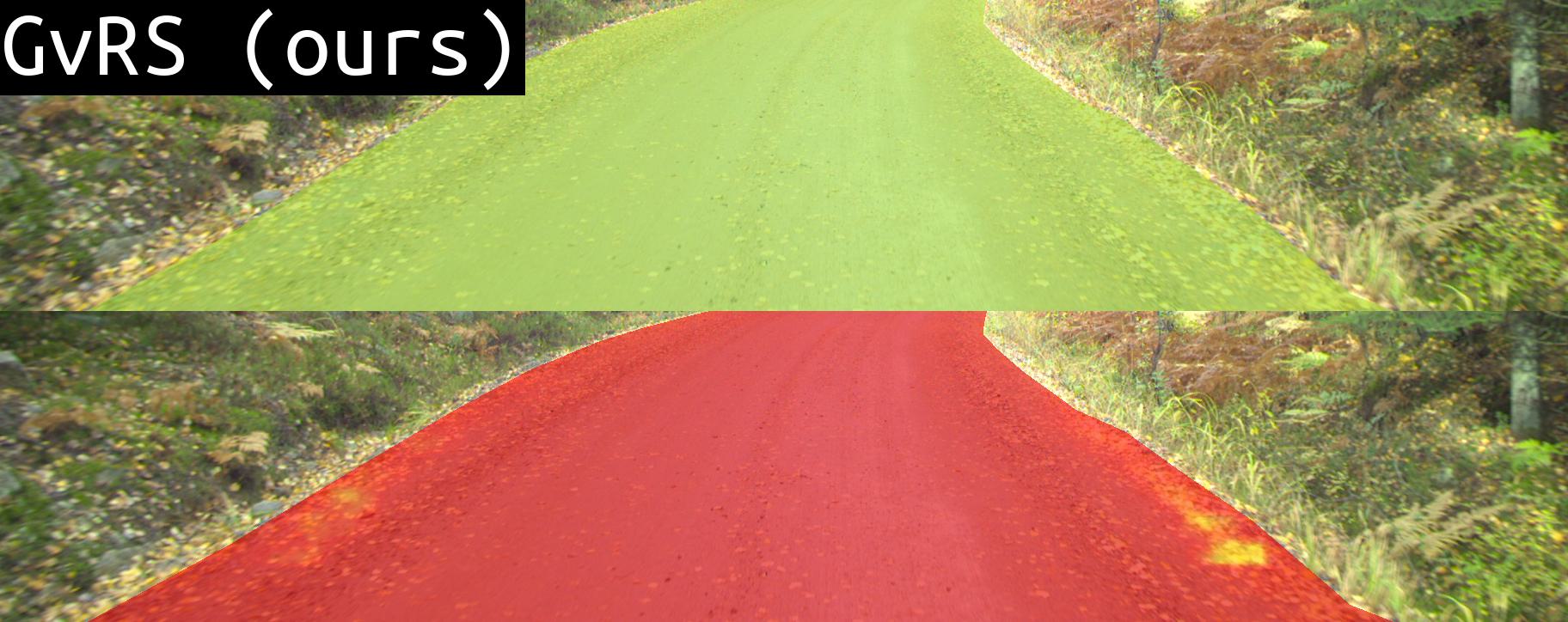} \\ \hline
\includegraphics[width=0.3\textwidth]{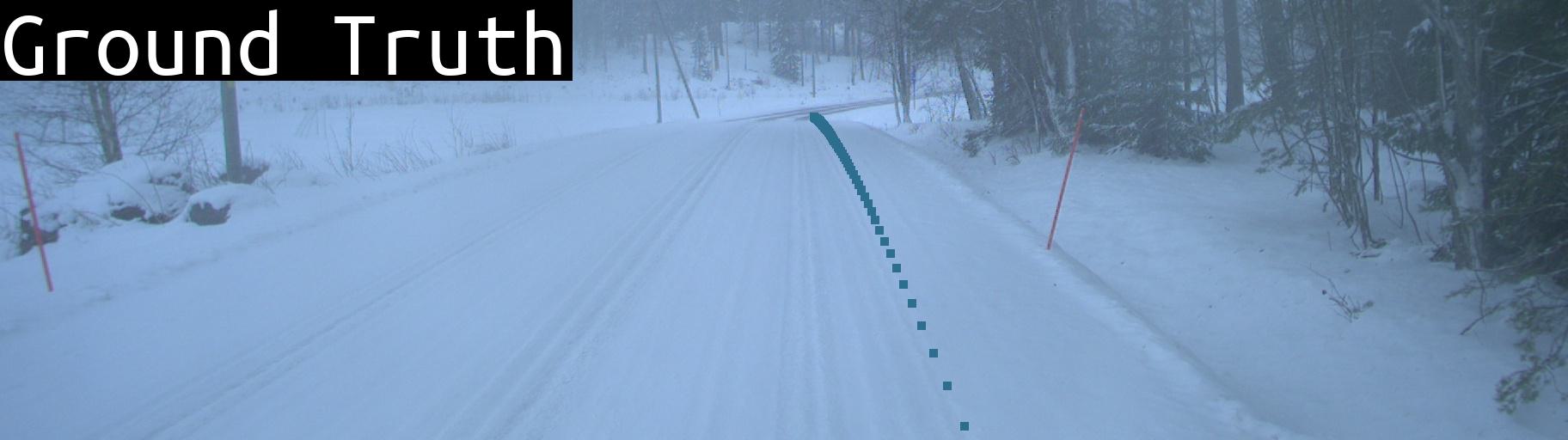} & \includegraphics[width=0.3\textwidth]{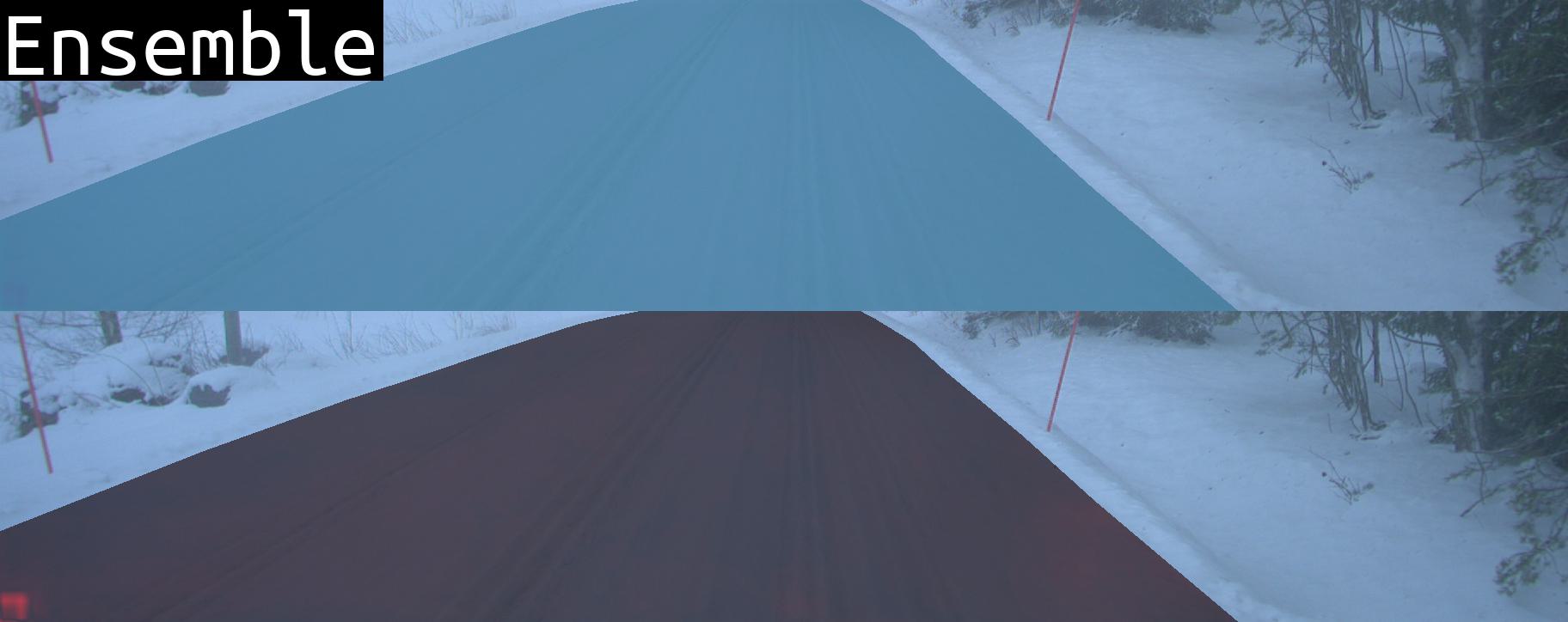} & \includegraphics[width=0.3\textwidth]{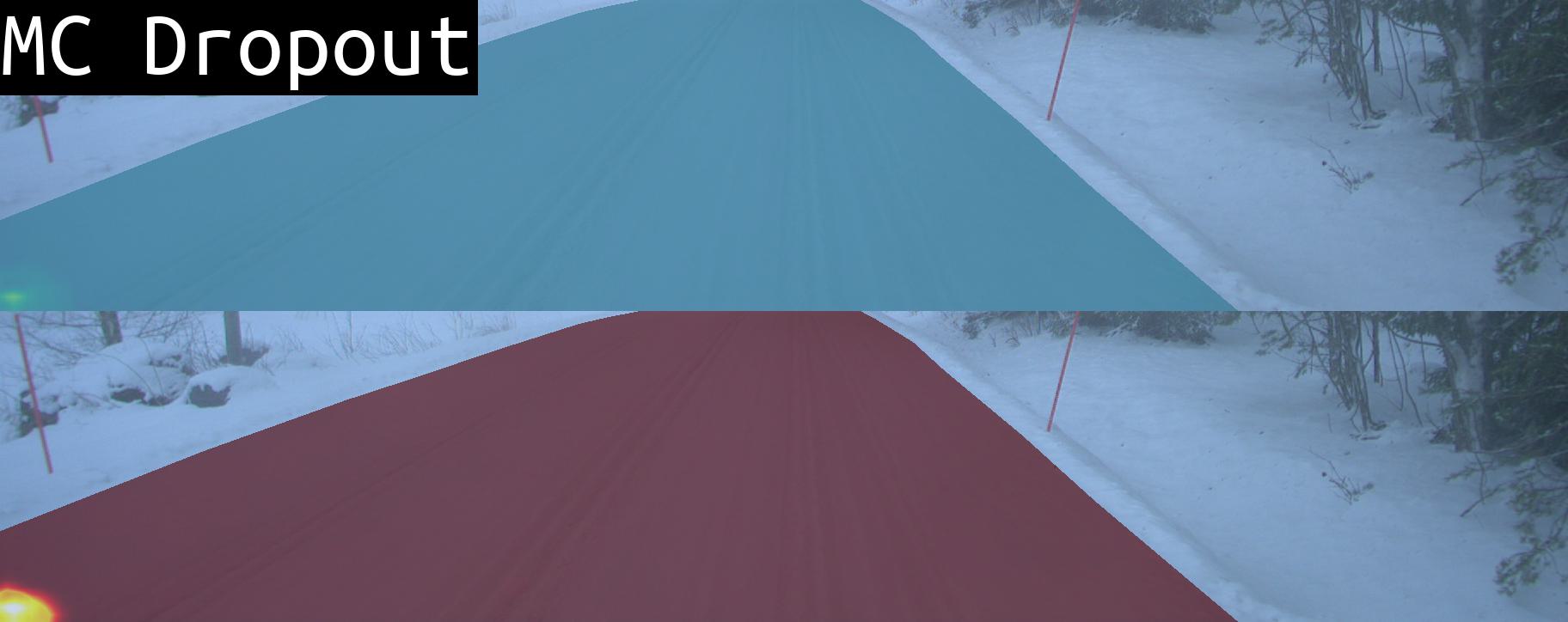} \\
\includegraphics[width=0.3\textwidth]{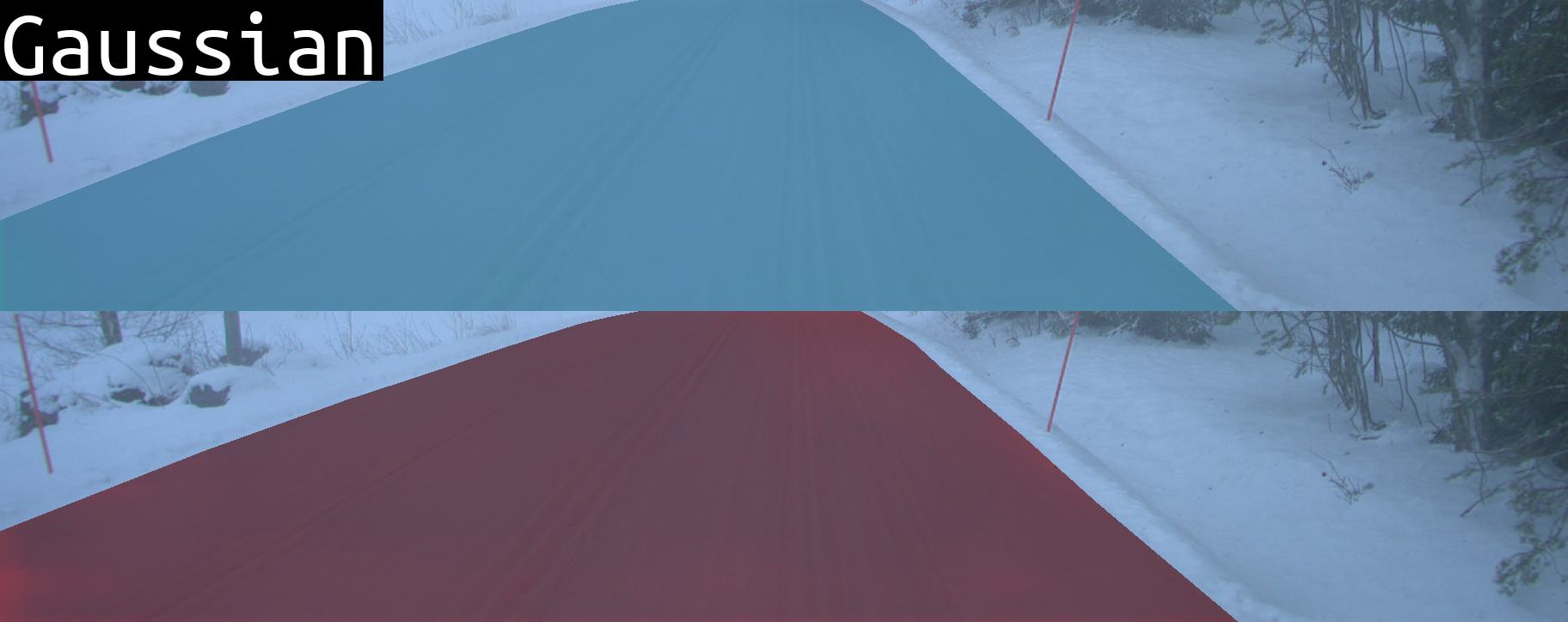} & \includegraphics[width=0.3\textwidth]{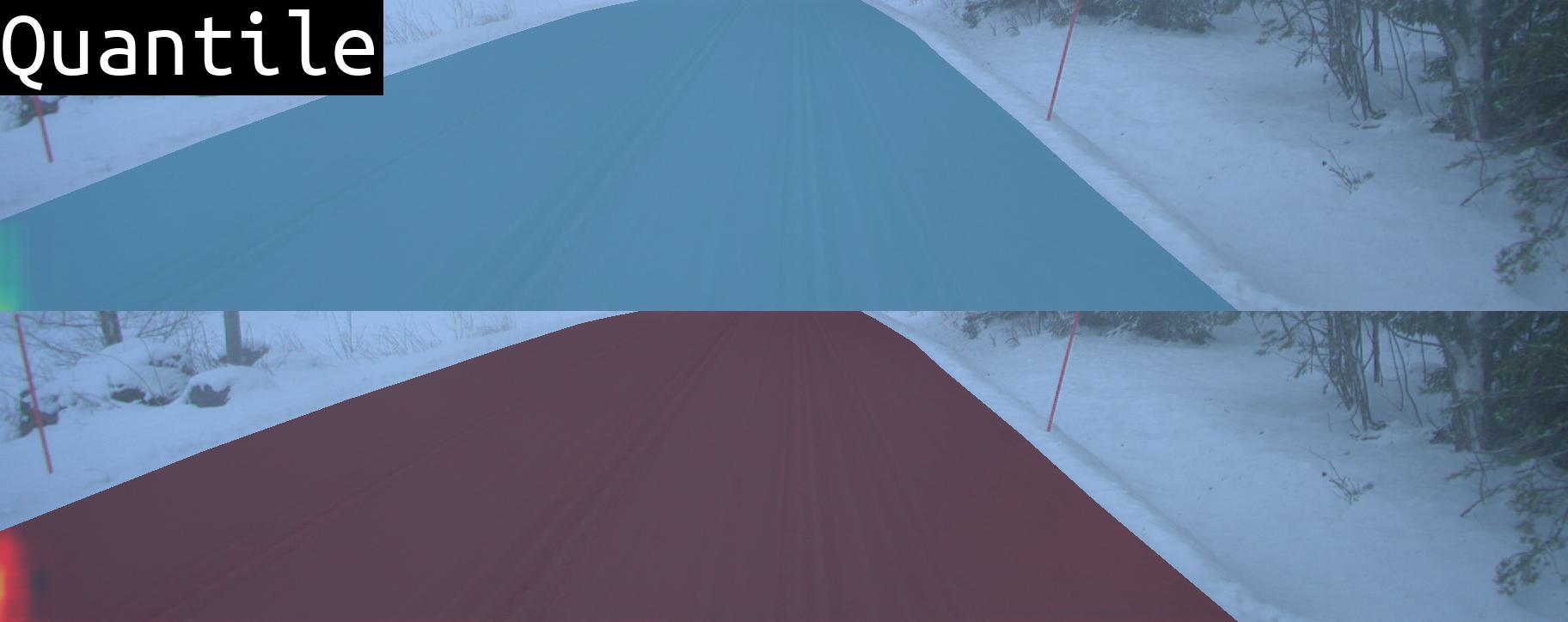} & \includegraphics[width=0.3\textwidth]{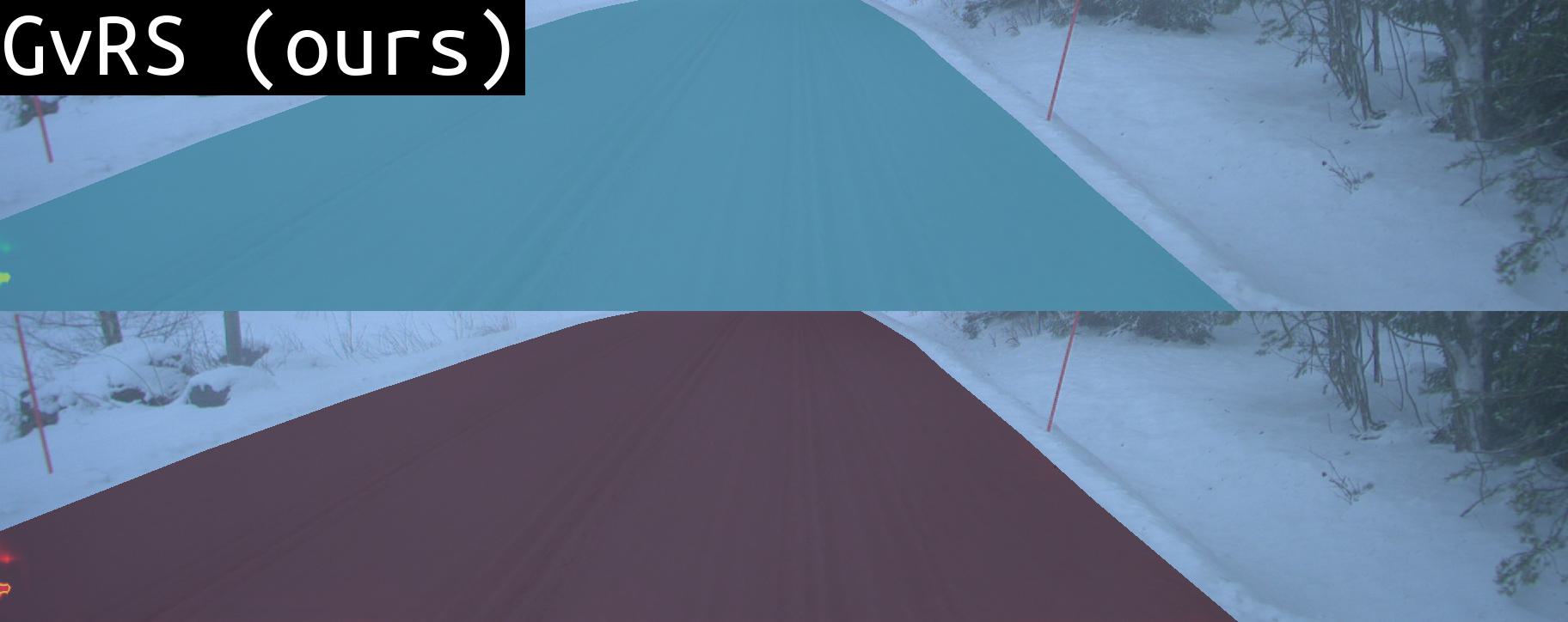} \\ \hline
\includegraphics[width=0.3\textwidth]{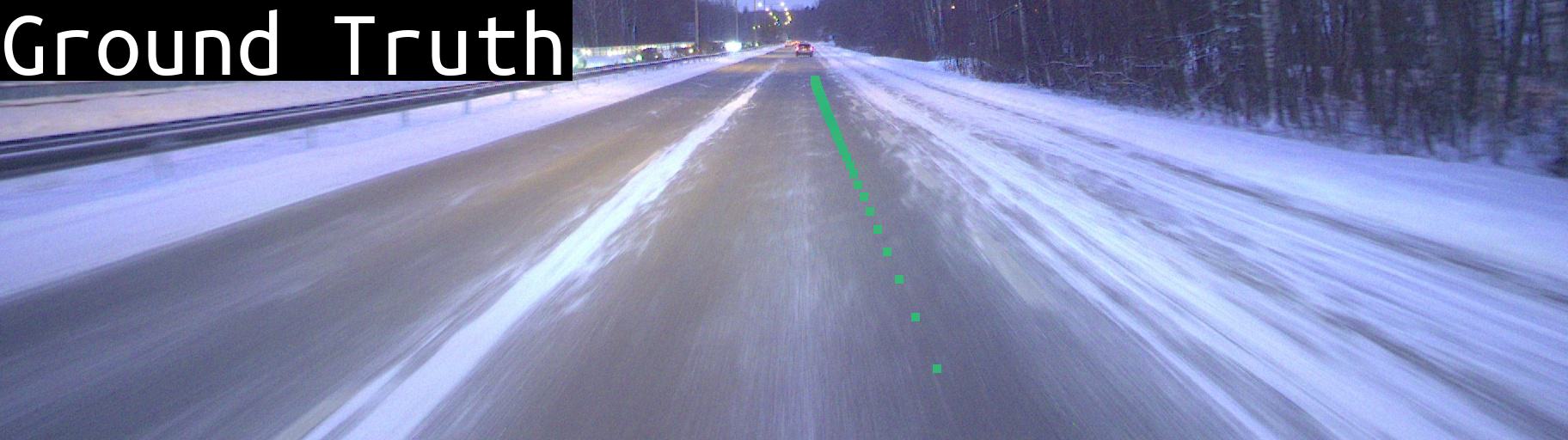} & \includegraphics[width=0.3\textwidth]{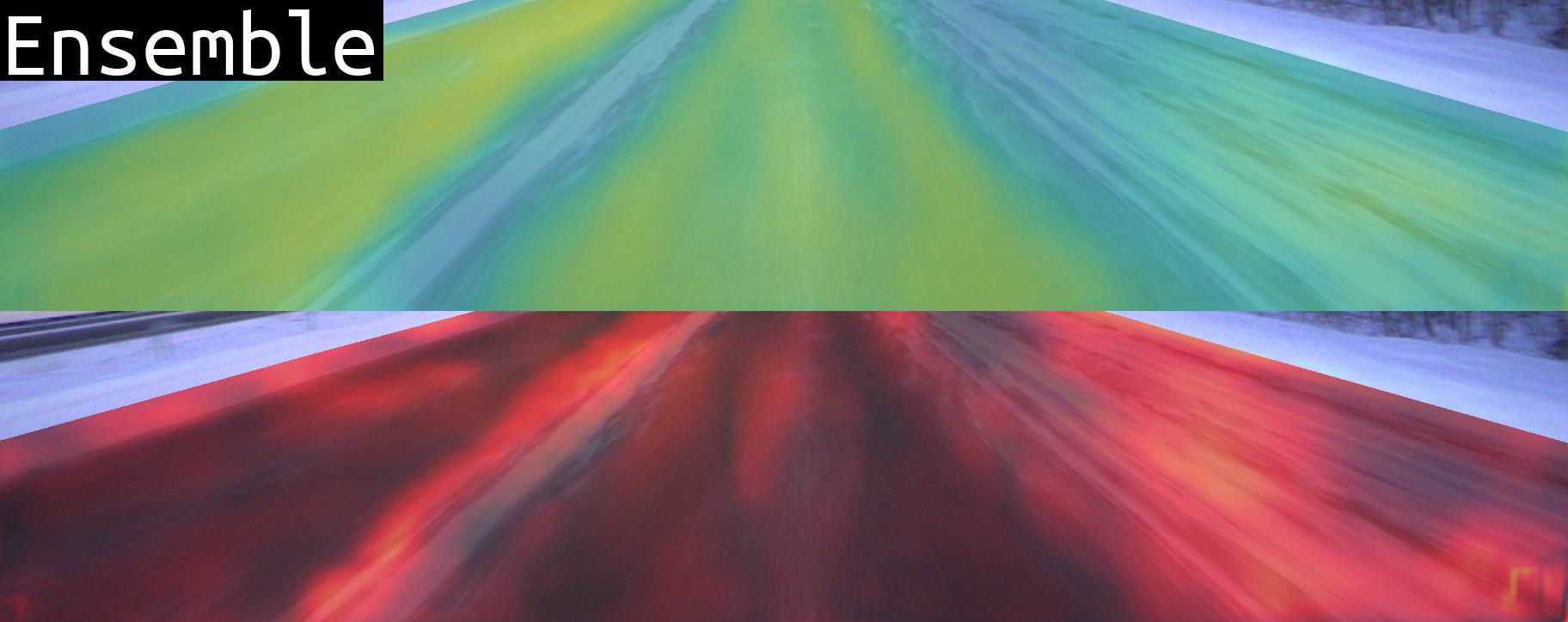} & \includegraphics[width=0.3\textwidth]{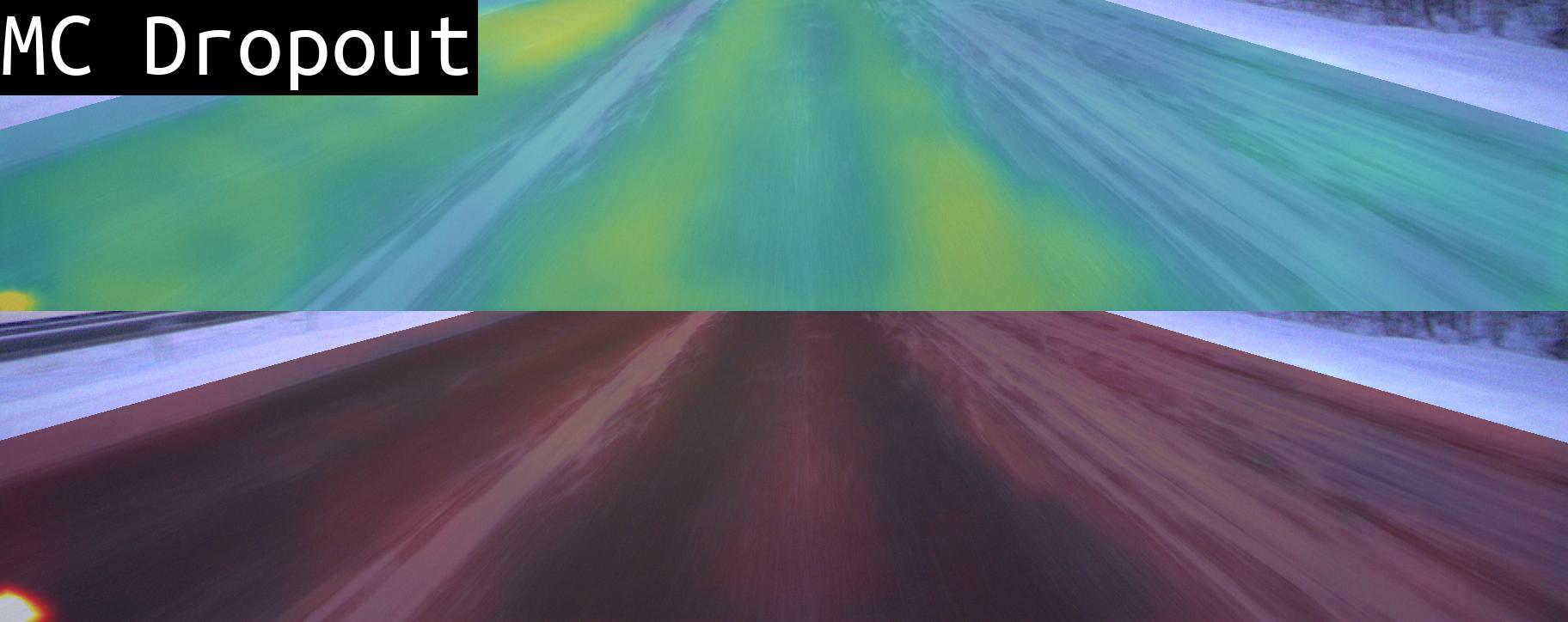} \\
\includegraphics[width=0.3\textwidth]{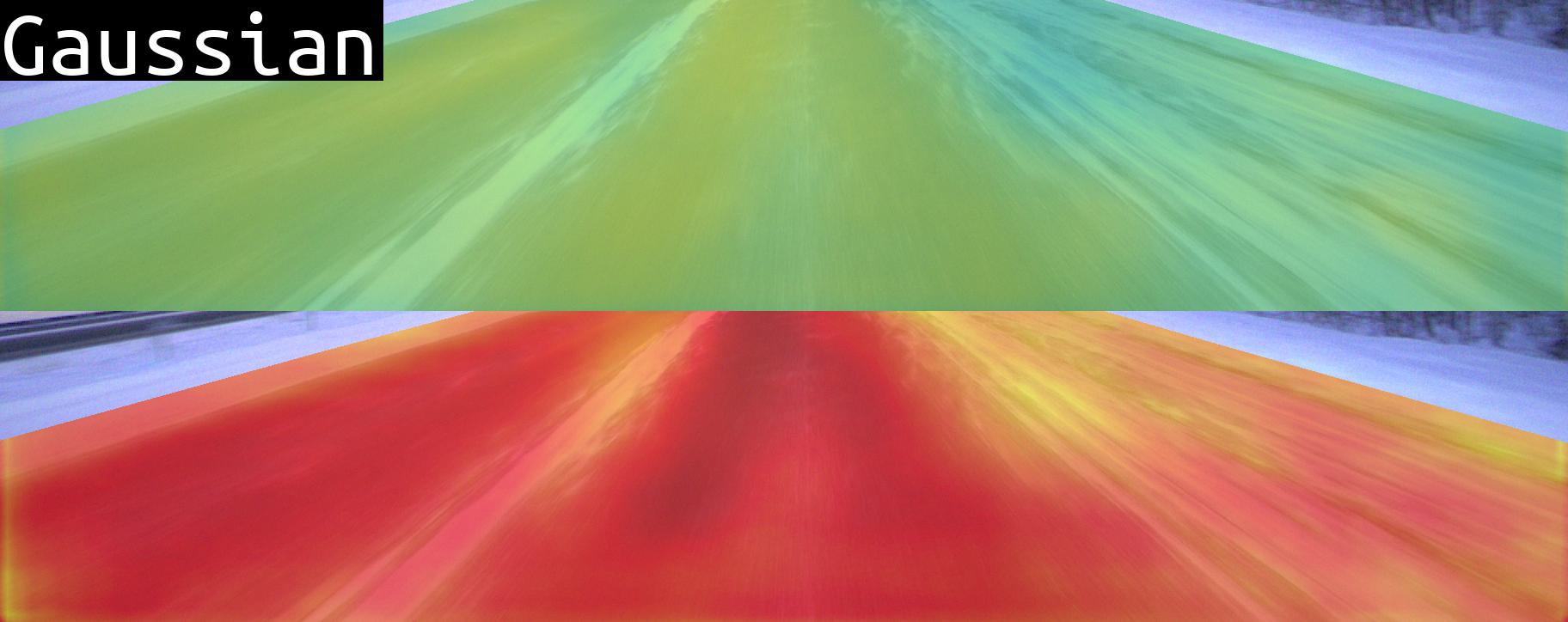} & \includegraphics[width=0.3\textwidth]{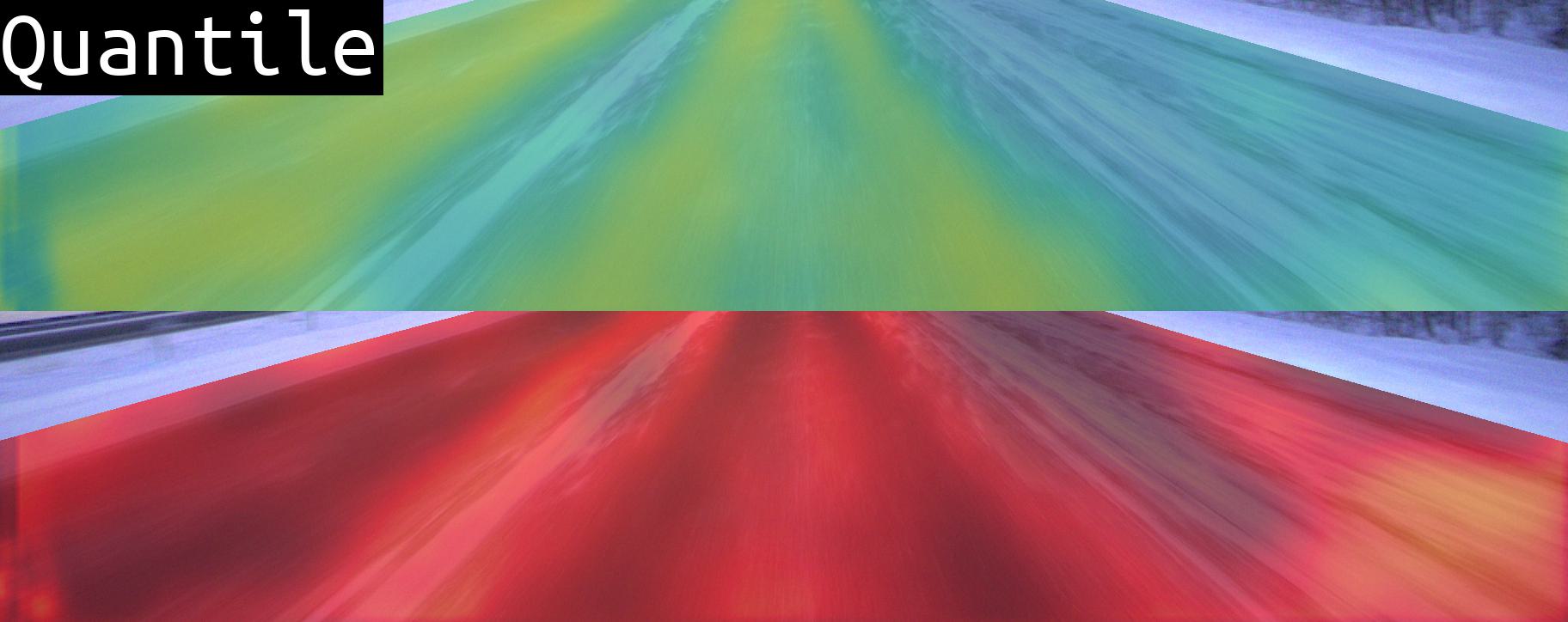} & \includegraphics[width=0.3\textwidth]{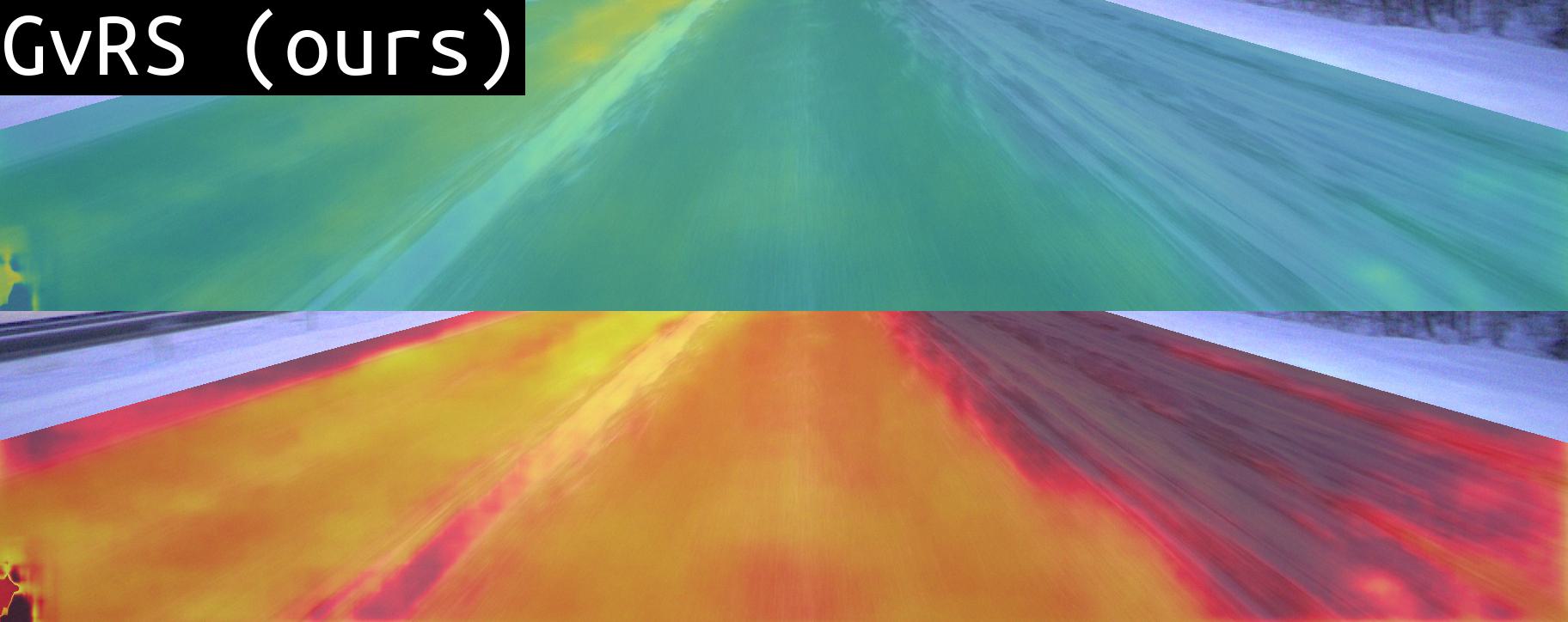} \\ \hline
\includegraphics[width=0.3\textwidth]{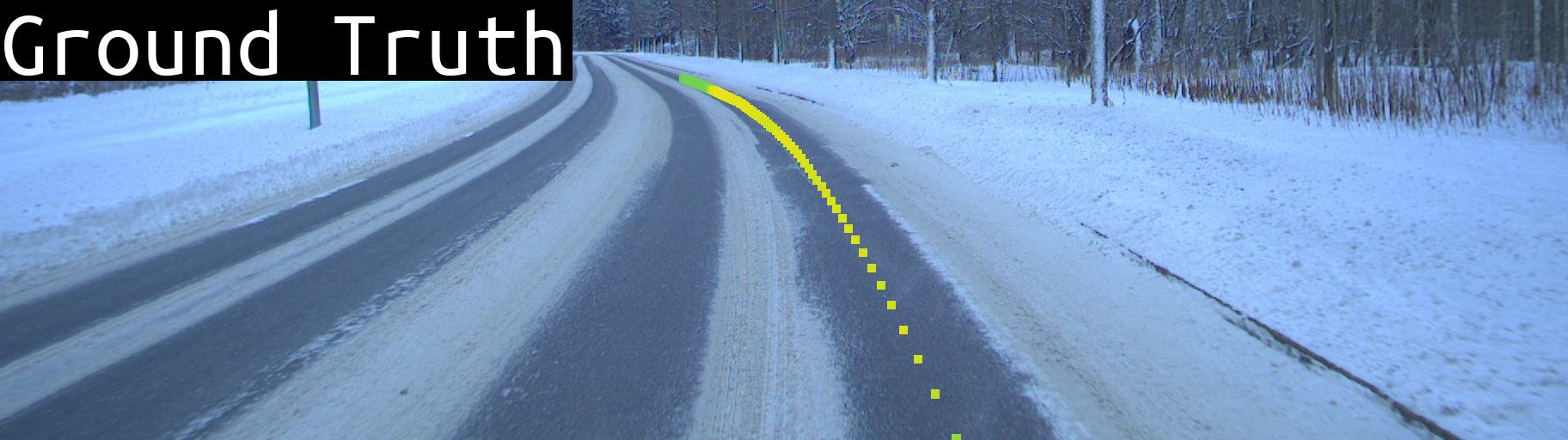} & \includegraphics[width=0.3\textwidth]{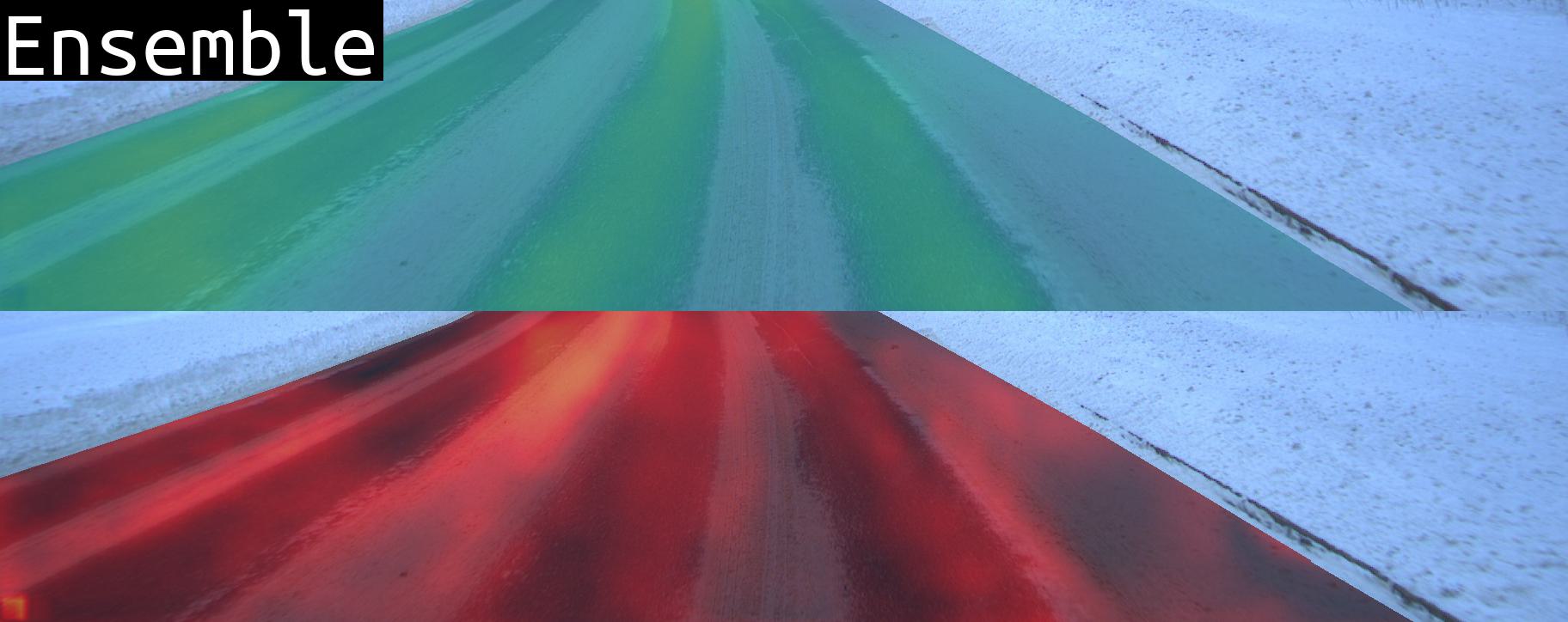} & \includegraphics[width=0.3\textwidth]{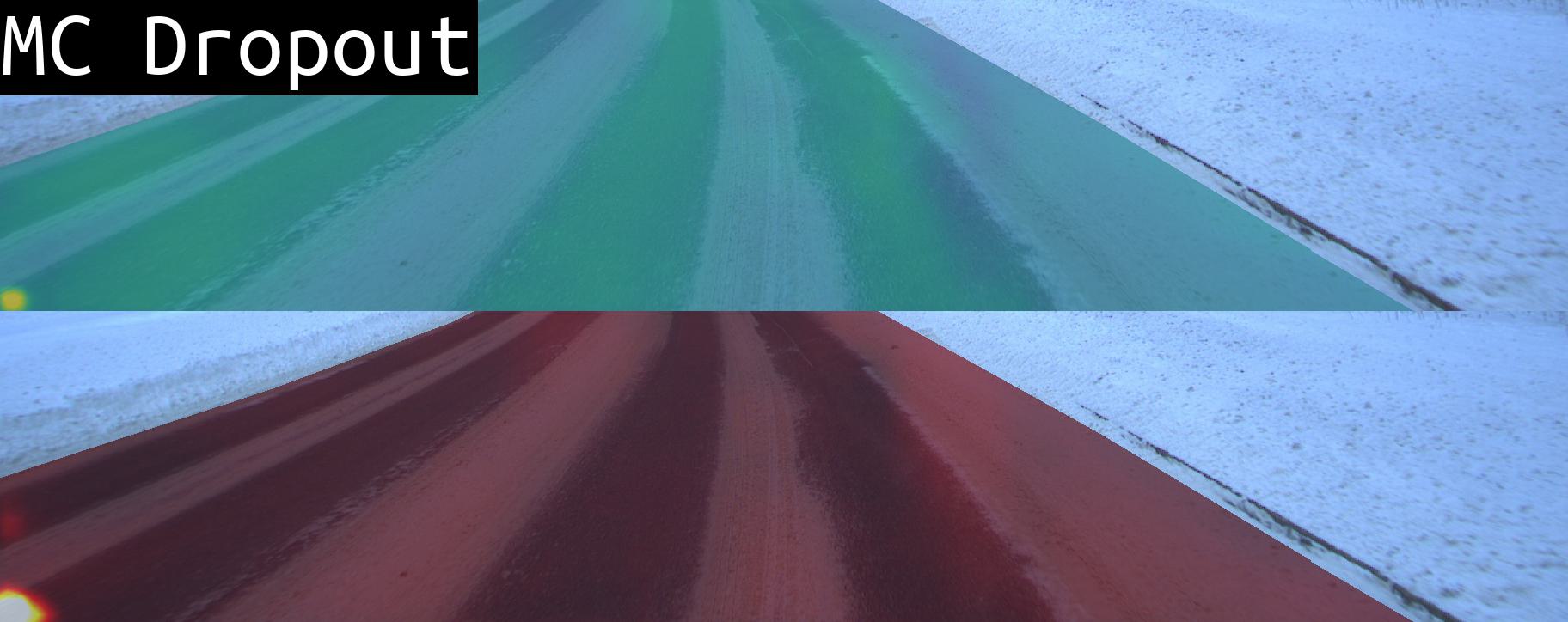} \\
\includegraphics[width=0.3\textwidth]{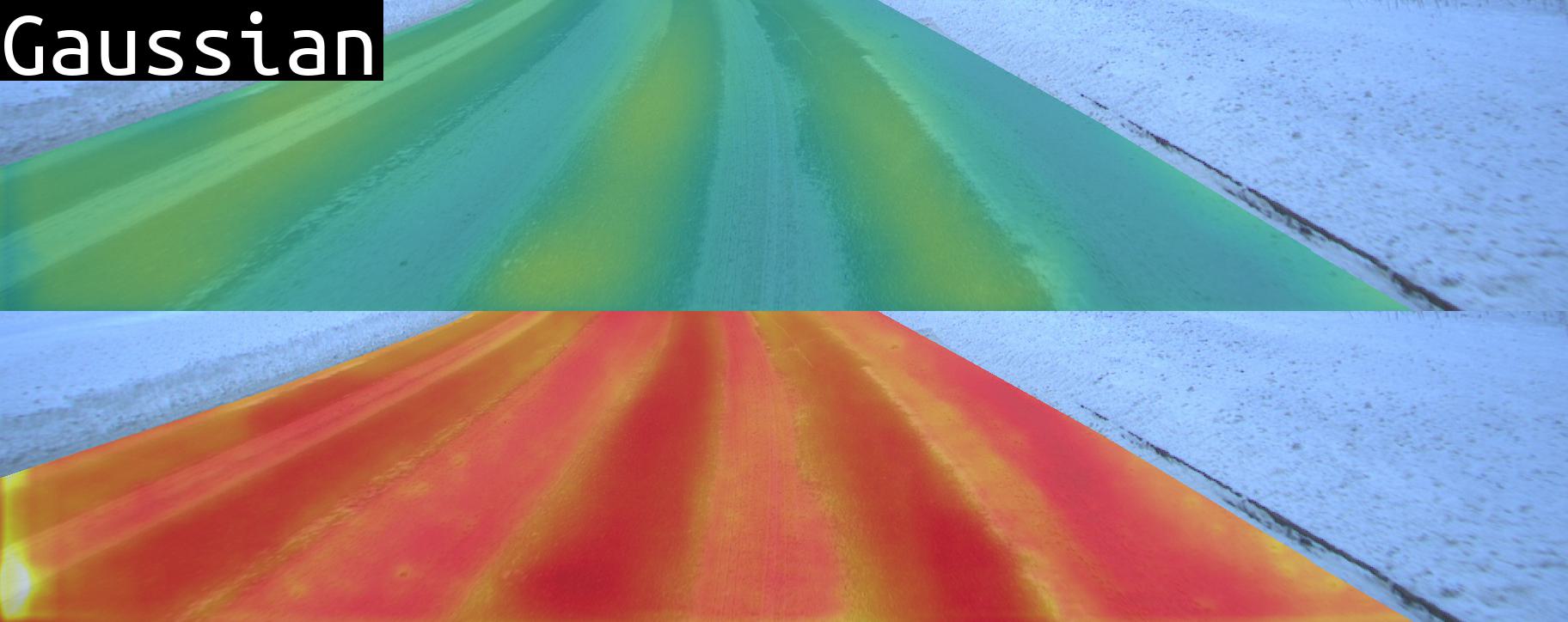} & \includegraphics[width=0.3\textwidth]{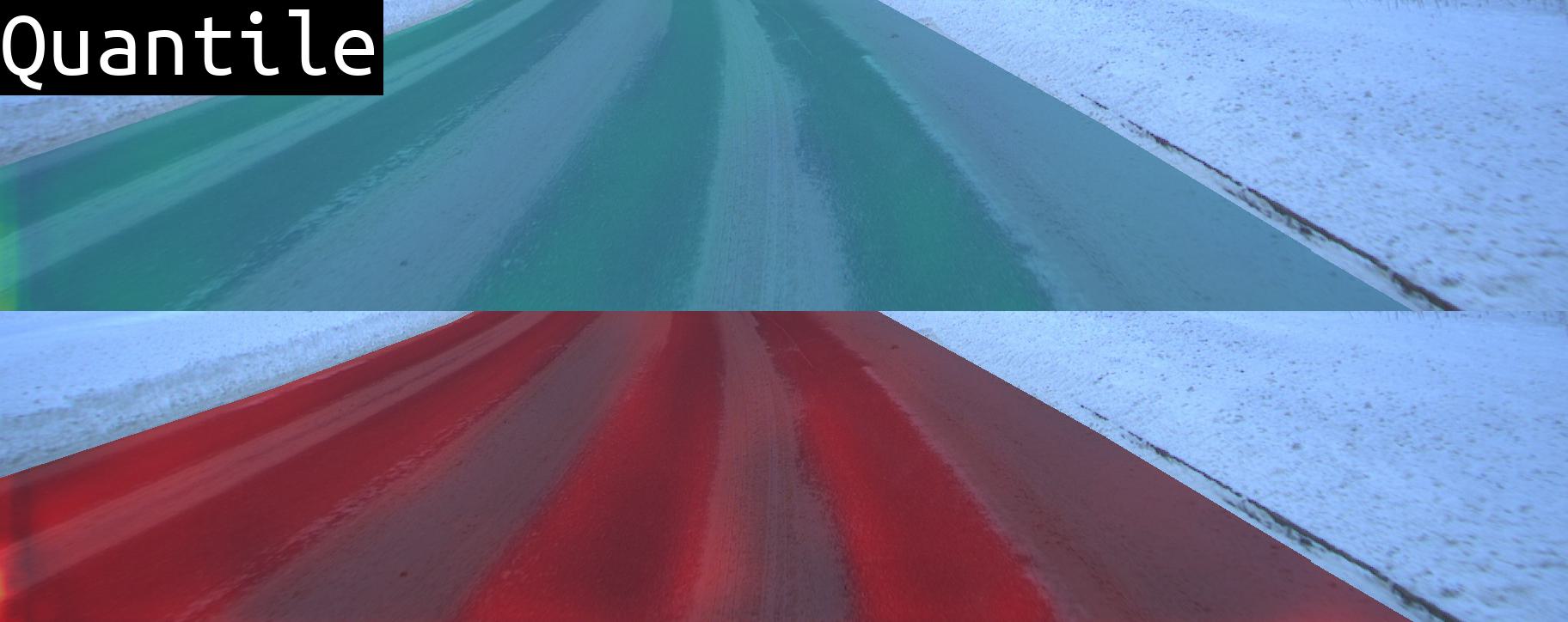} & \includegraphics[width=0.3\textwidth]{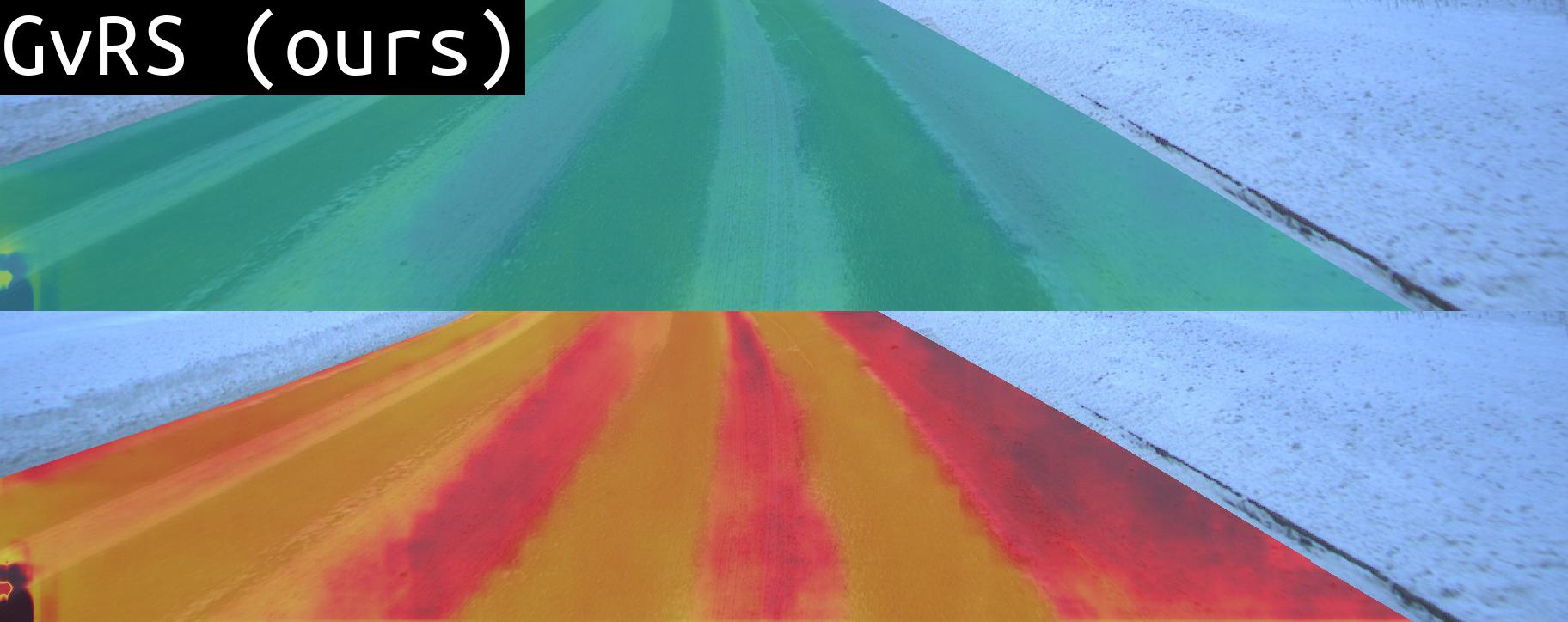} \\ \hline
\includegraphics[width=0.3\textwidth]{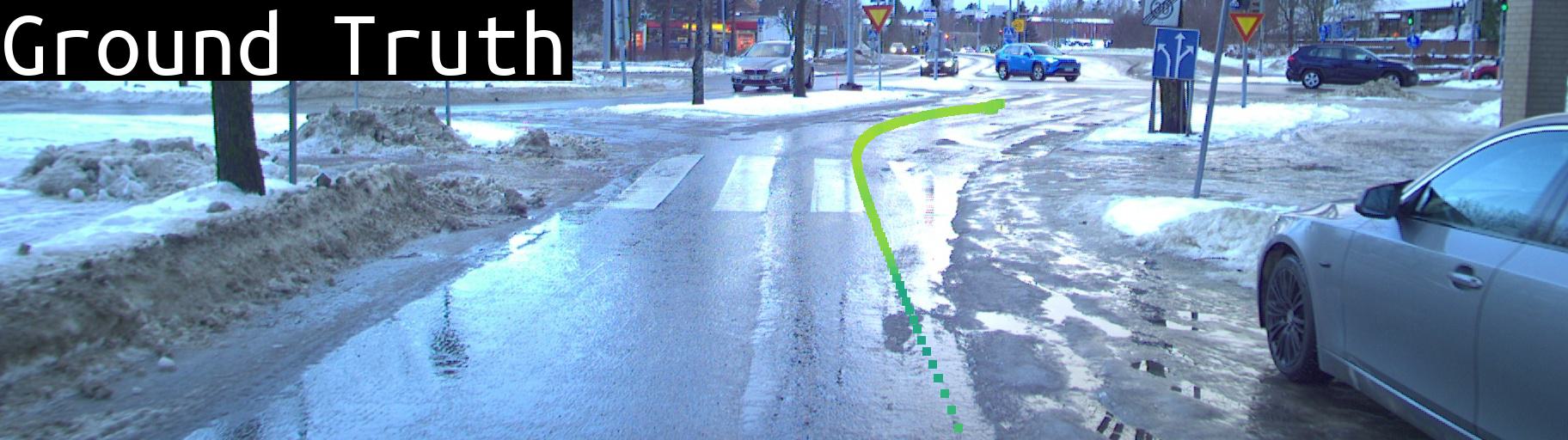} & \includegraphics[width=0.3\textwidth]{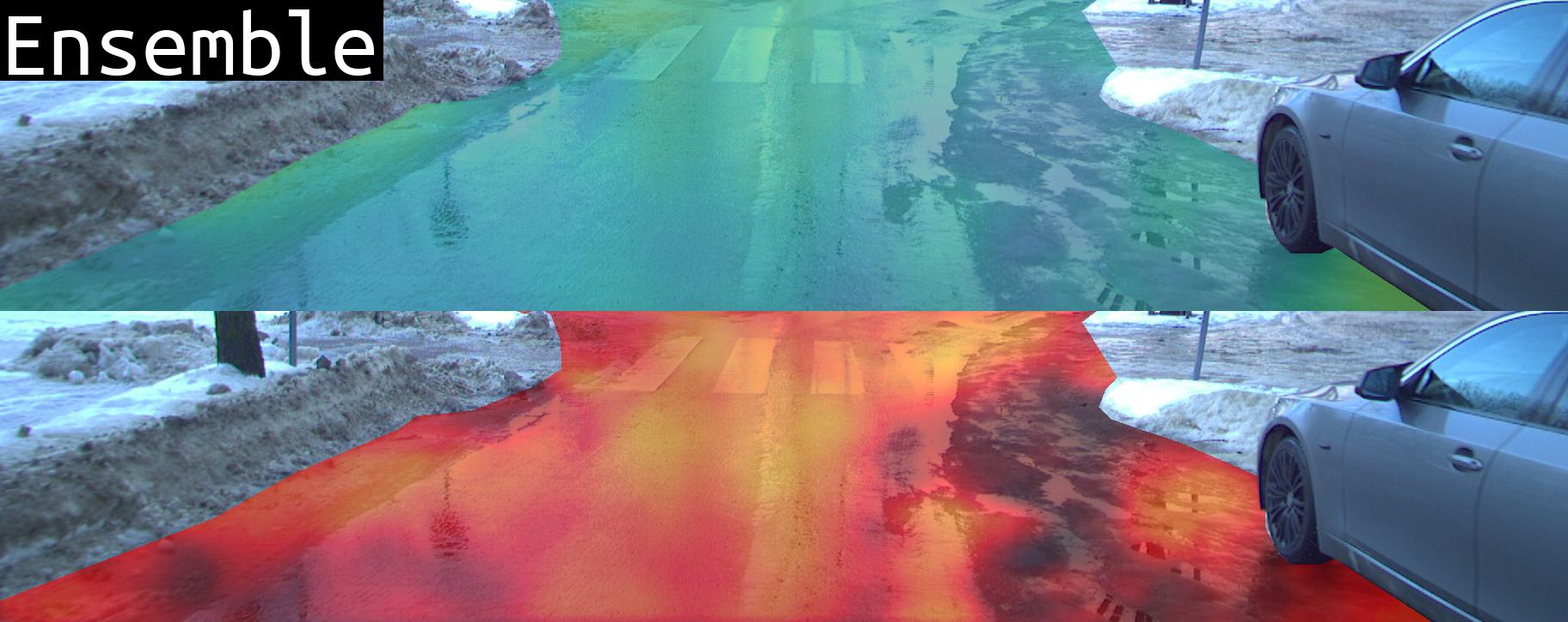} & \includegraphics[width=0.3\textwidth]{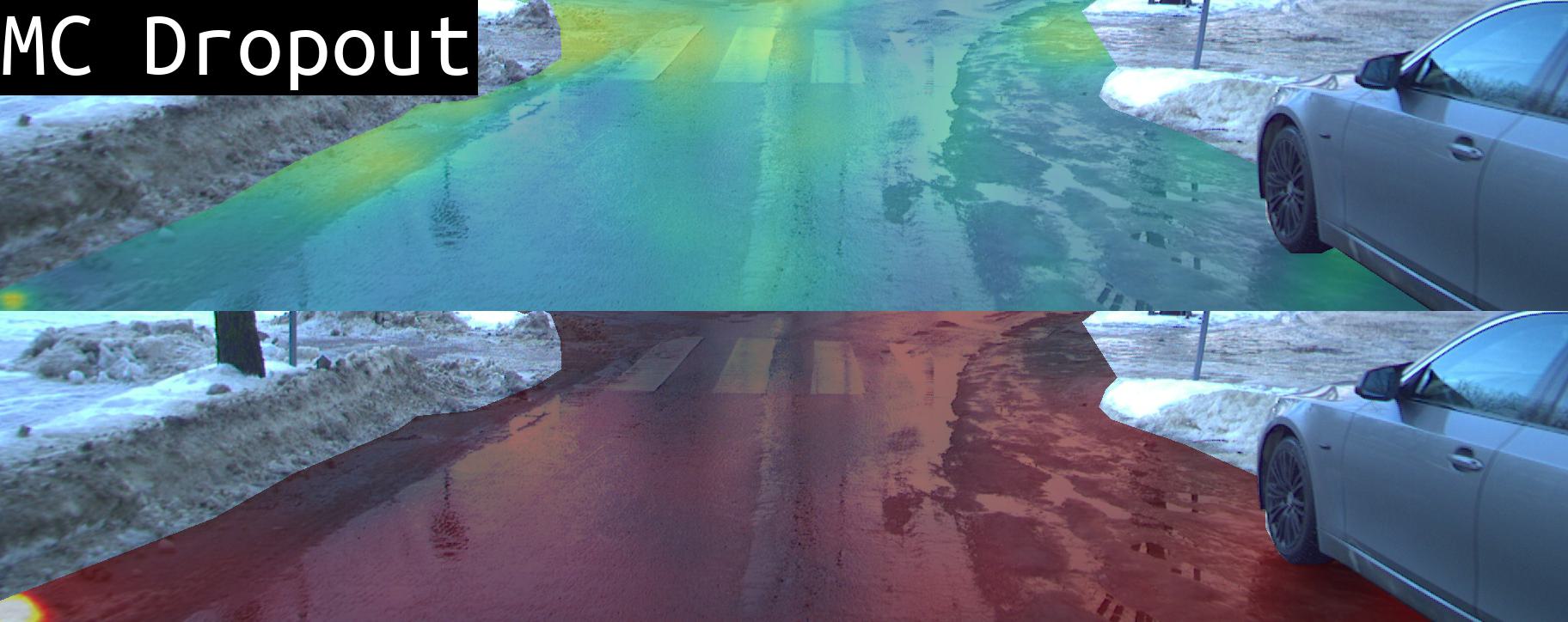} \\
\includegraphics[width=0.3\textwidth]{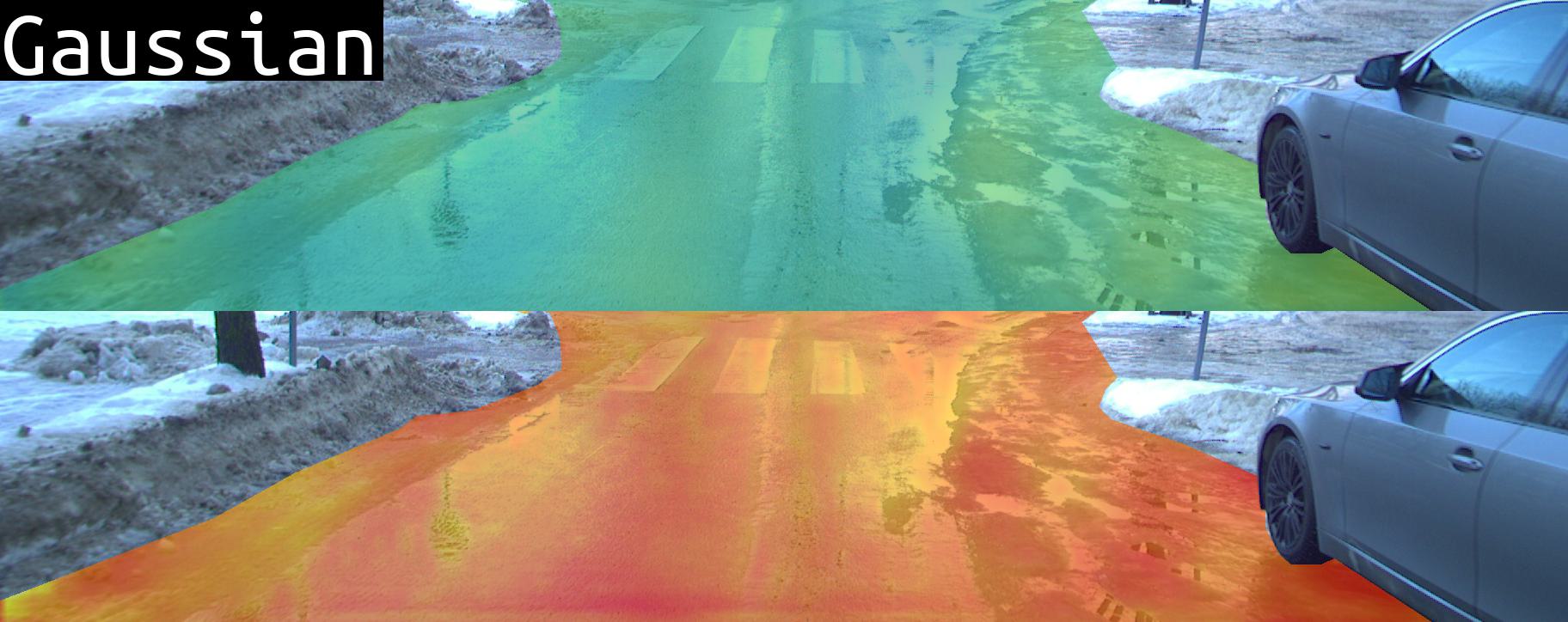} & \includegraphics[width=0.3\textwidth]{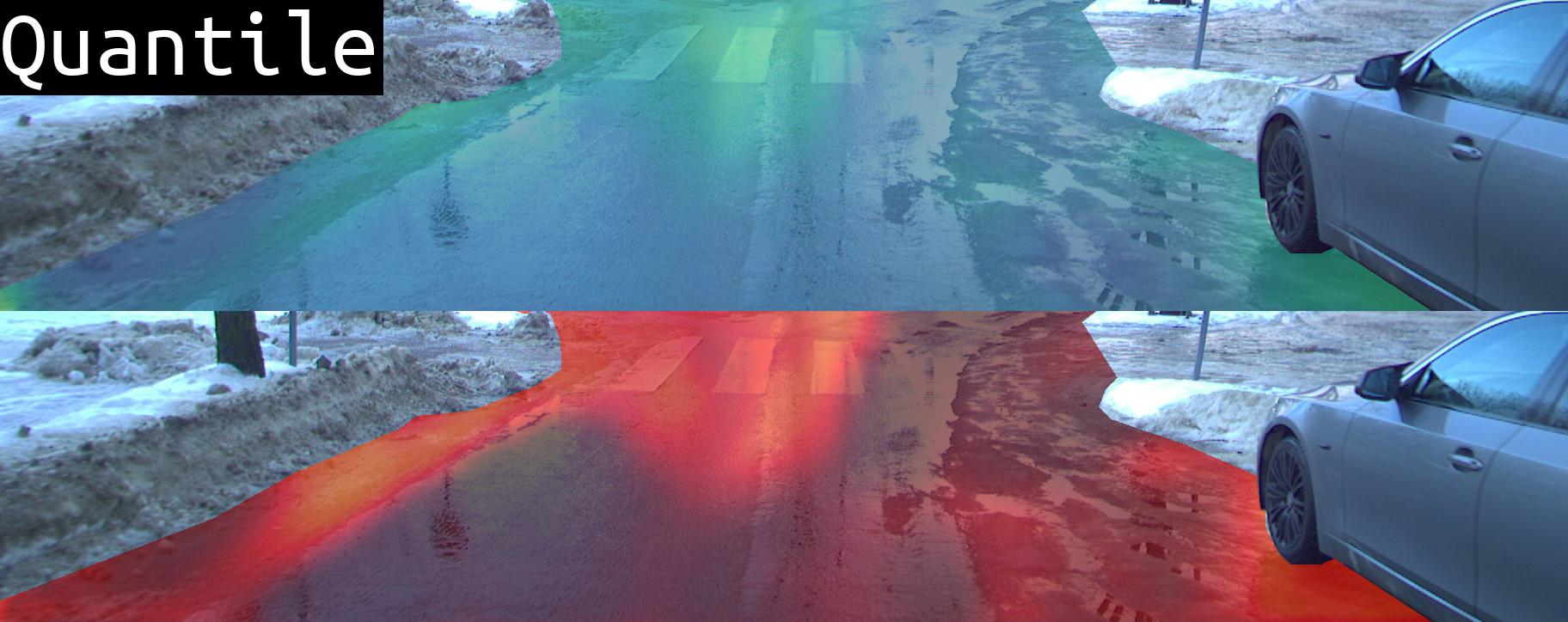} & \includegraphics[width=0.3\textwidth]{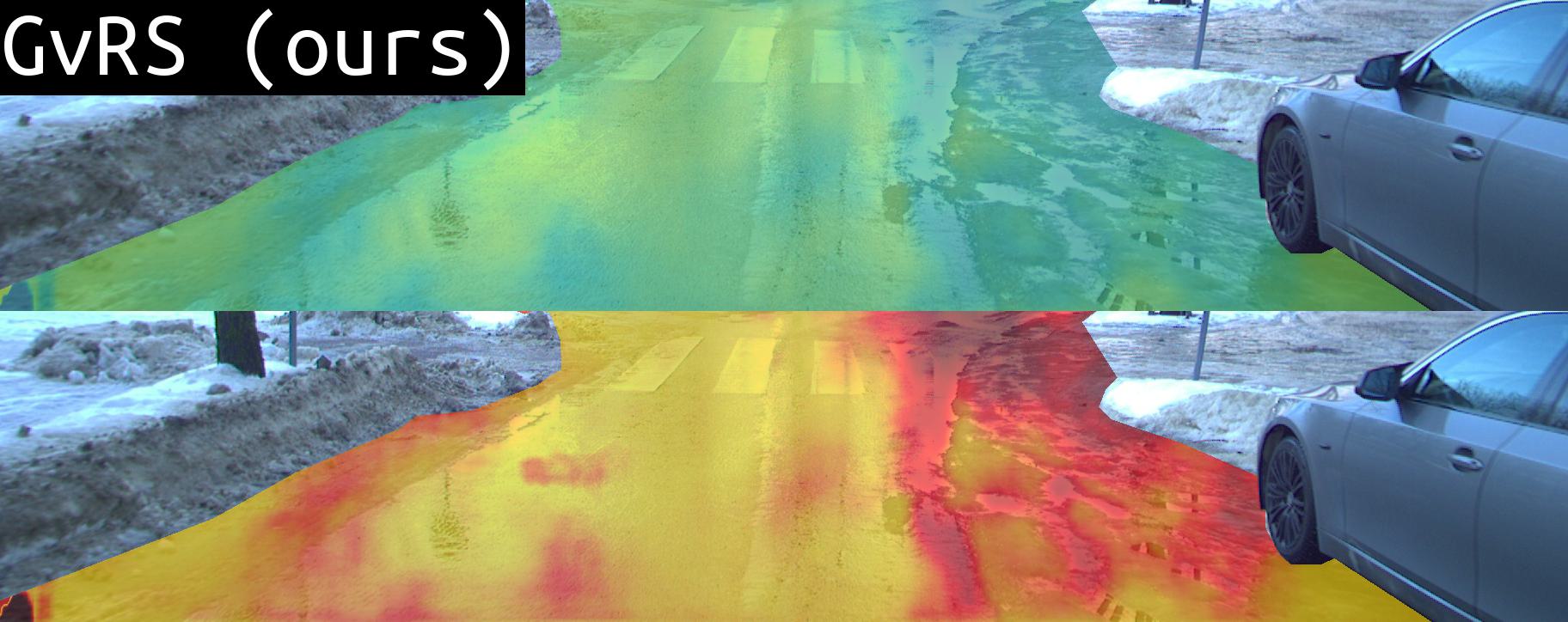}
\end{tabular}
\includegraphics[width=0.8\textwidth]{images/cbar_ready.pdf}
\caption{Additional visualizations of grip and grip uncertainty output on test set and test drive images. The first image for each example shows the ground truth data from the road weather sensor. For each model output, upper image shows the predicted grip distribution mean and the lower image shows the distance between predicted 5th percentile limit and the predicted grip distribution mean. The road area is manually segmented in the images for clarity.}\label{fig:extra_output_images}
\end{figure}

\end{document}